\documentclass[runningheads]{llncs}

\usepackage{eccv}

\usepackage{eccvabbrv}
\usepackage[utf8]{inputenc} % allow utf-8 input
\usepackage[T1]{fontenc}    % use 8-bit T1 fonts
\usepackage{lmodern}    % use 8-bit T1 fonts
\usepackage{booktabs}       % professional-quality tables
\usepackage{amsfonts}       % blackboard math symbols
\usepackage{nicefrac}       % compact symbols for 1/2, etc.
\usepackage{microtype}      % microtypography
\usepackage{amsmath}
\usepackage{graphicx}
\usepackage{subcaption}
\usepackage{algorithm}
\usepackage{algpseudocode}
\usepackage{mathtools}
\usepackage{longtable}

\usepackage{minitoc}

\renewcommand \thepart{}
\renewcommand \partname{}

\usepackage[accsupp]{axessibility}  % Improves PDF readability for those with disabilities.

\newcommand{\boldparagraph}[1]{\paragraph{\textbf{\textup{#1}}}}

\usepackage[dvipsnames]{xcolor}

\definecolor{review1}{rgb}{0.89,0.10,0.10}
\definecolor{review2}{rgb}{0.21,0.5,0.72}
\definecolor{review3}{rgb}{0.3,0.68,0.29}
\definecolor{review4}{rgb}{0.59,0.3,0.64}
\definecolor{review5}{rgb}{1,0.5,0}

\usepackage{diagbox} % for diagbox in tables
\usepackage{multirow} % for multirow headers in large table
\usepackage{colortbl} % for background color in table cells
\newcommand{\bestcolor}[0]{green!25}
\definecolor{impractical}{RGB}{255, 0, 0}
\usepackage{amssymb} % for \ding{51} and \ding{55}, i.e., checkmark and cross
\usepackage{pifont} % for \ding{51} and \ding{55}, i.e., checkmark and cross
\newcommand{\cmark}{\ding{51}}%
\newcommand{\xmark}{\ding{55}}%
\usepackage{nicefrac} % for \nicefrac, which results in nicer inline fractions
\def\estRscod{\widehat{\rm R}} %_{\rm SCOD}
\def\estRsel{\widehat{\rm R}_{\rm S}}
\def\esttpr{\widehat{\tpr}}
\def\estfpr{\widehat{\fpr}}
\def\areaSCODTpr{{\rm AuSRT}}
\def\areaRiskCoverage{{\rm AuRC}}
\def\areaROC{{\rm AuROC}}

\def\det#1{{\rm det(#1)}}
\def\Algo{{POSCOD}}
\def\tpr{{\rm tpr}}
\def\fpr{{\rm fpr}}
\def\Pout{\pi_{\rm O}}

\def\relCost{\alpha}
\def\tprMin{{\rm tpr}_{\rm min}}
\def\covMin{\rho_{\rm min}}
\def\Rsel{{\rm R_{\rm S}}}
\def\reject{{\rm reject}}

\def\beq{\begin{equation}}
\def\eeq{\end{equation}}
\def\beqn{\begin{eqnarray}}
\def\eeqn{\end{eqnarray}}
\def\beqns{\begin{eqnarray*}}
\def\eeqns{\end{eqnarray*}}

\def\Argmax{\mathop{\rm Argmax}}
\def\Argmin{\mathop{\rm Argmin}}

\def\ST{{\cal T}}

\def\EE{{\mathbb{E}}}

\def\SD{{\cal D}}

\def\SP{{\cal P}}

\def\SX{{\cal X}}
\def\SY{{\cal Y}}

\def\SN{{\cal N}}
\def\SH{{\cal H}}

\def\SD{{\cal D}}

\def\NN{\mathbb{N}}

\renewcommand{\Re}{\mathbb{R}}
\def\veps{\varepsilon}

\def\equ#1{(\ref{#1})}

\def\reject{{\rm reject}}

\def\##1{\relax\ifmmode\mathchoice      % italic bold vector, e.g. $\#{x}$
{\mbox{\boldmath$\displaystyle#1$}}
 {\mbox{\boldmath$\textstyle#1$}}
{\mbox{\boldmath$\scriptstyle#1$}}
{\mbox{\boldmath$\scriptscriptstyle#1$}}\else
\hbox{\boldmath$\textstyle#1$}\fi}
\def\leftbb{\mathopen{\rlap{$[$}\hskip1.3pt[}}
\def\rightbb{\mathclose{\rlap{$]$}\hskip1.3pt]}}

\usepackage[pagebackref,breaklinks,colorlinks,citecolor=eccvblue]{hyperref}
\begin{document}

\doparttoc % Tell to minitoc to generate a toc for the parts
\faketableofcontents % Run a fake tableofcontents command for the partocs

%\title{Theoretically grounded SCOD} 
\title{SCOD: From Heuristics to Theory}

%\titlerunning{Theoretically grounded SCOD}
\titlerunning{SCOD: From Heuristics to Theory}
%\author{Vojtech Franc\inst{1}\orcidlink{0000-1111-2222-3333} \and
%Jakub Paplham\inst{1}\orcidlink{1111-2222-3333-4444} \and
%Daniel Prusa\inst{1}\orcidlink{2222--3333-4444-5555}}
\author{Vojtech Franc\inst{1} \and
Jakub Paplham\inst{1} \and
Daniel Prusa\inst{1}}

%Vojtech Franc (Czech Technical University in Prague) <xfrancv@fel.cvut.cz> 
%Jakub Paplham (Czech Technical University in Prague) <paplhjak@fel.cvut.cz> 
%Daniel Prusa (Czech technical university in Prague) <prusapa1@cmp.felk.cvut.cz> 
%Primary Subject Area

\authorrunning{V.~Franc et al.}

\institute{Department of Cybernetics, Fakulty of Electrical Engineering, Czech Technical University in Prague, Czech Republic}
%\and
%Springer Heidelberg, Tiergartenstr.~17, 69121 Heidelberg, Germany

%\institute{Princeton University, Princeton NJ 08544, USA \and
%Springer Heidelberg, Tiergartenstr.~17, 69121 Heidelberg, Germany

\maketitle

\begin{abstract}
This paper addresses the problem of designing reliable prediction models that abstain from predictions when faced with uncertain or out-of-distribution samples - a recently proposed problem known as Selective Classification in the presence of Out-of-Distribution data (SCOD). We make three key contributions to SCOD. Firstly, we demonstrate that the optimal SCOD strategy involves a Bayes classifier for in-distribution (ID) data and a selector represented as a stochastic linear classifier in a 2D space, using i) the conditional risk of the ID classifier, and ii) the likelihood ratio of ID and out-of-distribution (OOD) data as input. This contrasts with suboptimal strategies from current OOD detection methods and the Softmax Information Retaining Combination (SIRC), specifically developed for SCOD. Secondly, we establish that in a distribution-free setting, the SCOD problem is not Probably Approximately Correct learnable when relying solely on an ID data sample. Third, we introduce \Algo{}, a simple method for learning a plugin estimate of the optimal SCOD strategy from both an ID data sample and an unlabeled mixture of ID and OOD data. Our empirical results confirm the theoretical findings and demonstrate that our proposed method, \Algo{}, outperforms existing OOD methods in effectively addressing the SCOD problem.

\keywords{out-of-distribution detection \and selective classification \and optimal strategy \and probably approximately correct learning}
\end{abstract}    
\section{Introduction}
\label{sec:intro}
Standard methods for learning predictors from data rely on the closed-world assumption, i.e., the training and testing samples are generated from the same distribution, so-called In-Distribution (ID). In real-world applications, ID test samples can be contaminated by samples from another distribution, the so-called Out-Of-Distribution (OOD), which is not represented by the training sample. In recent years, the growing interest in deep learning models capable of handling OOD data has resulted in numerous papers on effective OOD detection (OODD)~\cite{Hendrycks-baseline-ICLR17,liang2018enhancing,Dhamija-NIPS2018,Devries-arxiv2018,Malinin-NEURIPS2018,Granese-Nips2021,Chen-PAMI2022,Sun-NIPS2021,Song-Neurips2022,pmlr-v162-sun22d,Wang-ViM-CVPR2022}. Notably, despite the practical role of OOD detector as selector of input samples for ID classifier, prior work has not explicitly addressed the consideration of misclassified ID samples in selector design. The concept of classifiers equipped with selectors to reject predictions on likely misclassified input samples, known as selective classification (SC), has been studied separately in closed-world scenario~\cite{Chow-RejectOpt-TIT1970,Pietraszek-AbstainROC-ICML2005,Geifman-SelectClass-NIPS2017,Franc-Optimal-JMLR2023}. Recent research~\cite{Kim-UnifBench-ESA2021,Xia-SIRC-ACCV2022,Cen-Devil-ICLR2023,Narasimhan-Plugin-Arxiv2023} underscores the need for selective classifiers that simultaneously address OODD and SC goals. The newly introduced prediction problem is called Selective Classification in the presence of Out-of-Distribution data (SCOD)\cite{Xia-SIRC-ACCV2022}. 

SCOD aims to detect OOD samples, abstaining from predictions on them, while simultaneously minimizing the prediction error on accepted ID samples. To address this problem, \cite{Xia-SIRC-ACCV2022} introduces the Softmax Information Retaining Combination (SIRC) heuristic strategy. SIRC constructs selectors by combining two scores, one focused on detecting misclassified ID samples and the other on identifying OOD samples. Despite demonstrating superior performance over existing OOD detectors in the SCOD problem, as shown in empirical evidence by~\cite{Xia-SIRC-ACCV2022}, it remains unclear whether the SIRC strategy is optimal and whether the SCOD problem can be solved from the available data. 
In this paper, we address these questions, leveraging the answers to propose a theoretically grounded approach that consistently outperforms existing methods. 
%A trustworthy prediction model should detect OOD samples and refrain from making predictions on them, while simultaneously minimizing the prediction error on accepted ID samples. 
In particular, we provide the following contributions to the SCOD problem:
\begin{enumerate}
  \item We demonstrate that the optimal prediction strategy for solving the SCOD problem comprises the Bayes classifier for ID data and a selector represented as a stochastic linear classifier in a 2D space. The input features for this selector are the conditional risk of the ID classifier and the OOD/ID likelihood ratio. Our findings reveal that current OODD methods, as well as the SIRC, yield suboptimal strategies for the SCOD problem.
  \item We extend the concept of Probably Approximately Correct (PAC) learnability~\cite{Shwartz-ML-2014,Zhen-IsOODLearnable-NIPS2022} to address the SCOD problem. Additionaly, we prove that in a distribution-free setting, the SCOD problem is not PAC-learnable when the learning algorithm exclusively depends on an ID data sample.
  \item We introduce a method \Algo{} for learning the plugin estimate of the optimal SCOD strategy from both an ID data sample and an unlabeled mixture of ID and OOD data. \Algo{} simplifies the learning process to i) training an ID classifier through standard cross-entropy loss and ii) training a classifier using the binary cross-entropy (BCE) of a novel corrected sigmoid.
  \item We empirically confirm our theoretical findings and demonstrate that our proposed method, \Algo{}, outperforms existing OODD methods and SIRC when applied to the SCOD problem.
\end{enumerate}
It is worth noting that while the initial two contributions are theoretical in nature, their practical significance extends to any future SCOD method. The first contribution, characterizing the structure of the optimal strategy, effectively narrows the pool of predictors suitable for the SCOD problem. The second contribution establishes that attempts to devise efficient learning algorithms for SCOD, without assumptions on data distribution and relying solely on ID data, are futile. Notably, many existing OODD and SCOD methods, by making no explicit distribution assumptions and using only the ID sample, are unable to be PAC learners. The proposed algorithm \Algo{} serves as a prime example of methods that are in line with these guidelines, and our empirical findings confirm its superior performance compared to existing approaches.

Proofs of all theorems can be found in the Appendix.
\section{The SCOD problem and its optimal solution}
\label{sec:model}

%The terminology of ID and OOD samples comes from the setups when the training set contains only ID samples, while the test set contains a mixture of ID and OOD samples. 

SCOD is a decision-making problem that aims to design a selective classifier applied to samples from a mixture of ID and OOD. The selective classifier comprises two functions: a classifier of ID data and a selective function (or a {\em selector} for short). The selector determines which samples are accepted for prediction by the ID classifier and which are rejected as OOD samples or ID samples likely to be misclassified by the ID classifier. In this section, we first define the SCOD problem and the concept of optimal strategy following~\cite{Xia-SIRC-ACCV2022}. Then, we present our first contribution, which shows that {\em optimal SCOD strategies involve the Bayes classifier of ID data and a selector being a stochastic linear classifier in a 2D space, whose input features are the conditional risk of the ID classifier and the OOD/ID likelihood ratio}. Our result shows that existing OOD detection methods and the state-of-the-art method for SCOD, the SIRC~\cite{Xia-SIRC-ACCV2022}, return suboptimal strategies for the SCOD problem.

\subsection{Definitions}
\label{ssec:definitions}
\boldparagraph{Data distribution} 
%As in the previous work, we assume that at the deployment the samples are generated from a mixture of two distributions, the ID and the OOD~\cite{Zhen-NIPS2022}.
Let $\SX$ be a set of observable inputs, and $\SY$ a finite set of labels that can be assigned to ID data. ID samples $(x,y)\in\SX\times\SY$ are generated i.i.d. from a joint distribution $p_I\colon\SX\times\SY\rightarrow\Re_+$. OOD samples $x\in\SX$ are generated from a distribution $p_O\colon\SX\rightarrow\Re_+$. ID and OOD samples share the same input space $\SX$. Let $\emptyset$ be a special label to mark the OOD sample, and  $\bar{\SY}=\SY\cup \{\emptyset\}$ an extended label set. At the {\em deployment stage}, the samples $(x,\bar{y})\in\SX\times\bar{\SY}$ are generated from the joint distribution $p\colon\SX\times\bar{\SY}\rightarrow\Re_+$ defined as a mixture of ID and OOD~\cite{Zhen-IsOODLearnable-NIPS2022}:
\begin{equation}
  \label{equ:dataDistr}
   p(x,\bar{y}) = \left \{\begin{array}{rcl}
      p_O(x)\,\Pout &\quad\mbox{if} &\quad\bar{y}=\emptyset \,,\\
      p_I(x,\bar{y})\,(1-\Pout) &\quad \mbox{if}&\quad\bar{y}\in\SY\,,
   \end{array}
   \right .
\end{equation}
where $\Pout\in[0,1]$ is the portion of OOD data in the mixture.
%probability of observing the OOD sample. 
%Our OOD setup subsumes the standard non-OOD setup as a special case when $\pi=0$, and the reject option models that will be introduced below will become for $\pi=0$ the known reject option models for the non-OOD setup.
%Therefore, our results are valid even if the OOD samples were represented in the training set, e.g., OOD samples could correspond to a background or a garbage class.

\boldparagraph{Selective classifier} The ultimate goal is to design a reject option strategy $q\colon\SX\rightarrow\SD$, where $\SD=\SY\cup\{\reject\}$ is the decision set, which either predicts a label, $q(x)\in\SY$, or rejects the prediction, $q(x)=\reject$. Following~\cite{Geifman-SelectClass-NIPS2017}, we 
represent the reject option strategy $q$ by a selective classifier $(h,c)$ that comprises the ID classifier $h\colon\SX\rightarrow\SY$, and a stochastic selector $c\colon\SX\rightarrow[0,1]$ which outputs a probability that the input is accepted, i.e.,
\begin{equation}
\label{equ:selClassif}
   q(x)=(h,c)(x) =\left \{
    \begin{array}{rcl}
       h(x) & \quad\mbox{with probability} &\quad c(x) \,,\\
       \reject & \quad\mbox{with probability}&\quad 1-c(x) \,.
    \end{array}
    \right .
\end{equation}
We use the stochastic selector because it turns out to be an optimal solution in the general setting; however, we will show that in most practical settings, the deterministic strategy $c\colon\SX\rightarrow\{0,1\}$ suffices. 

\boldparagraph{Evaluation metrics} We define three base metrics to evaluate the performance of the SCOD strategy $(h,c)$. One role of the selector $c\colon\SX\rightarrow[0,1]$ is to discriminate ID/OOD samples. We consider ID and OOD samples as positive and negative classes, respectively. We evaluate the performance of the selector by the True Positive Rate (TPR) and the False Positive Rate (FPR). The TPR/FPR is the probability that the ID/OOD sample is accepted by the selector $c$, i.e.,
\begin{equation}
  %\label{equ:TPR}
  \tpr(c) = \EE_{x\sim p_I(x)} c(x)\quad \mbox{and}\quad\fpr(c) = \EE_{x\sim p_O(x)} c(x) \:.
  %\tpr(c) = \int_{\SX} p(x\mid \bar{y}\neq\emptyset )\, c(x)\, dx = \int_{\SX} p_I(x)\, c(x) \, dx \:.
\end{equation}
%The FPR is the probability that OOD sample is accepted by the selector $c$, i.e.,
%\begin{equation}
%  \label{equ:FPR}
%  \fpr(x) = \EE_{x\sim p_O(x)} c(x) 
%  \fpr(c) = \int_{\SX} p(x\mid \bar{y}=\emptyset)\, c(x)\, dx = \int_{\SX} p_O(x) \, c(x) \, dx \:.
%\end{equation}
%
%The second identity in~\equ{equ:TPR} and~\equ{equ:FPR} is obtained after substituting the definition of $p(x,\bar{y})$ from~\equ{equ:dataDistr}. 
The performance of the ID classifier $h\colon\SX\rightarrow\SY$ on the accepted samples w.r.t. user-defined loss $\ell\colon\SY\times\SY\rightarrow\Re_+$ is characterized by the selective risk~\cite{Geifman-SelectClass-NIPS2017} 
\[
%   \Rsel(h,c) = \frac{\int_{\SX}\sum_{y\in\SY} p_I(x,y)\,\ell(h(x),y)\,c(x)\, \,dx}{\tpr(c)} \;,
   \Rsel(h,c) = \frac{\EE_{(x,y)\sim p_I(x,y)} [\ell(y,h(x))\,c(x)]}{\tpr(c)} \;,
\]
which is defined for non-zero $\tpr(c)$. 
%Note that the selective risk $R_S(h,c)$ depends both on the ID classifier $h$ and the selector $c$.

\begin{definition}[SCOD problem]
\label{def:SCOD}
Let $\tprMin\in(0,1)$ be a user-defined minimum acceptable TPR and $\relCost\in[0,1]$ a relative cost associated with not rejecting an OOD sample. The SCOD problem involves solving
\begin{equation}
 \label{equ:SingeCostModel}
       \min_{\genfrac{}{}{0pt}{}{h\in\SY^\SX}{c\in[0,1]^\SX}} \big [ (1-\relCost)\,\Rsel(h,c) + \relCost\, \fpr(c)\big ]
\qquad\mbox{s.t.}\qquad \tpr(c) \geq \tprMin \:,
\end{equation}
where we assume that both minimizers exist. A selective classifier $(h^*,c^*)$ that solves~\equ{equ:SingeCostModel} is called an optimal SCOD strategy. 
We refer to $R(h,c)=(1-\relCost)\,\Rsel(h,c) + \relCost\, \fpr(c)$ as the SCOD risk. 
%We say that the problem is feasible if there exists a selective function $c$ such that $\tpr(c)\geq \phi_{\rm min}$ and $\fpr(c)\leq \rho_{\rm max}$.
\end{definition}
Def.~\ref{def:SCOD} is a slight generalization of the formulation proposed in~\cite{Xia-SIRC-ACCV2022}, which assumes only the 0/1 loss $\ell(y,y')=\leftbb y\neq y'\rightbb$. The analysis and methods in this paper apply to any loss $\ell\colon\SY\times\SY\rightarrow\Re_+$ such that $\ell(y,y')=0$ iff $y=y'$. 

The formulation of the SCOD problem~\equ{equ:SingeCostModel} is straightforward, intuitive, and offers several advantages over alternative formulations. For example, one could substitute TPR in the constraint with the total coverage $\rho(c)= \tpr(c)(1-\pi_O)+\fpr(c)\pi_O$. A notable advantage of the SCOD formulation~\equ{equ:SingeCostModel} is its independence on the portion of OOD data $\pi_O$ that is unknown and often non-stationary in practice. Additional benefits over alternatives are discussed in~\cref{asec:alternative_formulations}.

\subsection{The optimal SCOD strategy}
\label{sec:optScodStrategy}

In this section, we present our main result, which shows how to construct an optimal SCOD strategy.

\begin{theorem}\label{thm:BayesClsIsOptimal}
    Let $(h^*,c^*)$ be an optimal solution to~\equ{equ:SingeCostModel}. Then $(h_B,c^*)$, where $h_B$ is the Bayes ID classifier
    \begin{equation}
       \label{equ:BayesCls}
       h_B(x) \in \Argmin_{y'\in\SY}\sum_{y\in\SY}p_I(y\mid x)\ell(y,y')\,,
    \end{equation}
    is also optimal to~\equ{equ:SingeCostModel}.
\end{theorem}
\Cref{thm:BayesClsIsOptimal} ensures that the Bayes ID classifier $h_B$ is an optimal solution to~\equ{equ:SingeCostModel}. 
%Note that $h_B$ is not a unique solution as altering the predictions for non-accpeted inputs, $c(x)=0$, do not change the optimum. 
After approximating $h^*$, e.g., by our best estimate of $h_B$ learned from the data, the search for an optimal selector $c$ leads to the following problem:

\begin{problem}{\bf (Optimal SCOD selector $c$ for known ID classifier $h$)}
\label{prob:SingleCostModelFixedCls} Given ID classifier $h\colon\SX\rightarrow\SY$, and the user-defined parameters $\tprMin\in(0,1)$ and $\relCost\in[0,1]$, the optimal selector $c^*\colon\SX\rightarrow[0,1]$ is a solution to 
\begin{equation}
  \label{equ:JointRiskProblemFixedH}
   \min_{c\in[0,1]^\SX} \big [(1-\relCost)\,\Rsel(h,c)+\relCost\, \fpr(c)\big ] \qquad\mbox{s.t.}\qquad \tpr(c) \geq \tprMin\:.
\end{equation}
\end{problem}

\begin{theorem}\label{thm:SingleCostModel} 
Let $h\colon\SX\rightarrow\SY$ be any ID classifier and $r\colon\SX\rightarrow\Re$ its conditional risk $r(x)=\sum_{y\in\SY}{p_I(y\mid x)}\ell(y,h(x))$. Let $g(x)=p_O(x)/p_I(x)$ be the likelihood ratio of the OOD and ID samples. Then, the set of optimal solutions of Problem~\ref{prob:SingleCostModelFixedCls} contains the selector
\begin{equation}
  \label{equ:optimalScodSelector}
   c^*(x)=\left \{ \begin{array}{rcl}
     0 & \;\;\mbox{if} & \;\;s(x) > \lambda \\
     \tau & \;\;\mbox{if} &\;\; s(x) = \lambda \\
     1 & \;\;\mbox{if} & \;\;s(x) < \lambda 
   \end{array}\right . 
%\end{equation}
\quad\mbox{with the score}\quad
%\begin{equation}
  %\label{equ:optScore}
   s(x) = r(x) + \frac{\relCost \,\tpr_{\rm min}}{1-\relCost}\, g(x)
\end{equation}
for $\alpha\in[0,1)$ and $s(x)=g(x)$ for $\alpha=1$. The decision threshold $\lambda\in\Re$ and the randomization parameter $\tau\in [0,1]$ are implicitly defined by the distribution $\left(p_O\left(x\right), p_I\left(x,y\right)\right)$ and the problem parameters $(\ell,\tpr_{\rm min}\,, \relCost)$.
\end{theorem}
\Cref{thm:BayesClsIsOptimal,thm:SingleCostModel}  show that an optimal SCOD strategy $(h^*,c^*)$ can be constructed from the Bayes ID classifier $h^*=h_B$ and a linear stochastic classifier~\equ{equ:optimalScodSelector} operating in 2D features space, where the first coordinate is the conditional risk $r(x)=\sum_{y\in\SY}p_I(y\mid x)\ell(y,h^*(x))$ and the second coordinate is the OOD/ID likelihood ratio $g(x)=p_O(x)/p_I(x)$. The slope of the linear classifier is determined by the user-defined parameters $(\alpha,\tprMin)$. In general, the selector has to randomize with probability $\tau$ for boundary inputs $\SX_{s(x)=\lambda}=\{x\in\SX\mid s(x)=\lambda\}$. However, if $\SX$ is continuous, the set $\SX_{s(x)=\lambda}$ has in most cases the probability measure zero, and $\tau$ can be arbitrary, i.e., the deterministic $c^*(x)=\leftbb s(x) \leq \lambda\rightbb$ is optimal. Note that Thm.~\ref{thm:SingleCostModel} shows how to construct an {\em optimal selector for an arbitrary ID classifier} $h$, i.e., not only for the Bayes classifier~$h_B$.

\begin{figure}
  \centering  \includegraphics{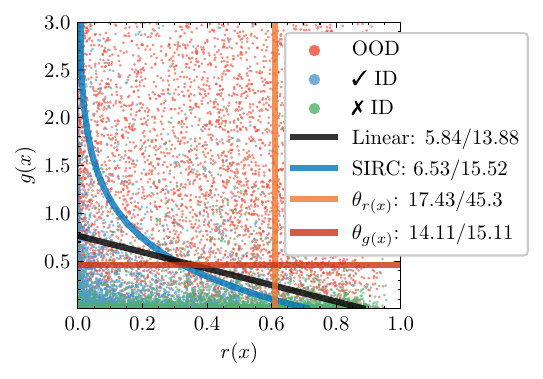}
  %\caption{The figure depicts four distinct selectors functioning as binary classifiers in 2D space. The selectors utilize two scores: i) a plugin conditional risk $\hat{r}(x)$, derived from the softmax posterior of a CNN trained on ImageNet (i.e., $\hat{r}(x)=1-$MSP score), and ii) a plugin estimate of the OOD/ID likelihood ratio, computed from the output of a CNN trained with standard Binary Cross-Entropy (BCE) to distinguish between ID data (ImageNet) and OOD data (SSB\_Hard). Test OOD data points are denoted in red, while correctly \cmark, and incorrectly \xmark, classified ID data points are denoted in blue/green. In addition to the proved optimal linear strategy~\equ{equ:optScore} and the SIRC~\equ{equ:SIRC}, we present two single-score strategies using either $\hat{r}(x)$ or $\hat{g}(x)$. All selector parameters are fine-tuned on test data to achieve a TPR of $90\%$ and minimal SCOD risk with $\alpha=0.5$. For each selector, we provide two metrics: $\areaSCODTpr\downarrow$ / ScodRisk at TPR=$90\%\downarrow$, see \cref{ssec:evaluation_metrics}.
  \caption{Selectors in 2D space using scores: $\hat r(x)$, a softmax score from an ImageNet classifier, and $\hat g(x)$, learned OOD/ID likelihood ratio. OOD data is shown in red, correctly \cmark, and incorrectly \xmark, classified ID data is shown in blue/green. Selector parameters tuned for $90\%$ TPR and minimal SCOD risk ($\alpha=0.5$). Metrics: $\areaSCODTpr\downarrow$ / ScodRisk at TPR=$90\%\downarrow$, details in \cref{ssec:evaluation_metrics}.}
  % Figure shows four different selectors as binary classifiers in 2D space whose coordinates are the two scores the selectors use: i) a plugin conditional risk $\hat{r}(x)$ computed from the soft-max posterior of a CNN trained on ImageNet (i.e. $\hat{r}(x)$=1-MSP score) and ii) a plugin estimate of OOD/ID likelihood ratio computed from the output of CNN trained with the standard BCE to separate ID data (ImageNet) and OOD data (SSB\_Hard). The red points are test OOD data and the blue/green points are correctly \cmark / incorrectly \xmark\, classified ID data. Besides the linear strategy~\equ{equ:optScore} (proved to be optimal) and the SIRC~\equ{equ:SIRC}, we show two single-score strategies using either $\hat{r}(x)$ or $\hat{g}(x)$. Parameters of all selectors are tuned on the test data to achieve TPR=$90\%$ and minimal SCOD risk with $\alpha=0.5$. For each selector, we show two metrics: $\areaSCODTpr$ / ScodRisk$@$Tpr=$90\%$, see \cref{ssec:evaluation_metrics}.
  %}
  \label{fig:ExampleSelectors}
\end{figure}

\subsection{Relation to existing OODD and SCOD strategies}
\label{ssec:existing_strategies}
\boldparagraph{Single-score strategy} Previous work on OODD focuses only on designing a good ID/OOD discriminator while ignoring the performance of the ID classifier on the accepted ID data. OODD methods output a single score $s\colon\SX\rightarrow\Re$ that is used to build a selector $c(x)=\leftbb s(x)\leq \lambda \rightbb$.
%where $\lambda\in\Re$ is a decision threshold chosen in the post-hoc evaluation. 
The goal is defined implicitly or explicitly as the Nayman-Pearson problem~\cite{Neyman-1928}, and the performance of the selector $c$ is usually evaluated by the ROC curve. Examples of single-score OOD methods involve the MSP score~\cite{Hendrycks-baseline-ICLR17}, ViM~\cite{Wang-ViM-CVPR2022}, GradNorm~\cite{Neal-GradNorm-ECCV2018}, etc. Our result shows that all existing single-score methods are not optimal, provided that one wants to solve the SCOD problem, which we verify experimentally in~\cref{sec:experiments}.
%In Sec~\ref{sec:experiments} we show experimentally that single-score strategies are consistently worse than the double-score strategies.
\boldparagraph{Double-score strategy} The SCOD problem we analyze in our paper was formulated in~\cite{Xia-SIRC-ACCV2022}, which also proposed the Softmax Information Retaining Combination (SIRC). SIRC is a heuristic strategy to combine two scores $s_1\colon\SX\rightarrow\Re$ and $s_2\colon\SX\rightarrow\Re$ into a single one:
\begin{equation}
  \label{equ:SIRC}
   s_{\rm SIRC}(x) = -(S_1^{max}-s_1(x))(1+\exp(-b(s_2(x)-a))
\end{equation}
where $S_1^{max}$ is an upper bound on $s_1(x)$, and $a$ and $b$ are hyper-parameters chosen based on a sample of ID data. The score $s_{\rm SIRC}$ is used to build a selector $c(x)=\leftbb s_{\rm SIRC}(x)\leq \lambda \rightbb$. The authors impose the following informal assumptions on the scores $s_1$ and $s_2$. The score $s_1$ should be i) higher for correctly classified ID samples and ii) lower for misclassified ID samples and OOD samples; in experiments, they set $s_1$ to the MSP score. The score $s_2$ should be lower for OOD data compared to ID data; in the experiments, $s_2$ score is either $L_1$-norm score~\cite{Huang-Importance-NIPS2021} or the negative of the Residual score~\cite{Wang-ViM-CVPR2022}. \cite{Xia-SIRC-ACCV2022} claim, and we confirm their findings in Sec.~\ref{sec:experiments}, that the SIRC strategy performs very well in practice. However, our result shows that the SIRC strategy is not optimal when the scores $s_1$ and $s_2$ approach their ideal setting, i.e. when $s_1(x)=r(x)$ and $s_2(x)=g(x)$. Then, the linear combination~\equ{equ:optimalScodSelector} performs better, which we verify experimentally in \cref{sec:experiments}.
\Cref{fig:ExampleSelectors} illustrates the proven optimal linear selector, SIRC, and two single-score selectors, functioning as binary classifiers in 2D space. The selectors are separating ImageNet samples as ID data and SSB\_Hard as OOD data.

\section{SCOD problem is not PAC learnable}
\label{sec:pac_learnability}

In the previous section, we showed how to construct an optimal strategy for the SCOD problem, provided $p_I(y\mid x)$ and $g(x)=p_O(x)/p_I(x)$ are known. The next key question is: Can the SCOD problem be 
%After introducing the SCOD problem and the optimal strategy in Def.~\ref{def:SCOD}, the key question arises: 
%Can this problem be 
effectively solved using available data? We build on the findings in~\cite{Zhen-IsOODLearnable-NIPS2022}, which establish the non-learnability of the OODD problem in a distribution-free setting~\footnote{Distribution-free setting implies learning guarantees for any data distribution.}, particularly when the learning algorithm relies {\em solely on an ID data sample}. Unlike the SCOD problem, the OODD problem is formulated as a cost-based minimization problem~\cite{Zhen-IsOODLearnable-NIPS2022}, which requires cost assignment for all decision outcomes in the prediction strategy. Extending these insights, we demonstrate that in the distribution-free setting, the SCOD problem is also non-learnable when only ID data is accessible. To achieve this, we broaden the Probably Approximately Correct (PAC) learnability concept~\cite{Shwartz-ML-2014} for the SCOD problem.

%After defining the SCOD problem and the concept of optimal strategy in Def.~\ref{def:SCOD}, the question is whether the problem can be solved from the data. We build on the work~\cite{Zhen-IsOODLearnable-NIPS2022} showing that the OODD problem is not learnable in the distribution-free setting~\footnote{The distribution-free setting means the learning is guaranteed for any data distribution.} when the learning algorithm uses only a sample of ID data. The OODD problem is defined as a cost-based minimization problem~\cite{Zhen-IsOODLearnable-NIPS2022}, which, unlike the SCOD problem, requires assigning a cost to all decisions of the prediction strategy. We extend the result of~\cite{Zhen-IsOODLearnable-NIPS2022} to the SCOD problem and show that in the distribution-free setting the SCOD problem is also not learnable when only ID data are available. To this end, we first extend the concept of the Probably Approximately Correct (PAC) learnability for the SCOD problem. 

\begin{definition}{\bf (PAC learnability of SCOD problem)}
 \label{def:PAC}
    Let $\ell\colon\SY\times\SY\rightarrow\Re_+$ be a loss function and $(\tprMin,\relCost)\in (0,1)^2$ user-defined parameters of the SCOD problem~\equ{equ:SingeCostModel}. A hypothesis class $\SH\subset \{(h,c)\in\SY^\SX \times{[0,1]^\SX}\}$~\footnote{We use shortcuts $\SY^\SX=\{h\colon\SX\rightarrow\SY\}$, $[0,1]^\SX=\{c\colon\SX\rightarrow[0,1]\}$.} is PAC learnable if there exist a function $m\colon (0,1)^3\rightarrow\NN$ and a learning algorithm $A\colon\cup_{m=1}^\infty (\SX\times\SY)^m\rightarrow\SH$ with the following property: For every $(\veps_1,\veps_2,\delta)\in(0,1)^3$, every OOD distribution $p_O\colon\SX\rightarrow\Re_+$, every ID distribution $p_I\colon\SX\times\SY\rightarrow\Re_+$ and every $\pi_O\in[0,1]$, when running the algorithm $A$ on $m\geq m(\veps_1,\veps_2,\delta)$ examples i.i.d. drawn from $p_I(x,y)$, the algorithm returns a selective classifier $(h_m,c_m)$ such that with a probability of at least $1-\delta$ it holds that
    \[
       R(h_m,c_m)-R(h^*,c^*)\leq \veps_1\quad\mbox{and}\quad \tpr(c_m)\geq \tprMin-\veps_2\:,
    \]
    where $R(h,c)=(1-\alpha)\Rsel(h,c)+\alpha\,\fpr(c)$ is the SCOD risk and $(h^*,c^*)$ is an optimal SCOD strategy.
\end{definition}
Def.~\ref{def:PAC} establishes PAC learnability, indicating the existence of an algorithm searching in a hypothesis space $\SH$, which can discover an $(\veps_1,\veps_2)$-optimal solution for the SCOD problem~\equ{equ:SingeCostModel} with an arbitrarily low probability of failure $\delta\in(0,1)$, given a sufficient number of ID data. Importantly, this guarantee is distribution-free, applying universally to every mixture~\equ{equ:dataDistr} of ID and OOD data.

%The PAC learnability in Def.~\ref{def:PAC} implies, that there exists an algorithm searching in a hypothesis space $\SH$ which is able to find an $(\veps_1,\veps_2)$-optimal solution of the SCOD problem~\equ{equ:SingeCostModel} with arbitrary low probability of failure $\delta\in(0,1)$, provided it uses sufficient amount of ID data. The guarantee is distribution-free, i.e., it must be valid for every mixture~\equ{equ:dataDistr} of ID and OOD data. 

\begin{theorem}
\label{thm:SCODisNotPAC}
    Let $\SH\subset \{(h,c)\in\SY^\SX \times[0,1]^\SX\}$ be a non-trivial hypothesis space such that there exist two selective classifiers $(h_1,c_1)\in\SH$ and $(h_2,c_2)\in\SH$ for which $c_1\neq c_2$. Then, the hypothesis space $\SH$ is not PAC learnable in the sense of Definition~\ref{def:PAC}.
\end{theorem}
The implication from Theorem~\ref{thm:SCODisNotPAC} is that, generally, achieving arbitrarily precise approximation of the optimal SCOD strategy using only ID data is unattainable unless the hypothesis space is trivial~\footnote{The trivial hypothesis space involves a single selector, reducing the SCOD problem to standard prediction under the closed-world assumption - known to be learnable when $\SH$ has finite complexity.}. Although our result is negative, it has an important implication: {\em Attempts to develop an efficient learning algorithm for a SCOD problem that does not make an assumption about the data distribution and uses only ID data are futile}.

%Motivated by this, in the next section, we introduce a learning approach that relaxes the assumptions of Theorem~\ref{thm:SCODisNotPAC}. Specifically, we assume: i) the availability of a sample of unlabeled mixtures of ID and OOD data, and ii) the existence of a nonlinear transform rendering ID and OOD data normally distributed. While lacking formal proof ensuring PAC learnability under these assumptions, we offer compelling empirical support for the proposed method.

%Theorem~\ref{thm:SCODisNotPAC} implies that in general there is no hope to learn a precise approximation of the optimal the optimal SCOD strategy using only the ID data, unless the hypothesis space is trivial~\footnote{The trivial hypothesis space involves a single selector, in which case the SCOD problem reduces to standard prediction under the closed-world assumption, which is well known to be learnable when $\SH$ has a finite complexity (e.g. VC dimension)}. Motivated by this result, in the following section, we propose a learning which weakens the assumption of Theorem~\ref{thm:SCODisNotPAC}, namely: we assume that i) a sample of unlabeled mixture of ID and OOD data is available and ii) there is a nonlinear transform which makes ID and OOD data to be normally distributed. We do not have a formal proof that would guarantee the PAC learnability under these two assumptions; however, we provide convincing empirical support for the proposed method.

\section{Plugin estimate of the optimal SCOD strategy}
\label{sec:plugin_estimate_of_optimal_scod}

In this section, we leverage the theoretical insights from preceding sections to introduce a method for learning a plugin estimate of the optimal SCOD strategy. We adopt the framework proposed by~\cite{Katz-Habitats-ICML2022}, where in addition to the ID data sample $\ST_{I}=((x^I_i,y^I_i)\in\SX\times\SY\mid i=1,\ldots,m)$ generated from $p_I(x,y)$ we also have an unlabeled sample of a mixture of ID and OOD data $\ST_{U}=(x^U_i\in\SX\mid i=1,\ldots,n)$ generated from $p(x)=\pi_O^{\rm tr}\, p_O(x) + (1-\pi_O^{\rm tr}) p_I(x)$. Collecting the data for the unlabeled mixture can, e.g., involve just recording the input samples from a real deployment of the predictor $h$. We use data $\ST_I$ and $\ST_U$ to learn a plugin estimate of the optimal SCOD strategy derived in Sec~\ref{sec:optScodStrategy}.

%We assume the setting proposed by~\cite{Katz-Habitats-ICML2022}, when we have a sample of ID data $\ST_{I}=((x^I_i,y^I_i)\in\SX\times\SY\mid i=1,\ldots,m)$ generated i.i.d. from $p_I(x,y)$ and a sample of an unlabeled mixture of ID and OOD data $\ST_{M}=(x^M_i\in\SX\mid i=1,\ldots,n)$ generated i.i.d. from $p(x)=\pi_O p_O(x) + (1-\pi_O) p_I(x)$. 

\boldparagraph{ID classifier} We can use any method to train the ID classifier $h$ from $\ST_I$ that provides an estimate of the class posterior $p_I(y\mid x)$. In our experiments, we use a CNN with softmax decision layer trained by cross-entropy loss producing $\hat{p}_I(y\mid x)$. Then, we construct the plug-in Bayes ID classifier and conditional risk:
\begin{equation}
  \label{equ:pluginBayes}
  \hat{h}_{B}(x) \in \Argmin_{y'\in\SY}\sum_{y\in\SY} \hat{p}_I(y\mid x)\ell(y,y')
%\end{equation}
%and the plug-in conditional risk
%\begin{equation}
%\label{equ:pluginCondRisk}
\;\;\mbox{and}\;\;\hat{r}(x)=\sum_{y\in\SY}\hat{p}_I(y\mid x) \ell(y, \hat{h}_B(x)) \:.
\end{equation}
Note that in case of $0/1$-loss, $\ell(y,y')=\leftbb y\neq y'\rightbb$, $\hat{h}_B(x)=\Argmax_{y\in\SY}\hat{p}_I(y\mid x)$ is the standard MAP rule, and $\hat{r}(x)=1-\max_{y\in\SY}\hat{p}_I(y\mid x)$ is the 1-MSP rule~\cite{Hendrycks-baseline-ICLR17}.

\boldparagraph{Selector} Once we have $\hat{r}$, it remains to estimate $g(x)=p_O(x)/p_I(x)$, in order to build the plugin estimator of the optimal selector $c^*(x)$ given by~\equ{equ:optimalScodSelector}. We create a sequence $\ST_{IU}=((x_i,z_i)\in\SX\times\{I,U\}\mid i=1,\ldots,n+m)$ by randomly re-shuffling a concatenation of $\ST_I$ and $\ST_U$, and setting $z_i=I$ when $x_i$ is from $\ST_I$ and $z_i=U$ when $x_i$ is from $\ST_U$. The sequence $\ST_{IU}$ can be seen as a random sample from a mixture 
\begin{equation}
  \label{equ:IDvsUnlabMixture}
  p(x,z) = \left \{ \begin{array}{rcl}
       \pi_U\, (p_O(x)\;\pi_O^{\rm tr} + p_I(x)\, (1-\pi_O^{\rm tr})) & \quad\mbox{if}\quad { }& z = U\\
       (1-\pi_U)\, p_I(x) & \quad\mbox{if}\quad { }& z=I
  \end{array}
  \right .
\end{equation}
where $\pi_U=n/(m+n)$ is the {\em known portion} of $\ST_U$ in $\ST_{IU}$. It follows directly from~\equ{equ:IDvsUnlabMixture} that the desired OOD/ID likelihood ratio reads
\begin{equation}
 \label{equ:gCorrection}
    g(x)=\frac{p_O(x)}{p_I(x)} = \frac{p(z=U\mid x)}{p(z=I\mid x)} \frac{1-\pi_U}{\pi_U\pi_O^{\rm tr}} - \frac{1-\pi_O^{\rm tr}}{\pi_O^{\rm tr}} \:.
\end{equation}
We propose to approximate the unknown $p(z\mid x)$ by a {\em corrected sigmoid model} (CSM) $p(z\mid x; \#\theta, a)$, the use of which is motivated by Theorem~\ref{thm:CorrectedSigmoid}. 

\begin{theorem}\label{thm:CorrectedSigmoid}
Let $\#\phi\colon\SX\rightarrow\Re^d$ be a feature map. Assume that the features $\#\phi(x)$ computed on the ID and OOD data are normally distributed, i.e, $p_I(x;\#\mu_I,\#C) = \SN(\#\phi(x);\#\mu_I,\#C)$ and $p_O(x;\#\mu_O,\#C) = \SN(\#\phi(x);\#\mu_O,\#C)$. Then, the posterior $p(z\mid x)$ derived from the distribution~\equ{equ:IDvsUnlabMixture} is an element of $\SP=\{ p(z\mid x;\#\theta,a) \mid a = \pi_U(1-\pi_O^{\rm tr})/(1-\pi_U), \#\theta\in\Re^{d+1}\}$ where 
\begin{equation}
  \label{equ:correctedSigmoid}
    p(z=I\mid x;\#\theta,a) = \frac{1}{1+|a|+\exp(\#\theta^T[\#\phi(x);1])} \:,
\end{equation}
is the corrected sigmoid.
\end{theorem}
We estimate the parameters $(\#\theta,a)$ of CSM by the Maximum Likelihood (ML) method, which corresponds to minimizing the binary cross-entropy (BCE) of the proposed CSM~\equ{equ:correctedSigmoid} on the sequence $\ST_{IU}$. Let $(\hat{\theta},\hat{a})$ be the parameters estimated from $\ST_{IU}$. By Theorem~\ref{thm:CorrectedSigmoid}, the unknown $\pi_O^{\rm tr}$ can be recovered from the parameter $\hat{a}$ by $\hat{\pi}_O^{\rm tr}=1+|\hat{a}| -|\hat{a}|/\pi_U$. Finally, we substitute $p(z\mid x; \hat{\#\theta},\hat{a})$ and $\hat{\pi}_O^{\rm tr}$ into formula~\equ{equ:gCorrection} to obtain an estimate of the OOD/ID likelihood ratio $\hat{g}$, and use it to obtain a plugin estimate of the optimal score~\equ{equ:optimalScodSelector}, that is, 
\begin{equation}
  \label{equ:selectorScore}
   \hat{s}(x) = \hat{r}(x) + \frac{\alpha\,\tprMin}{1-\alpha}\hat{g}(x)    
%\end{equation}
\;\;\;\;\mbox{where}\;\;\;\;
%\begin{equation}
% \label{equ:pluginLratio}
   \hat{g}(x)=\frac{p(z=U\mid x;\hat{\#\theta},\hat{a})}{p(z=I\mid x;\hat{\#\theta},\hat{a})} \frac{1-\pi_U}{\pi_U\hat{\pi}_O^{\rm tr}}\:.    
\end{equation}
In the formula for $\hat{g}$ we omit the additive term present in~\equ{equ:gCorrection}, as it is absorbed by the decision threshold $\lambda$ of the linear selector~\equ{equ:optimalScodSelector}. The value of $\lambda$ is adjusted on $\ST_I$ to achieve the target $\tpr(c)=\tprMin$. Algo.~\ref{algo} summarizes the proposed method to learn the Plugin estimate of the Optimal SCOD strategy (\Algo{}).

\begin{algorithm}
 \caption{\Algo{}}
  \begin{algorithmic}[1]
      \Require ID data $\ST_I=((x_i^I,y_i^I)\in\SX\times\SY\mid i=1,\ldots,m)$, unlabeled mixture of ID and OOD $\ST_U=(x_i^U\in\SX\mid i=1,\ldots,n)$, problem parameters $(\alpha,\tprMin)\in (0,1)^2$, target loss $\ell\colon\SY\times\SY\rightarrow\Re_+$.
      \Ensure Selective classifier $(\hat{h},\hat{c})$ for the SCOD problem, Def.~\ref{def:SCOD}.
      \State Train $\hat{p}_I(y\mid x)$ on $\ST_I$ using the cross-entropy loss.
      \State Construct the plugin Bayes ID classifier $\hat{h}$ and its conditional risk $\hat{r}$ by~\equ{equ:pluginBayes}.
      %
      %\State Construct the plugin conditional risk $\hat{r}$ by~\equ{equ:pluginCondRisk}.
      %
      \State Create $\ST_{IU}=((x_i,z_i)\in\SX\times\{I,U\}\mid i=1,\ldots,n+m)$ by randomly re-shuffling inputs from $\ST_I$ and $\ST_U$. 
      \State Train $\hat{\#\theta}$ and $\hat{a}$ by minimizing the BCE of the corrected sigmoid~\equ{equ:correctedSigmoid} on $\ST_{IU}$. 
      \State Compute $\pi_U=n/(n+m)$ and $\hat{\pi}_O^{\rm tr}=1+|\hat{a}|-\frac{|\hat{a}|}{\pi_U}$.
      \State Construct the selector score $\hat{s}$ by~\equ{equ:selectorScore}.
      \State Tune $\lambda$ of $\hat{c}(x)=\leftbb \hat{s}(x) \leq \lambda\rightbb$ on $\ST_I$ to achieve the target $\tpr(\hat{c})=\tprMin$.
  \end{algorithmic}
  \label{algo}
\end{algorithm}

\boldparagraph{Computational requirements and implementation}  \Algo{} converts the learning process for the SCOD strategy into two steps. First, it involves training the ID classifier by minimizing the standard cross-entropy loss. Second, it includes learning a binary classifier by minimizing the BCE of the corrected sigmoid. Implementing this in Pytorch requires modifying two lines of code with the BCE of the standard sigmoid. Notably, \Algo{} does not use hyperparameters. 

\boldparagraph{Assumptions} \Algo{} relies on two assumptions. Firstly, the CSM in Eq.~\equ{equ:correctedSigmoid} should effectively represent $p(z\mid x)$. Secondly, the OOD prior $\pi_O^{\rm tr}$ needs to be learnable. Note that ID and {\em OOD are not required to be transformable to normal distributions}. However, when they are, Thm.~\ref{thm:CorrectedSigmoid} guarantees that CSM is exact. Notably, the unknown {\em OOD prior $\pi_O^{\rm tr}$ in the training sample $\ST_U$ does not need to match the OOD prior $\pi_O$ in the test sample} because the SCOD problem~\equ{equ:SingeCostModel} is independent of $\pi_O$; the estimate of $\pi_O^{\rm tr}$ is used only to adjust the scale of the learned likelihood ratio through~\equ{equ:gCorrection}. 
To ensure the learnability of $\pi_O^{\rm tr}$, the OOD must be a proper novelty distribution with respect to ID, as defined in Def. 4 and Prop. 5 in \cite{Blanchard-SemiSuperNovDet-JMLR2010}. OOD is considered proper if it cannot be decomposed into a mixture of ID and any other distribution on $\SX$. This assumption is practical and is typically satisfied in real-world scenarios. The suitability of CSM as a proxy for $p(z\mid x)$ should be experimentally validated for specific datasets, a validation we demonstrate in various benchmarks.

\boldparagraph{Relation to existing work} In prior research~\cite{Katz-Habitats-ICML2022,Narasimhan-Plugin-Arxiv2023}, the standard sigmoid was used to model $p(z\mid x)$. That is, they optimize $\#\theta$ to fit $p(z\mid x;\#\theta,a)$ to $\ST_{IU}$ while keeping $a$ fixed at 0. Using the standard sigmoid has two drawbacks: i) the model is incorrect because $a = \pi_U(1-\pi_O^{\rm tr})/(1-\pi_U)=0$ only if $\ST_U$ contains clean OOD data, i.e., $\pi_O^{\rm tr}=1$; % ii) it does not provide an estimate of the unknown~$\pi_O$.
ii) it does not estimate~$\pi_O^{\rm tr}$ (needed to compute $\hat g(x)$).

%The method is based on the identity [Blanchard]
%\[
%     \frac{p_O(x)}{p_I(x)=\frac{}}
%\]

\section{Relation to existing literature}

%OOD detection (OODD) has garnered significant attention in prior work, e.g.~\cite{Hendrycks-baseline-ICLR17,liang2018enhancing,Dhamija-NIPS2018,Devries-arxiv2018,Malinin-NEURIPS2018,Granese-Nips2021,Chen-PAMI2022,Sun-NIPS2021,Song-Neurips2022,pmlr-v162-sun22d,9879414}, with a focus on distinguishing ID and OOD samples. Although OOD detectors are ultimately employed as selectors alongside ID classifiers, the explicit consideration of misclassified ID samples when designing the selector has not been previously addressed. 

Recent insights emphasize the necessity, when designing selectors for ID classifiers in OOD settings, to reject not only OOD samples but also ID samples prone to misclassification. This problem has been termed Unknown Detection~\cite{Kim-UnifBench-ESA2021}, Unified Open-Set Recognition~\cite{Cen-Devil-ICLR2023}, and Selective Classification in the presence of Out-of-Distribution (SCOD) data~\cite{Xia-SIRC-ACCV2022}. ~\cite{Xia-SIRC-ACCV2022} formally defined the SCOD problem and introduced Softmax Information Retaining Combination (SIRC), a method tailored for SCOD. We analyze the SCOD problem and demonstrate that both existing OODD methods and SIRC deviate from the optimal SCOD strategy that we derived in our paper. We introduce a novel method, \Algo{}, learning a plugin estimate of the optimal SCOD strategy and empirically show that it outperforms both SIRC and existing OODD methods.

The PAC learnability of OODD was examined in~\cite{Zhen-IsOODLearnable-NIPS2022}, defining the optimal OODD strategy as an unconstrained cost-based minimization problem. Their results demonstrate that OODD is not PAC-learnable in the distribution-free setting when the learning algorithm relies solely on an ID data sample. We extend their findings and establish that, in the distribution-free setting, SCOD is also not PAC-learnable when only ID data is available.

The work of~\cite{Katz-Habitats-ICML2022} introduced a method to learn an OODD from an unlabeled mixture of ID and OOD data. Their formulation of the optimal strategy involves constrained optimization, different from the SCOD problem analyzed in our paper. Thanks to the known form of the optimal strategy derived in our paper, our proposed method, \Algo{}, simplifies the learning process by i) training an ID classifier using standard cross-entropy loss and ii) training a classifier via the BCE of the novel CSM. This simplified approach contrasts with the Augmented Lagrangian Method employed by~\cite{Katz-Habitats-ICML2022} to optimize the constrained problem.

Our work aligns with~\cite{Narasimhan-Plugin-Arxiv2023}, establishing an optimal strategy for a different SCOD formulation and proposing a plugin estimator. Key distinctions include i) our TPR-constrained SCOD formulation removing depenence on the unknown OOD portion in the test sample (c.f.~\cref{asec:alternative_formulations}),  ii) a novel technique for finding the optimal strategy applicable to all distributions, iii) providing an explicit formula to compute parameters of the optimal strategy and iv) using the CSM instead of the standard sigmoid to model the OOD/ID likelihood ratio. 

\vspace{-0.15cm}
\section{Experiments}
\label{sec:experiments}
In this section, we empirically validate the theoretical results presented in~\cref{thm:SingleCostModel}, confirming that the optimal SCOD selector indeed constitutes a linear combination of $g(x)$ and $r(x)$. Additionally, we show that our proposed method, \Algo{}, built on the plugin estimate of the optimal SCOD strategy, see \cref{sec:plugin_estimate_of_optimal_scod}, surpasses the current state-of-the-art on real-world datasets.

%In this section, we empirically validate~\cref{thm:SingleCostModel}, which states that the optimal SCOD selector is a linear combination of $g(x)$ and $r(x)$, and we demonstrate that our proposed method, \Algo{}, based on learning the plugin estimate of the optimal SCOD strategy (\cref{sec:plugin_estimate_of_optimal_scod}), outperforms the current state-of-the-art on real-world data.

\subsection{Evaluation metrics}
\label{ssec:evaluation_metrics}
Let $\ST = \ST_I\cup \ST_O$ be a sample of evaluation data where $\ST_{I}=((x^I_i,y^I_i)\in\SX\times\SY\mid i=1,\ldots,m)$ is a sample of i.i.d. ID from $p_I(x,y)$ and $\ST_{O}=((x^O_i, \emptyset)\in\SX\times\bar{\SY}\mid i=1,\ldots,n)$ is a sample of i.i.d. OOD from $p_O(x)$. Let $s\colon\SX\rightarrow\Re$ be a score, $c(x;\lambda)=\leftbb s(x)\leq \lambda \rightbb$ a selector and $h\colon\SX\rightarrow\SY$ an ID classifier. To summarize the performance of a selective classifier $(h,c)$ on the evaluation data $\ST$, we employ the \textit{Area Under SCOD Risk - True positive rate} ($\areaSCODTpr{}\downarrow$) curve which evaluates the overall performance of the selective classifier on the SCOD problem, see Def. \ref{def:SCOD}. 
For any variable $x$, let $\hat{x}$ be its empirical estimate. %As we are only provided with a sample of data, we use the empirical estimates of the metrics, e.g., instead of TPR, we use $\smash{\esttpr(c) = \frac{1}{m}\sum_{(x,y)\in\ST_I}\,c(x)}$. Analogously, for the FPR and the selective risk. % we use $\smash{\estfpr(c) = \frac{1}{n}\sum_{(x,\emptyset)\in\ST_O}\,c(x)}\,$ and $\estRsel(h,c) = \nicefrac{(\frac{1}{m}\sum_{(x,y)\in\ST_I}\,\ell(h(x),y)\,c(x))}{\esttpr(c)}$. 
The SCOD risk (refer to~\cref{equ:SingeCostModel}) at a given TPR is estimated by ${\estRscod(\tprMin;h,c) = \min_{\lambda}\,(1-\relCost)\,\estRsel(h,c)+\relCost\, \estfpr(c)\;} \mbox{s.t.}\; \esttpr(c) \geq \tprMin$. The $\areaSCODTpr$ is then computed as the area under the curve given by the points 
$\big\{\big(\estRscod(\tprMin), \tprMin\big)\mid \tprMin\in(0,1]\big\}$. See~\cref{asec:imagenet,asec:cifar10,asec:cifar100} for more metrics.

%three metrics: i) the \textit{Area Under Risk - Coverage} ($\areaRiskCoverage$$\downarrow$) curve which measures the model's ability to discriminate between correctly classified (\cmark\:ID) and incorrectly classified (\xmark\:ID) data, ii) the \textit{Area Under Receiver Operating Characteristic} ($\areaROC\uparrow$) curve which evaluates how well the model distinguishes between ID and OOD data, and iii)

\subsection{Experimental setup}
\label{ssec:experiments_methodology}
\boldparagraph{Datasets}
We assess SCOD performance on datasets adopted from the OpenOOD benchmark~\cite{Yang-Nips2022}. Detailed description of the datasets is provided in~\cref{assec:data}.

%and  across three distinct ID datasets: CIFAR10, CIFAR100, and ImageNet-1K. As OOD datasets, we consider CIFAR100, CIFAR10, MNIST, Places365, SVHN, Textures, Tiny ImageNet, SSB\_hard, Ninco, iNaturalist, Textures, and OpenImage\_O.

\boldparagraph{Models}
As the ID classifier $h(x)$ on CIFAR-10/100, we use pre-trained ResNet-18 from the OpenOOD benchmark~\cite{Yang-Nips2022}. For $h(x)$ on ImageNet-1K, we use a pre-trained ResNet-50 model from Torchvision \cite{Torchvision-GitHub2016}. For details, see~\cref{assec:models}.

\boldparagraph{Learning the likelihood ratio}
To estimate the OOD/ID likelihood ratio $g(x)$ on real-world data, we train a classifier with BCE of the standard sigmoid (a.k.a logistic regression) and a \textit{corrected sigmoid} model (refer to~\cref{sec:plugin_estimate_of_optimal_scod}). We reuse the feature representation of the ID classifier $h(x)$ to learn the likelihood ratio. Up to $50\%$ of OOD samples are reserved for training, and evaluations are done exclusively on unseen samples. Refer to \cref{assec:training_likelihood_ratio} for more details.

\boldparagraph{Double-score parameter setting strategies} 
As mentioned in \cref{ssec:existing_strategies}, the current state-of-the-art employs a nonlinear combination $s_{\rm SIRC}(x)$ of scores $s_1(x)$ and $s_2(x)$ with hyperparameters $a$ and $b$. In contrast, our proposed approach utilizes a linear combination $s_{\rm Linear}(x) = s_1(x) + \beta s_2(x)$ with a single hyperparameter $\beta$. Two key questions arise when comparing the approaches: Given the optimal parameters $a,b,\beta$, which model performs better? More importantly, which model performs better in a scenario when the optimal parameters are unknown? To answer the first question, we search for the best hyperparameters by optimizing the SCOD risk on the test set and refer to scores using the parameters as \textit{tuned}. The \textit{tuned} scores can not be used in practice, however, they provide an upper bound on the model's performance. For practical deployment of $s_{\rm SIRC}$, \cite{Xia-SIRC-ACCV2022} offers a heuristic to set the parameters using the empirical mean $\mu_{s_2}$ and standard deviation $\sigma_{s_2}$ of the score $s_2(x)$ on ID data. For the linear strategy, we derive an explicit formula, see \cref{equ:optimalScodSelector}. We refer to scores using these parameters as \textit{plugin} and they can be used in practice without test data tuning. We evaluate both \textit{tuned} and \textit{plugin} settings. Details on the tuning can be found in~\cref{assec:hyperparameters}.

\subsection{Experimental validation of theoretical results}
To validate Thm.~\ref{thm:SingleCostModel}, we approximate the likelihood ratio $g(x)$ on real-world data using a logistic regression model trained on features of the ID classifier $h(x)$. This model is impossible to obtain in practice, given the rarity of clean OOD data samples. Nonetheless, it provides the best available approximation of the true likelihood ratio $g(x)$. For the linear strategy, we employ the plug-in conditional risk $\hat{r}(x) = 1 - {\rm MSP}$ as the score $s_1(x)$ and use the approximated $\hat{g}(x)$ as the score $s_2(x)$. In the case of SIRC, we use $1-\hat{r}(x)$ and $-\hat{g}(x)$ as scores to meet the assumptions outlined in \cref{ssec:existing_strategies}. %We also assess a baseline double-score strategy $s(x) = s_1(x) \cdot s_2(x)$~\cite{Xia-SIRC-ACCV2022}. 
The performance of the selectors on ImageNet using these scores is summarized in~\cref{tab:imagenet_clean_likelihood}. 
%The linear strategy consistently outperforms SIRC and the baseline. Using the approximation of the optimal scores $\hat{r}(x)$ and $\hat{g}(x)$, the linear strategy also surpasses all other combinations of scoring functions of contemporary OOD detectors [cite:TODO] by a large margin in both the tuned and plugin setups across all ID datasets. For detailed results on all datasets, please refer to~\cref{asec:imagenet,asec:cifar10,asec:cifar100}.
%
The linear strategy consistently outperforms both SIRC and the baseline. %Using the approximations of the optimal scores, $\hat{r}(x)$ and $\hat{g}(x)$, 
The linear strategy with $\hat{r}(x)$ and $\hat{g}(x)$ also surpasses all other combinations of scoring functions employed by contemporary OOD detectors~\cite{Djurisic-ASH-ICLR2023,Weitang-EBO-NIPS2020,Huang-Importance-NIPS2021,Hendrycks-MLS-ICML2022,Hendrycks-baseline-ICLR17,liang2018enhancing,Sun-NIPS2021,Wang-ViM-CVPR2022}. This dominance holds true across all ID datasets, showcasing substantial advantages in both tuned and plugin setups. For detailed results on all datasets, refer to~\cref{asec:imagenet,asec:cifar10,asec:cifar100}.

\begin{table}[htbp]
\centering
%\caption{\areaSCODTpr{}$\downarrow$ in \% points for selective classifiers $(h,c)$ constructed from an ID classifier $h(x)$ and selectors $c(x)=\leftbb s(x)\leq \lambda \rightbb$. Results are shown for ID ImageNet and several possible OOD datasets. Rows of the Table correspond to different scores $s(x)$ and show the performance of both: i) tuned, and ii) plugin double-score methods. The relative cost is $\alpha = 0.5$. Rows marked in {\color{impractical} red} are not applicable in practice, as they use $\hat{g}(x)$ learned on clean OOD data.}
\caption{\areaSCODTpr{}$\downarrow$ in \% for selective classifiers $(h,c)$ from ID classifier $h(x)$ and selectors $c(x)=\leftbb s(x)\leq \lambda \rightbb$. Results on ID ImageNet and various OOD datasets. Rows represent different scores $s(x)$, showcasing performance for i) tuned, and ii) plugin double-score methods. {\color{impractical}Red} rows use $\hat{g}(x)$ learned on clean OOD data. Relative cost: $\alpha = 0.5$.}
%\resizebox{\columnwidth}{!}{%
\begin{tabular}{l|c|c|c|c|c}
\backslashbox{\it Score}{\it Dataset} & ssb\_hard & ninco & inaturalist & textures & openimage\_o\\ \hline
\color{impractical} Tuned SIRC             & 6.59 & \bf6.61 & 4.78 & \underline{7.45} & \underline{6.21} \\ \hline
\color{impractical} Tuned Linear           & \bf5.88 & \underline{6.67} & \bf4.04 & \bf4.23 & \bf5.82 \\
\hline \hline
\color{impractical} Plugin SIRC & $13.24$ & $10.94$ & $6.39$ & $10.49$ & $8.59$ \\ \hline
\color{impractical} Plugin Linear & \underline{6.10} & $8.05$ & \underline{4.35} & \underline{4.48} & $7.37$ \\% \hline
%\color{impractical} Multiplication   & 16.97 & 17.41 & 12.88 & 13.51 & 17.03 \\
\hline \hline
MSP~\cite{Hendrycks-baseline-ICLR17}             & 17.47 & 13.59 & 9.35 & 12.15 & 11.05 \\ \hline
\color{impractical} Likelihood Ratio & 14.14 & 14.04 & 11.22 & 11.75 & 14.00 \\
\end{tabular}
%}
\label{tab:imagenet_clean_likelihood}
\end{table}
\subsection{The \Algo{} algorithm}
\label{ssec:practical_application_of_optimal_plugin_scod_strategy}
To show the practicality of our results without requiring a clean OOD data sample, we employ a framework using an unlabeled mixture of ID and OOD data, as detailed in~\cref{sec:plugin_estimate_of_optimal_scod}. We demonstrate that the linear strategy combining $\hat{r}(x)$ and $\hat{g}(x)$ learned  by the proposed \Algo{}~\cref{algo} consistently outperforms SIRC and contemporary OOD scoring functions \cite{Djurisic-ASH-ICLR2023,Weitang-EBO-NIPS2020,Huang-Importance-NIPS2021,Hendrycks-MLS-ICML2022,Hendrycks-baseline-ICLR17,liang2018enhancing,Sun-NIPS2021,Wang-ViM-CVPR2022} across all ID datasets. Additionally, our results reveal that utilizing the standard sigmoid to approximate the likelihood ratio $g(x)$, as done in prior work~\cite{Katz-Habitats-ICML2022,Narasimhan-Plugin-Arxiv2023}, leads to suboptimal performance. Results for ID ImageNet using the scores $\hat{r}(x)$ and $\hat{g}(x)$ are shown in~\cref{tab:imagenet_ploscod}. In~\cref{asec:ablation}, we show that the results hold for a wide range of priors $\pi_O^{\rm tr}$ and relative costs $\alpha$. For comprehensive results across all datasets and compared scores, refer to~\cref{asec:imagenet,asec:cifar10,asec:cifar100}.

\begin{table}[htbp]
\centering
%\caption{Comparison of double-score strategies using logistic regression with standard and corrected sigmoid to approximate the likelihood ratio (LR) $g(x)$ from a mixture of data, see
%~\cref{sec:correctedSigmoid}. Table shows \areaSCODTpr{}$\downarrow$ in \% points for selective classifiers $(h,c)$ constructed from an ID classifier $h(x)$ and selectors $c(x)=\leftbb s(x)\leq \lambda \rightbb$. Results are shown for practically usable plugin double-score strategies on ID ImageNet. The relative cost is $\alpha = 0.5$.}
\caption{Comparison of double-score strategies using logistic regression with standard and corrected sigmoid for approximating the likelihood ratio (LR) $g(x)$ from a mixture of data with $\pi_O^{\rm tr}=\nicefrac{1}{2}$ (see ~\cref{sec:plugin_estimate_of_optimal_scod}). Table displays \areaSCODTpr{}$\downarrow$ in \% for selective classifiers $(h,c)$ from an ID classifier $h(x)$ and selectors $c(x)=\leftbb s(x)\leq \lambda \rightbb$. Results show practically usable plugin double-score strategies on ID ImageNet. Relative cost: $\alpha = 0.5$.}
%\resizebox{\columnwidth}{!}{%
\begin{tabular}{l|l|c|c|c|c|c}
Sigmoid& \backslashbox{\small \it Score}{\small \it Dataset} & ssb\_hard & ninco & inaturalist & textures & openimage\_o\\ \hline
\multirow{3}{*}{\rotatebox{0}{Corrected}} & SIRC%(MSP, -LR)
& 
14.27   & 12.67   & {8.02}    & {10.50}   & {9.75}    \\ \cline{2-7}
& Linear %(1-MSP, LR)
& 
\bf6.37 & \bf7.44 & \bf4.16 & \bf4.38 & \bf6.48 \\ 
\cline{2-7}
& LR &
14.25 & 16.67 & 12.70 & 13.19 & 15.65
\\
\hline \hline
\multirow{3}{*}{\rotatebox{0}{Standard}} 
& SIRC %(MSP, -LR)
& 
11.27 & 9.28 & 4.34 & 5.94 & 7.22 \\ 
\cline{2-7}
& Linear %(1-MSP, LR)
& 
7.28 & 8.83 & 5.15 & 5.33 & 7.22 
\\ 
\cline{2-7}
& LR & 
13.88 & 15.21 & 11.32 & 11.54 & 13.57 
%\\
%\hline \hline
%& MSP 
%& 17.38 & 13.43 & 9.35 & 12.08 & 11.03 \\ 
\end{tabular}
%}
\label{tab:imagenet_ploscod}
\end{table}

\boldparagraph{Comparison with contemporary scoring functions}
Our results demonstrate that when provided with reasonable estimates $\hat{r}(x)$, $\hat{g}(x)$ of the optimal scores, the linear strategy is a state-of-the-art SCOD selector. However, we refrain from making this claim with currently employed OODD scoring functions. In~\cref{tab:sirc_vs_linear_with_posthoc} we show the performance of SIRC and the linear strategy on ImageNet when combining contemporary scores.~\Cref{tab:sirc_vs_linear_with_posthoc} only shows a subset of the evaluated scores; for SIRC, the best-performing score, and scores~\cite{Huang-Importance-NIPS2021,Wang-ViM-CVPR2022} mirroring the original setup in~\cite{Xia-SIRC-ACCV2022}. For the linear double-score strategy, the three best-performing combinations are shown. In some cases, SIRC outperforms the linear strategy when using contemporary OODD scores. However, with the \Algo{} estimate $\hat{g}(x)$, the linear strategy outperforms all other methods by a large margin. For results with all scores on all datasets, refer to~\cref{asec:imagenet,asec:cifar10,asec:cifar100}.

\begin{table}[htbp]
\centering
\caption{\areaSCODTpr{}$\downarrow$ in \% points for practically usable strategies. The table contains single-score strategies that when combined with MSP achieve the best results. The likelihood ratio (LR) $\hat{g}(x)$ was approximated by \Algo{}. In some cases, SIRC outperforms the linear strategy. When using the plugin LR estimate, the linear strategy significantly outperforms SIRC and all single-score strategies. The best results with contemporary OODD scores are shown in bold. The best results overall are highlighted in green.}
%\resizebox{\columnwidth}{!}{%
\begin{tabular}{l|l|c|c|c|c|c}
\multicolumn{2}{c|}{{\backslashbox{\it  Score}{\it  Dataset}}} &  ssb\_hard &  ninco &  inaturalist &  textures & openimage\_o\\ \hline
\multirow{6}{*}{\rotatebox{90}{\bf Single Score}} 
&  ASH~\cite{Djurisic-ASH-ICLR2023}
& 19.04 & 13.76 & 6.95 & 7.03 & 8.85
\\ \cline{2-7}
&  GradNorm~\cite{Huang-Importance-NIPS2021}             
& 23.95 & 22.89 & 12.95 & 13.88 & 17.49
\\ \cline{2-7}
&  $L_1$-norm~\cite{Huang-Importance-NIPS2021}         
& 32.27 & 36.06 & 39.22 & 28.37 & 32.86 
\\ \cline{2-7}
&  ODIN~\cite{liang2018enhancing}   
& 19.79 & 16.78 & 10.08 & 11.16 & 11.54
\\ \cline{2-7}
&  ReAct~\cite{Sun-NIPS2021}              
& 18.88 & 14.53 & 7.23 & 9.00 & 9.46 
\\ \cline{2-7}
&  Residual~\cite{Wang-ViM-CVPR2022}           
& 40.72 & 35.98 & 37.03 & 18.59 & 31.93
\\
\hline\hline
\multirow{4}{*}{\rotatebox{90}{{\bf Linear}}} 
&  ASH~\cite{Djurisic-ASH-ICLR2023}
& 18.74 & 13.47 & 6.82 & \bf6.95 & \bf8.66
\\ \cline{2-7}
&  ODIN~\cite{liang2018enhancing}          
& 17.40 & 13.51 & 9.27 & 12.21 & 11.02
\\ \cline{2-7}
&  ReAct~\cite{Sun-NIPS2021}        
& 17.85 & 13.37 & \bf6.76 & 8.62 & 8.77
\\ \cline{2-7}
&  \Algo{} LR
& \cellcolor{\bestcolor} 6.37 & \cellcolor{\bestcolor}7.44 & \cellcolor{\bestcolor}4.16 & \cellcolor{\bestcolor}4.38 & \cellcolor{\bestcolor}6.48
\\
\hline\hline
\multirow{4}{*}{\rotatebox{90}{{\bf SIRC}}} 
&  GradNorm~\cite{Huang-Importance-NIPS2021}             
& \bf16.99 & \bf12.96 & 7.70 & 10.24 & 9.80
\\ \cline{2-7}
&  $L_1$-norm~\cite{Huang-Importance-NIPS2021}          
& 17.21 & 13.54 & 9.50 & 11.90 & 10.96
\\ \cline{2-7}
&  Residual~\cite{Wang-ViM-CVPR2022}           
& 17.49 & 13.43 & 9.20 & 9.54 & 10.62
\\ \cline{2-7}
&  \Algo{} LR
& 14.27 & 12.67 & 8.02 & 10.50 & 9.75
\end{tabular}
%}
\label{tab:sirc_vs_linear_with_posthoc}
\end{table}

\section{Conclusions}

This study addresses the SCOD problem~\cite{Xia-SIRC-ACCV2022}, focusing on scenarios in which ID test samples are contaminated by OOD data. Our key contributions include demonstrating that the optimal SCOD strategy involves a Bayes classifier for ID data and a selector corresponding to a stochastic linear classifier in a 2D space. This contrasts with suboptimal strategies used in contemporary OOD detection methods and SIRC~\cite{Xia-SIRC-ACCV2022}, the current state-of-the-art on the SCOD problem. We establish the non-learnability of SCOD in a distribution-free setting when relying solely on an ID data sample. This result highlights the inherent challenges of PAC learning for SCOD without access to OOD data. We introduced \Algo{}, a method for learning the plugin estimate of the optimal SCOD strategy from both an ID data sample and an unlabeled mixture of ID and OOD data. Empirical validations confirm our theoretical findings and  demonstrate that \Algo{} outperforms existing OODD methods and SIRC in solving the SCOD problem. 
%These results underscore the practical significance of our proposed approach for developing robust and effective classifiers in the presence of OOD data. 

%The main insight is that the optimal prediction strategy must trade-off the ability to detect misclassified examples and to distinguish ID from OOD samples. This is in contrast to existing OOD methods that output a single uncertainty score. We propose a simple and effective double-score strategy that allows us to boost performance of two existing OOD methods by combining their uncertainty scores. Finally, we suggest improved evaluation metrics for assessing OOD methods that simultaneously evaluate all aspects of the OOD methods and are directly related to the optimal OOD strategy under the proposed reject option models.

\bibliographystyle{splncs04}
\bibliography{main}

\begin{thebibliography}{10}
\providecommand{\url}[1]{\texttt{#1}}
\providecommand{\urlprefix}{URL }
\providecommand{\doi}[1]{https://doi.org/#1}

\bibitem{Bitterwolf-NINCO-ICML2023}
Bitterwolf, J., Mueller, M., Hein, M.: In or out? fixing imagenet
  out-of-distribution detection evaluation. In: ICML (2023),
  \url{https://proceedings.mlr.press/v202/bitterwolf23a.html}

\bibitem{Blanchard-SemiSuperNovDet-JMLR2010}
Blanchard, G., Lee, G., Scott, C.: Semi-supervised novelty detection. Journal
  of Machine Learning Research  (2010)

\bibitem{Cen-Devil-ICLR2023}
Cen, J., Luan, D., Zhang, S., Pei, Y., Zhang, Y., Zhao, D., Shen, S., Che, Q.:
  The devil is in the wrongly-classified samples: towards unified open-set
  recognition. In: ICLR (2023)

\bibitem{Chen-PAMI2022}
Chen, G., Peng, P., Wang, X., Tian, Y.: Adversarial reciprocal points learning
  for open set recognition. IEEE Transactions on Pattern Analysis and Machine
  Intelligence  \textbf{44}(11),  8065--8081 (2022)

\bibitem{Chow-RejectOpt-TIT1970}
Chow, C.: On optimum recognition error and reject tradeoff. IEEE Transactions
  on Information Theory  \textbf{16}(1),  41–46 (1970)

\bibitem{Cimpoi-Textures-CVPR2014}
Cimpoi, M., Maji, S., Kokkinos, I., Mohamed, S., Vedaldi, A.: Describing
  textures in the wild. In: 2014 IEEE Conference on Computer Vision and Pattern
  Recognition. pp. 3606--3613 (2014). \doi{10.1109/CVPR.2014.461}

\bibitem{Devries-arxiv2018}
DeVries, T., Taylor, G.W.: Learning confidence for out-of-distribution
  detection in neural networks. arXiv preprint:1802.04865  (2018)

\bibitem{Dhamija-NIPS2018}
Dhamija, A.R., G\"{u}nther, M., Boult, T.: Reducing network agnostophobia. In:
  Bengio, S., Wallach, H., Larochelle, H., Grauman, K., Cesa-Bianchi, N.,
  Garnett, R. (eds.) Advances in Neural Information Processing Systems.
  vol.~31. Curran Associates, Inc. (2018)

\bibitem{Djurisic-ASH-ICLR2023}
Djurisic, A., Bozanic, N., Ashok, A., Liu, R.: Extremely simple activation
  shaping for out-of-distribution detection. In: The Eleventh International
  Conference on Learning Representations (2023),
  \url{https://openreview.net/forum?id=ndYXTEL6cZz}

\bibitem{Zhen-IsOODLearnable-NIPS2022}
Fang, Z., Li, Y., Lu, J., Dong, J., Han, B., Liu, F.: Is out-of-distribution
  detection learnable? In: Koyejo, S., Mohamed, S., Agarwal, A., Belgrave, D.,
  Cho, K., Oh, A. (eds.) Advances in Neural Information Processing Systems.
  vol.~35, pp. 37199--37213. Curran Associates, Inc. (2022)

\bibitem{Franc-Optimal-JMLR2023}
Franc, V., Prusa, D., Voracek, V.: Optimal strategies for reject option
  classifiers. Journal of Machine Learning Research  \textbf{24}(11),  1--49
  (2023)

\bibitem{Geifman-SelectClass-NIPS2017}
Geifman, Y., El-Yaniv, R.: Selective classification for deep neural networks.
  In: Advances in Neural Information Processing Systems 30. pp. 4878--4887
  (2017)

\bibitem{Granese-Nips2021}
Granese, F., Romanelli, M., Gorla, D., Palamidessi, C., Piantanida, P.: Doctor:
  A simple method for detecting misclassification errors. In: Ranzato, M.,
  Beygelzimer, A., Dauphin, Y., Liang, P., Vaughan, J.W. (eds.) Advances in
  Neural Information Processing Systems. vol.~34, pp. 5669--5681. Curran
  Associates, Inc. (2021)

\bibitem{Hendrycks-MLS-ICML2022}
Hendrycks, D., Basart, S., Mazeika, M., Zou, A., Kwon, J., Mostajabi, M.,
  Steinhardt, J., Song, D.: Scaling out-of-distribution detection for
  real-world settings. In: Chaudhuri, K., Jegelka, S., Song, L., Szepesvari,
  C., Niu, G., Sabato, S. (eds.) Proceedings of the 39th International
  Conference on Machine Learning. Proceedings of Machine Learning Research,
  vol.~162, pp. 8759--8773. PMLR (Jul 2022)

\bibitem{Hendrycks-baseline-ICLR17}
Hendrycks, D., Gimpel, K.: A baseline for detecting misclassified and
  out-of-distribution examples in neural networks. In: Proceedings of
  International Conference on Learning Representations (2017)

\bibitem{Huang-Importance-NIPS2021}
Huang, R., Geng, A., Li, Y.: On the importance of gradients for detecting
  distributional shifts in the wild. In: Advances in Neural Information
  Processing Systems (2021)

\bibitem{Huang-MOS-CVPR2021}
Huang, R., Li, Y.: Mos: Towards scaling out-of-distribution detection for large
  semantic space. In: Proceedings of the IEEE/CVF Conference on Computer Vision
  and Pattern Recognition (CVPR). pp. 8710--8719 (June 2021)

\bibitem{Katz-Habitats-ICML2022}
Katz-Samuels, J., Nakhleh, J., Nowak, R., Li, Y.: Training ood detectors in
  their natural habitats. In: ICML (2022)

\bibitem{Kim-UnifBench-ESA2021}
Kim, J., Koo, J., Hwang, S.: A unified benchmark for the unknown detection
  capability of deep neural networks. Expert Syst. Appl.  \textbf{229},  120461
  (2021), \url{https://api.semanticscholar.org/CorpusID:244773165}

\bibitem{Krizhevsky-Learning-2009}
Krizhevsky, A.: Learning multiple layers of features from tiny images. Tech.
  rep. (2009)

\bibitem{Kuznetsova-OpenImages-IJCV2020}
Kuznetsova, A., Rom, H., Alldrin, N., Uijlings, J., Krasin, I., Pont-Tuset, J.,
  Kamali, S., Popov, S., Malloci, M., Kolesnikov, A., Duerig, T., Ferrari, V.:
  The open images dataset v4: Unified image classification, object detection,
  and visual relationship detection at scale. International Journal of Computer
  Vision  \textbf{128}(7),  1956–1981 (Mar 2020).
  \doi{10.1007/s11263-020-01316-z},
  \url{http://dx.doi.org/10.1007/s11263-020-01316-z}

\bibitem{Le-TIN-2015}
Le, Y., Yang, X.S.: Tiny imagenet visual recognition challenge (2015)

\bibitem{Lecun-MNIST-1998}
Lecun, Y., Bottou, L., Bengio, Y., Haffner, P.: Gradient-based learning applied
  to document recognition. Proceedings of the IEEE  \textbf{86}(11),
  2278--2324 (1998). \doi{10.1109/5.726791}

\bibitem{liang2018enhancing}
Liang, S., Li, Y., Srikant, R.: Enhancing the reliability of
  out-of-distribution image detection in neural networks. In: International
  Conference on Learning Representations (2018)

\bibitem{Weitang-EBO-NIPS2020}
Liu, W., Wang, X., Owens, J., Li, Y.: Energy-based out-of-distribution
  detection. In: Larochelle, H., Ranzato, M., Hadsell, R., Balcan, M., Lin, H.
  (eds.) Advances in Neural Information Processing Systems. vol.~33, pp.
  21464--21475. Curran Associates, Inc. (2020),
  \url{https://proceedings.neurips.cc/paper_files/paper/2020/file/f5496252609c43eb8a3d147ab9b9c006-Paper.pdf}

\bibitem{Torchvision-GitHub2016}
maintainers, T., contributors: Torchvision: Pytorch's computer vision library.
  \url{https://github.com/pytorch/vision} (2016)

\bibitem{Malinin-NEURIPS2018}
Malinin, A., Gales, M.: Predictive uncertainty estimation via prior networks.
  In: Bengio, S., Wallach, H., Larochelle, H., Grauman, K., Cesa-Bianchi, N.,
  Garnett, R. (eds.) Advances in Neural Information Processing Systems.
  vol.~31. Curran Associates, Inc. (2018)

\bibitem{Narasimhan-Plugin-Arxiv2023}
Narasimhan, H., Krisna~Menon, A., Jitkrittum, W., Kumar, S.: Plugin estimators
  for selective classification with out-of-distribution detection. arXiv
  preprint:2301.12386v4  (2023)

\bibitem{Neal-GradNorm-ECCV2018}
Neal, L., Olson, M., Fern, X., Wong, W.K., Li, F.: Open set learning with
  counterfactual images. In: Proceedings of the European Conference on Computer
  Vision (ECCV) (September 2018)

\bibitem{Netzer-SVHN-NIPS2011}
Netzer, Y., Wang, T., Coates, A., Bissacco, A., Wu, B., Ng, A.Y.: Reading
  digits in natural images with unsupervised feature learning. In: NIPS
  Workshop on Deep Learning and Unsupervised Feature Learning 2011 (2011),
  \url{http://ufldl.stanford.edu/housenumbers/nips2011_housenumbers.pdf}

\bibitem{Neyman-1928}
Neyman, J., Person, E.: On the use and interpretation of certain test criteria
  for purpose of statistical inference. Biometrica pp. 175--240 (1928)

\bibitem{Pietraszek-AbstainROC-ICML2005}
Pietraszek, T.: Optimizing abstaining classifiers using {ROC} analysis. In:
  Proceedings of the 22nd International Conference on Machine Learning. p.
  665–672 (2005)

\bibitem{Ridnik-ImageNet-21K-NIPS2021}
Ridnik, T., Ben-Baruch, E., Noy, A., Zelnik-Manor, L.: Imagenet-21k pretraining
  for the masses. In: Thirty-fifth Conference on Neural Information Processing
  Systems Datasets and Benchmarks Track (Round 1) (2021),
  \url{https://openreview.net/forum?id=Zkj_VcZ6ol}

\bibitem{Shwartz-ML-2014}
Shalev-Shwartz, S., Ben-David, S.: Understanding Machine Learning: From Theory
  to Algorithms. Cambridge University Press, USA (2014)

\bibitem{Song-Neurips2022}
Song, Y., Sebe, N., Wang, W.: Rankfeat: Rank-1 feature removal for
  out-of-distribution detection. In: Koyejo, S., Mohamed, S., Agarwal, A.,
  Belgrave, D., Cho, K., Oh, A. (eds.) Advances in Neural Information
  Processing Systems. vol.~35, pp. 17885--17898. Curran Associates, Inc. (2022)

\bibitem{Sun-NIPS2021}
Sun, Y., Guo, C., Li, Y.: React: Out-of-distribution detection with rectified
  activations. In: Ranzato, M., Beygelzimer, A., Dauphin, Y., Liang, P.,
  Vaughan, J.W. (eds.) Advances in Neural Information Processing Systems.
  vol.~34, pp. 144--157. Curran Associates, Inc. (2021)

\bibitem{pmlr-v162-sun22d}
Sun, Y., Ming, Y., Zhu, X., Li, Y.: Out-of-distribution detection with deep
  nearest neighbors. In: Chaudhuri, K., Jegelka, S., Song, L., Szepesvari, C.,
  Niu, G., Sabato, S. (eds.) Proceedings of the 39th International Conference
  on Machine Learning. Proceedings of Machine Learning Research, vol.~162, pp.
  20827--20840. PMLR (Jul 2022)

\bibitem{Torralba-TinyImages-TPAMI2008}
Torralba, A., Fergus, R., Freeman, W.T.: 80 million tiny images: A large data
  set for nonparametric object and scene recognition. IEEE Transactions on
  Pattern Analysis and Machine Intelligence  \textbf{30}(11),  1958--1970
  (2008). \doi{10.1109/TPAMI.2008.128}

\bibitem{Horn-iNaturalist-CVPR2018}
Van~Horn, G., Mac~Aodha, O., Song, Y., Cui, Y., Sun, C., Shepard, A., Adam, H.,
  Perona, P., Belongie, S.: The inaturalist species classification and
  detection dataset. In: Proceedings of the IEEE Conference on Computer Vision
  and Pattern Recognition (CVPR) (June 2018)

\bibitem{Vaze-Openset-ICLR2022}
Vaze, S., Han, K., Vedaldi, A., Zisserman, A.: Open-set recognition: a good
  closed-set classifier is all you need? In: International Conference on
  Learning Representations (2022)

\bibitem{Wang-ViM-CVPR2022}
Wang, H., Li, Z., Feng, L., Zhang, W.: Vim: Out-of-distribution with
  virtual-logit matching. In: 2022 IEEE/CVF Conference on Computer Vision and
  Pattern Recognition (CVPR). pp. 4911--4920 (2022)

\bibitem{Xia-SIRC-ACCV2022}
Xia, G., Bouganis, C.S.: Augmenting softmax information for selective
  classification with out-of-distribution data. In: Proceedings of the Asian
  Conference on Computer Vision (ACCV). pp. 1995--2012 (December 2022)

\bibitem{Yang-Nips2022}
Yang, J., Wang, P., Zou, D., Zhou, Z., Ding, K., Peng, W., Wang, H., Chen, G.,
  Li, B., Sun, Y., Du, X., Zhou, K., Zhang, W., Hendrycks, D., Li, Y., Liu, Z.:
  Openood: Benchmarking generalized out-of-distribution detection. In:
  Conference on Neural Information Processing Systems (NeurIPS 2022) Track on
  Datasets and Benchmar (2022)

\bibitem{Zhou-Places-PAMI2017}
Zhou, B., Lapedriza, A., Khosla, A., Oliva, A., Torralba, A.: Places: A 10
  million image database for scene recognition. IEEE Transactions on Pattern
  Analysis and Machine Intelligence  (2017)

\end{thebibliography}

\clearpage
\appendix
\centerline{\textbf{\Large  \textit{Supplementary:}
SCOD: From Heuristics to Theory}}

\section{Alternative formulations of the SCOD problem}
\label{asec:alternative_formulations}

In this paper, we address the SCOD problem~\equ{equ:SingeCostModel}, further denoted SCOD-tpr, which aims to minimize the SCOD risk $R(h,c)=(1-\alpha)\Rsel(h,c)+\alpha \fpr(c)$ subject to a constraint ensuring a minimum TPR of $\tprMin$. The TPR represents the probability that an ID sample will be accepted for classification by a selector $c(x)$, defined as $\tpr(c)=\EE_{x\sim p_I(x)}c(x)$. 
An alternative formulation, discussed in~\cite{Narasimhan-Plugin-Arxiv2023}, replaces TPR with total coverage $\rho(c)=\tpr(c)(1-\pi_O)+\fpr(c)\pi_O$ which represents the probability of accepting an input sample, whether generated from ID or OOD. Consequently, we arrive at the SCOD-coverage formulation:
\begin{equation}
 \label{equ:ScodCoverage}
       \min_{\genfrac{}{}{0pt}{}{h\in\SY^\SX}{c\in[0,1]^\SX}} \big [ (1-\relCost)\,\Rsel(h,c) + \relCost\, \fpr(c)\big ]
     \qquad\mbox{s.t.}\qquad \rho(c) \geq \covMin \:,
\end{equation}
where $\covMin\in (0,1)$ is a user-defined minimum acceptable coverage. 

Both formulations are well-defined, and the choice between them, based on whether one favors accepting a specific portion of ID or a portion of all samples, depends on the specific application. However, the SCOD-tpr formulation analyzed in this paper offers several practical advantages over the SCOD-coverage formulation:
\begin{enumerate}
  \item  The optimal strategy $(h^*,c^*)$ for solving the SCOD-tpr formulation~\equ{equ:SingeCostModel} is independent of the portion of OOD data $\pi_O$. None of the key quantities, namely $R_S(h,c)$, $\tpr(c)$, and $\fpr(c)$, are influenced by $\pi_O$. This property is crucial because, in practice, $\pi_O$ is not only unknown but often nonstationary. In contrast, the SCOD-coverage formulation relies on the OOD portion $\pi_O$ through the coverage $\rho(c)$. Consequently, it is not applicable in a nonstationary setup.
  \item The optimal selector for both the SCOD-tpr and SCOD-coverage formulations relies on a linear combination of the conditional risk $r(x)$ and the ID/OOD score $g(x)=p_I(x)/p_O(x)$. Specifically, the optimal score is $s(x)=r(x)+\beta\, g(x)$ where $\beta\in\Re$ is a multiplier dependent on the problem parameters. In~\equ{equ:optimalScodSelector}, we demonstrated that the optimal multiplier $\beta$ for the SCOD-tpr can be analytically computed as $\beta=\alpha \tprMin/(1-\alpha)$. Conversely, for SCOD-coverage, there is no analytical formula for computing $\beta$, and it necessitates solving an optimization problem based on quantities reliant on clean OOD data (refer to~\cite{Narasimhan-Plugin-Arxiv2023} for further details).
\end{enumerate}
In summary, both SCOD-tpr and SCOD-coverage formulations are well-defined and intuitive. However, SCOD-tpr is independent of $\pi_O$, possesses an analytically solvable multiplier $\beta$ for the optimal combination of $r(x)$ and $g(x)$, and learning the optimal selective classifier does not necessitate clean OOD samples. Conversely, SCOD-coverage formulation inherently depends on $\pi_O$, requires optimization of the multiplier $\beta$ for combining $r(x)$ and $g(x)$, and learning the optimal selective classifier requires clean OOD samples.

%\boldparagraph{Formulation}
%\hspace{-0.5em}\revone has raised concerns regarding the formulation of the SCOD problem (\Cref{def:SCOD}). The suggestion is to substitute the constraint on $\tpr$ with a constraint on coverage. Both formulations are valid; coverage constraint can be found in \cite{Narasimhan-Plugin-Arxiv2023}. While coverage can be determined at test-time, the threshold $\lambda$ (see \cref{thm:SingleCostModel}) can be tuned using validation data and then fixed. Therefore, accessing $\tpr$ at test time is not necessary. In the paper, we show several advantages of constraining $\tpr$:
%\begin{enumerate}
%    \item The problem definition becomes independent of the prior $\pi_O$; hence the optimal solution is independent of prior
%    \item constraining $\tpr$ enables the explicit formulation of the optimal score, \cref{equ:optScore}. In contrast, the coverage-based formulation necessitates solving an optimization task to find the mixing coefficient for $r(x)$ and $g(x)$, see \cite{Narasimhan-Plugin-Arxiv2023}
    %\item constraining $\tpr$ aligns with the task's goal; practitioners are likely to prefer accepting a specific portion of ID rather than of all samples.
%\end{enumerate}
\section{Implementation Details}
\label{asec:implementation_details}
\subsection{Datasets}
\label{assec:data}
We adopt datasets from the OpenOOD benchmark \cite{Yang-Nips2022} and assess SCOD performance across three ID datasets: CIFAR-10, CIFAR-100, and ImageNet-1K. 

\boldparagraph{ID ImageNet-1K}
In line with \cite{Yang-Nips2022}, we employ 45,000 images from the ImageNet-1K validation set as the \textit{in-distribution} (ID) data for evaluation. The official ImageNet-1K test set could \textit{not} be utilized for this purpose as the ground-truth labels are \textit{not} publicly accessible. For the near-OOD group we employ the SSB\_Hard \cite{Vaze-Openset-ICLR2022} and NINCO \cite{Bitterwolf-NINCO-ICML2023} datasets. In the far-OOD group we include the iNaturalist \cite{Huang-MOS-CVPR2021,Horn-iNaturalist-CVPR2018}, Textures \cite{Cimpoi-Textures-CVPR2014}, and OpenImage\_O \cite{Wang-ViM-CVPR2022,Kuznetsova-OpenImages-IJCV2020} datasets. A brief overview of the datasets is provided below:

\begin{enumerate}
\item \textbf{SSB\_Hard} \cite{Vaze-Openset-ICLR2022}: A dataset comprising 49,000 images covering 980 categories selected from ImageNet-21K \cite{Ridnik-ImageNet-21K-NIPS2021} that are absent in ImageNet-1K. We evaluate on the entire dataset.

\item \textbf{NINCO} \cite{Bitterwolf-NINCO-ICML2023}: A dataset of 5,879 images with manually filtered-out noise. The majority of the dataset was extracted from the species subset of iNaturalist \cite{Horn-iNaturalist-CVPR2018}. It is intentionally designed to be challenging to differentiate from ImageNet-1K samples, with examples such as a marbled newt considered OOD, while a common newt is considered ID. We evaluate on the entire dataset.
\item \textbf{OpenImage\_O} \cite{Wang-ViM-CVPR2022,Kuznetsova-OpenImages-IJCV2020}: A dataset consisting of 17,632 images manually selected from the test set of the OpenImages dataset \cite{Kuznetsova-OpenImages-IJCV2020}. We use a subset defined by the OpenOOD \cite{Yang-Nips2022} benchmark, comprising 15,869 images.
\item \textbf{iNaturalist} \cite{Huang-MOS-CVPR2021, Horn-iNaturalist-CVPR2018}: A dataset consisting of 10,000 images randomly sampled from 110 manually selected \textit{plant} classes not present in ImageNet-1K. The samples were obtained by \cite{Huang-MOS-CVPR2021} from the full iNaturalist dataset \cite{Horn-iNaturalist-CVPR2018}. We evaluate on all 10,000 samples.
\item \textbf{Textures} \cite{Cimpoi-Textures-CVPR2014}: A dataset consisting of 5,640 images, split into 47 texture classes (e.g., braided, striped, wrinkled). We evaluate on the entire dataset.
\end{enumerate}

\boldparagraph{ID CIFAR-10}
In line with \cite{Yang-Nips2022}, we use 9,000 images from the CIFAR-10 \cite{Krizhevsky-Learning-2009} test set as the \textit{in-distribution} (ID) data for evaluation. For the near-OOD group, we employ the CIFAR-100 \cite{Krizhevsky-Learning-2009} and Tiny-ImageNet (TIN) \cite{Le-TIN-2015} datasets. In the far-OOD group, we include the Street View House Numbers (SVHN) \cite{Netzer-SVHN-NIPS2011}, Places365 \cite{Zhou-Places-PAMI2017}, MNIST \cite{Lecun-MNIST-1998}, and Textures \cite{Cimpoi-Textures-CVPR2014} datasets. A brief overview of the datasets and a description of how we utilize them are provided below:

\begin{enumerate}
\item \textbf{CIFAR-100} \cite{Krizhevsky-Learning-2009}: Dataset of 60,000 images across 100 classes, selected from the Tiny Images \cite{Torralba-TinyImages-TPAMI2008} dataset such that there is no semantic overlap with CIFAR-10. The test set comprises 10,000 images, however, we evaluate the methods only on a subset of 9,000 images defined by the OpenOOD \cite{Yang-Nips2022} benchmark.
\item \textbf{TIN} \cite{Le-TIN-2015}: Dataset of 100,000 images divided into 200 classes. For every class, there are 500 training images, 50 validating images, and 50 test images. Following the OpenOOD \cite{Yang-Nips2022} benchmark, we evaluate the methods on the validation set and only use 7793 of the 10,000 images as the OOD dataset, removing samples that semantically overlap with CIFAR-10.

\item \textbf{SVHN} \cite{Netzer-SVHN-NIPS2011}: Dataset of 99,289 house numbers taken from Google Street View images. We use the pre-processed \textit{cropped} variant of the dataset, and evaluate the methods on the test set comprising 26,032 images.

\item \textbf{Places365} \cite{Zhou-Places-PAMI2017}: Scene recognition dataset comprising over 10,000,000 images divided into 434 scene classes. Following the OpenOOD benchmark \cite{Yang-Nips2022}, we use the \textit{Places365-Standard} version of the dataset and evaluate on the validation set. We use only 35,195 out of the 36,000 images due to semantic overlap of the samples with CIFAR-10.

\item \textbf{Textures} \cite{Cimpoi-Textures-CVPR2014}: A dataset consisting of 5,640 images, split into 47 texture classes (e.g., braided, striped, wrinkled). We evaluate on the entire dataset.

\item \textbf{MNIST} \cite{Lecun-MNIST-1998}: Dataset comprising of 70,000 images of handwritten digits. Following the OpenOOD benchmark \cite{Yang-Nips2022}, we evaluate on the entire dataset.
\end{enumerate}

\boldparagraph{ID CIFAR-100}
In line with \cite{Yang-Nips2022}, we use 9,000 images from the CIFAR-100 \cite{Krizhevsky-Learning-2009} test set as the \textit{in-distribution} (ID) data for evaluation. For the near-OOD group, we employ the CIFAR-10 \cite{Krizhevsky-Learning-2009} and Tiny-ImageNet (TIN) \cite{Le-TIN-2015} datasets. In the far-OOD group, we include the Street View House Numbers (SVHN) \cite{Netzer-SVHN-NIPS2011}, Places365 \cite{Zhou-Places-PAMI2017}, MNIST \cite{Lecun-MNIST-1998}, and Textures \cite{Cimpoi-Textures-CVPR2014} datasets. A brief overview of the datasets and a description of how we utilize them are provided below:

\begin{enumerate}
\item \textbf{CIFAR-10} \cite{Krizhevsky-Learning-2009}: Dataset of 60,000 images across 10 classes, selected from the Tiny Images \cite{Torralba-TinyImages-TPAMI2008} dataset such that there is no semantic overlap with CIFAR-100. The test set comprises 10,000 images, however, we evaluate the methods only on a subset of 9,000 images defined by the OpenOOD \cite{Yang-Nips2022} benchmark.

\item \textbf{TIN} \cite{Le-TIN-2015}: Dataset of 100,000 images divided into 200 classes. For every class, there are 500 training images, 50 validating images, and 50 test images. Following the OpenOOD \cite{Yang-Nips2022} benchmark, we evaluate the methods on the validation set and only use 6,526 of the 10,000 images as the OOD dataset, removing samples that semantically overlap with CIFAR-100.

\item \textbf{SVHN} \cite{Netzer-SVHN-NIPS2011}: Dataset of 99,289 house numbers taken from Google Street View images. We use the pre-processed \textit{cropped} variant of the dataset, and evaluate the methods on the test set comprising 26,032 images.

\item \textbf{Places365} \cite{Zhou-Places-PAMI2017}: Scene recognition dataset comprising over 10,000,000 images divided into 434 scene classes. Following the OpenOOD benchmark \cite{Yang-Nips2022}, we use the \textit{Places365-Standard} version of the dataset and evaluate on the validation set. We use only 33,773 out of the 36,000 images due to semantic overlap of the samples with CIFAR-100.

\item \textbf{Textures} \cite{Cimpoi-Textures-CVPR2014}: A dataset consisting of 5,640 images, split into 47 texture classes (e.g., braided, striped, wrinkled). We evaluate on the entire dataset.

\item \textbf{MNIST} \cite{Lecun-MNIST-1998}: Dataset comprising of 70,000 images of handwritten digits. Following the OpenOOD benchmark \cite{Yang-Nips2022}, we evaluate on the entire dataset.
\end{enumerate}

\subsection{Models}
\label{assec:models}
\boldparagraph{CIFAR-10/100}
For the ID classifier $h(x)$ on CIFAR-10/100, we use pre-trained ResNet-18 classifiers from the OpenOOD benchmark by~\cite{Yang-Nips2022}. All of the models were originally trained using standard cross-entropy loss, employing the SGD optimizer with a momentum of $0.9$, a learning rate set to $0.1$, and a cosine annealing decay schedule. Additionally, a weight decay of $0.0005$ was applied. During evaluation, we apply input normalization identical to the one used during the training of the classifier; the normalization is different for CIFAR-10 and CIFAR-100.

\boldparagraph{ImageNet-1K}
For the ID classifier $h(x)$ on ImageNet-1K, we use a pre-trained ResNet-50 model from Torchvision \cite{Torchvision-GitHub2016}. Specifically, we use the \texttt{IMAGENET1K\_V1} model. During evaluation, we apply input normalization identical to the one used during the training of the classifier.

\subsection{Approximating the likelihood ratio}
\label{assec:training_likelihood_ratio}
To estimate the OOD/ID likelihood ratio $g(x)$ on real-world data, we train a classifier with BCE of the standard sigmoid (a.k.a logistic regression) and a \textit{corrected sigmoid} model (refer to \cref{sec:plugin_estimate_of_optimal_scod}) using features from the ID classifier. In other words, we replace the last layer of the classifier $h(x)$ with a linear layer with a single output. Additionally, we add a \texttt{Dropout} layer with $p_{\rm train}=0.2$ before the linear layer. For the experiments on CIFAR-10/100, we allow all weights of $h(x)$ to be modified. For experiments on ImageNet-1K, we only learn the last layer.

\boldparagraph{Training details}
During training, we apply several data augmentations, including rotation up to 20 degrees, random resized cropping, and horizontal mirroring. For ImageNet-1K, we extend the augmentation strategy to incorporate variations in brightness, contrast, saturation, and hue. We train the models with binary cross-entropy loss using the AMSGrad variant of the Adam optimizer with a learning rate of $0.003$ for a total of $200$ epochs. After every $50$ epochs, the learning rate is decreased by a factor of $10$. The final model is selected based on the validation loss. For further details, please refer to the implementation.

\boldparagraph{Data splitting}
When training the likelihood ratio approximation $\hat{g}(x)$, we use the dedicated training sets of ImageNet-1K, CIFAR-10, CIFAR-100, Tiny-ImageNet, and SVHN. Conversely, for MNIST, Places365, SSB\_Hard, NINCO, iNaturalist, Textures, and OpenImage\_O, we adopt a random split approach, allocating 50\% of each dataset for training $\hat{g}(x)$ and the remaining 50\% for evaluation. Note that the evaluations are always reported exclusively on samples unseen during training.

\boldparagraph{Mixture}
In \cref{{sec:plugin_estimate_of_optimal_scod}} we adopt the framework proposed by~\cite{Katz-Habitats-ICML2022}, where in addition to the ID data sample $\ST_{I}=((x^I_i,y^I_i)\in\SX\times\SY\mid i=1,\ldots,m)$ generated from $p_I(x,y)$ we also have an unlabeled sample of a mixture of ID and OOD data $\ST_{U}=(x^U_i\in\SX\mid i=1,\ldots,n)$ generated from $p(x)=\pi_O p_O^{\rm tr}(x) + (1-\pi_O^{\rm tr}) p_I(x)$. We demonstrate that within this framework, the SCOD problem can be effectively addressed using the proposed \Algo{} algorithm. In our experimental setup, we set $\pi_O^{\rm eval}$ to 50\%, indicating an equal distribution of ID and OOD samples in the unlabeled mixture used during evaluation. We use the same evaluation mixture irrespective of the training prior $\pi_O^{\rm tr}$. Results for varying $\pi_O^{\rm tr}$ can be found in~\cref{asec:ablation}.

\subsection{Hyperparameters of double-score strategies}
\label{assec:hyperparameters}
The current state-of-the-art SCOD selector $s_{\rm SIRC}(x) = -(S_1^{max}-s_1(x))(1+\exp(-b(s_2(x)-a))$ employs a nonlinear combination of scores $s_1(x)$ and $s_2(x)$ with two hyperparameters $a, b$. In contrast, our approach utilizes a linear combination $s_{\rm Linear}(x) = s_1(x) + \beta s_2(x)$ with a single hyperparameter $\beta$, for which we derived an explicit formula, see \cref{equ:optimalScodSelector}. When comparing the two stragies in the \textit{tuned} setup, see \cref{ssec:experiments_methodology}, we search for the best hyperparameters $a,b,\beta$, optimizing the SCOD risk on the test set; providing an upper bound on the model’s performance. 

\boldparagraph{Hyperparameters of SIRC}
For practical deployment of $s_{\rm SIRC}$, a heuristic is provided by \cite{Xia-SIRC-ACCV2022} to set the parameters $a,b$ using the empirical mean $\mu_{s_2}$ and standard deviation $\sigma_{s_2}$ of the score $s_2(x)$ on ID data. Specifically, they propose setting $a_{\rm plug} = \mu_{s_2} - 3\sigma_{s_2}$, and $b_{\rm plug} = \nicefrac{1}{\sigma_{s_2}}$. When searching for the optimal parameters $a$ and $b$, we use these heuristic formulae as a rough estimate. We search over parameters $(a,b) \in [\mu_{s_2} - 6\sigma_{s_2},\; \mu_{s_2}]\times[\nicefrac{1}{10\sigma_{s_2}},\; \nicefrac{10}{\sigma_{s_2}}] = [a_{\rm plug} - 3\sigma_{s_2},\; a_{\rm plug} + 3\sigma_{s_2}]\times[\frac{1}{10}\cdot b_{\rm plug},\; 10\cdot b_{\rm plug}]$. We sample both intervals at $40$ evenly spaced values and at the heuristic values themselves; resulting in $41^2=1681$ possible hyperparameter settings.

\boldparagraph{Hyperparameter of linear strategy}
For the linear strategy, we derive an explicit formula $\beta_{\rm plug} = \frac{\relCost \,\tpr_{\rm min}}{1-\relCost}$, see \cref{equ:optimalScodSelector}. However, when searching for the optimal parameter $\beta$ on a test set, we parameterize the linear combination by an angle $\vartheta$, $s_{\rm Linear} = \cos{\vartheta}s_1(x) + \sin{\vartheta}s_2(x)$ and search over the interval $\vartheta \in [0, 2\pi]$. This is equivalent to the parameterization $\beta=\tan{\vartheta}$. The relative scale of the two scores is absorbed in the parameterization by the scale of the decision threshold $\lambda$. We sample $1600$ evenly spaced values from the interval, as well as $\frac{\pi}{2}, \pi, \frac{3\pi}{2}$; resulting in $1603$ possible hyperparameter settings.

\subsection{Evaluation Metrics}
\label{assec:metrics}
\boldparagraph{Definition}
To summarize the performance of a selective classifier $(h,c)$ where $c(x;\lambda)=\leftbb s(x)\leq \lambda \rightbb$, on the evaluation data $\ST$, we employ three metrics: \begin{enumerate}
    \item \textit{Area Under Risk - Coverage} ($\areaRiskCoverage$$\downarrow$) curve which measures the model's ability to discriminate between correctly classified (\cmark\:ID) and incorrectly classified (\xmark\:ID) data,
    
    \item \textit{Area Under Receiver Operating Characteristic} ($\areaROC\uparrow$) curve which evaluates how well the model distinguishes between ID and OOD data, and 

    \item \textit{Area Under SCOD Risk - True positive rate} ($\areaSCODTpr{}\downarrow$) curve which is used to evaluate the overall performance of the selective classifier on the SCOD problem, see Def. \ref{def:SCOD}. 
    
\end{enumerate}    

AuRC and AuROC are standard metrics for selective classification (SC) and out-of-distribution detection (OODD). AuSRT directly addresses the SCOD objective. Currently, single-score methods tend to address either SC or OODD; but not both at the same time. The SCOD problem addresses both tasks at the same time.

\boldparagraph{Empirical Estimates}
As we are only provided with a sample of data, we use the empirical estimates of the metrics. E.g., instead of TPR, we use 
$$
\esttpr(c) = \frac{1}{m}\sum_{(x,y)\in\ST_I}\,c(x).
$$
Analogously, for the FPR and the selective risk we use:
$$\estfpr(c) = \frac{1}{n}\sum_{(x,\emptyset)\in\ST_O}\,c(x)\,$$ and $$\estRsel(h,c) = \frac{\frac{1}{m}\sum_{(x,y)\in\ST_I}\,\ell(h(x),y)\,c(x)}{\esttpr(c)}.$$ 
The SCOD risk,~\cref{equ:SingeCostModel}, at a given TPR is estimated by
$$\estRscod(\tprMin;h,c) = \min_{\lambda}\,(1-\relCost)\,\estRsel(h,c)+\relCost\, \estfpr(c)\; \mbox{s.t.}\; \esttpr(c) \geq \tprMin,\:$$ 
where ${c(x;\lambda)=\leftbb s(x)\leq \lambda \rightbb}.$ The $\areaSCODTpr$ is then computed as the area under the curve given by points $\big\{\big(\estRscod(\tprMin), \tprMin\big)\mid \tprMin\in(0,1]\big\}$.

\clearpage
\section{Ablation}
\label{asec:ablation}
\boldparagraph{Relative cost ablation}
Our analysis on ImageNet, see~\cref{tab:changing_relative_cost}, reveals that the linear strategy using the plugin score,~\cref{equ:optimalScodSelector}, consistently performs better than SIRC and single-score strategies, across different values of the relative cost $\alpha\in[0,1]$. In other words, the linear strategy outperforms alternatives independent of the relative weights assigned to the OODD and SC tasks.

\begin{table}[htbp]
\centering
\caption{\areaSCODTpr{}$\downarrow$ in \% points for the \Algo{} approximated likelihood ratio (LR) when varying the relative cost $\alpha$. Results are shown for practically usable plugin double-score strategies on ID ImageNet.}
%\resizebox{\columnwidth}{!}{%
\begin{tabular}{l|r|c|c|c|c|c}
 & \backslashbox{\it Score}{\it $\alpha$} & $0.10$ & $0.25$ & $0.5$ & $0.75$ & $0.9$ \\ \hline
\multirow{2}{*}[-0.05cm]{\rotatebox{0}{ssb\_hard}} 
& SIRC%(MSP, -LR) 
& $\bf8.43$
& $10.62$ 
& $14.27$ 
& $17.93$ 
& $20.13$ 
\\ \cline{2-7}
& Linear%(1-MSP, LR)
& $8.44$ 
& $\bf7.13$ 
& $\bf6.37$ 
& $\bf7.22$ 
& $\bf9.97$ 
%\\ \cline{2-7}
%& Multiplication & $43.28$& $39.22$ & $32.47$ & $25.71$ & $21.66$ 
\\ \hline\hline
\multirow{2}{*}[-0.05cm]{\rotatebox{0}{ninco}} 
& SIRC%(MSP, -LR) 
& $\bf8.10$
& $9.81$ 
& $12.67$ 
& $15.53$ 
& $17.24$ 
\\ \cline{2-7}
& Linear%(1-MSP, LR)
& $8.32$
& $\bf7.78$ 
& $\bf7.44$ 
& $\bf7.66$ 
& $\bf8.39$ 
%\\ \cline{2-7}
%& Multiplication & $44.18$  & $42.0$ & $38.35$ & $34.71$ & $32.52$ 
\\ \hline\hline
\multirow{2}{*}[-0.05cm]{\rotatebox{0}{inaturalist}} 
& SIRC%(MSP, -LR) 
& $7.17$
& $7.49$ 
& $8.02$
& $8.56$ 
& $8.88$ 
\\ \cline{2-7}
& Linear%(1-MSP, LR)
& $\bf6.80$
& $\bf5.75$ 
& $\bf4.16$ 
& $\bf2.73$ 
& $\bf2.01$ 
%\\ \cline{2-7}
%& Multiplication & $42.02$  & $35.73$ & $25.25$ & $14.77$ & $8.48$ 
\\ \hline\hline
\multirow{2}{*}[-0.05cm]{\rotatebox{0}{textures}} 
& SIRC%(MSP, -LR) 
& $7.67$
& $8.73$ 
& $10.50$ 
& $12.29$ 
& $13.35$ 
\\ \cline{2-7}
& Linear%(1-MSP, LR)
& $\bf6.81$
& $\bf5.81$ 
& $\bf4.38$ 
& $\bf3.21$ 
& $\bf2.84$ 
%\\ \cline{2-7}
%& Multiplication & $42.37$ & $36.67$ & $27.17$ & $17.66$ & $11.96$ 
\\ \hline\hline
\multirow{2}{*}[-0.05cm]{\rotatebox{0}{openimage\_o}} 
& SIRC%(MSP, -LR) 
& $7.52$
& $8.35$ 
& $9.75$ 
& $11.15$ 
& $11.99$ 
\\ \cline{2-7}
& Linear%(1-MSP, LR)
& $\bf7.28$
& $\bf6.86$ 
& $\bf6.48$ 
& $\bf6.44$ 
& $\bf6.81$ 
%\\ \cline{2-7}
%& Multiplication & $44.75$ & $42.48$ & $38.69$ & $34.91$ & $32.64$ \\
\end{tabular}
%}
\label{tab:changing_relative_cost}
\end{table}

\boldparagraph{Prior $\pi_O^{\rm tr}$ ablation} In the main paper body, we demonstrate that \Algo{} is an effective approach for learning an estimate $\hat g(x)$ of the OOD/ID likelihood ratio when the unlabeled mixture $\mathcal{T}_U$ used for training contains an equal ratio of ID and OOD samples, i.e., $\pi_O^{\rm tr} \approx \nicefrac{1}{2}$. For other settings of $\pi_O^{\rm tr}$, we show results on ID ImageNet in~\cref{tab:changing_prior_p_o}. The linear strategy outperforms other approaches in a wide range of $\pi_O^{\rm tr}$. However, with $\pi_O^{\rm tr} \approx 0$, \Algo{} will likely fail, as it would be difficult to estimate $p_O(x)$ using $p(z=M\,|\,x)$.

Note that we use the test time ratio of OOD data $\pi_O^{\rm eval} \approx \nicefrac{1}{2}$ irrespective of the prior in the training mixture $\pi_O^{\rm tr}$. We can justify this, as the optimal SCOD strategy is independent of $\pi_O^{\rm eval}$, see~\cref{ssec:definitions}. The model can therefore be used in settings where $\pi_O^{\rm eval}$ is not stationary.

\begin{table}[htbp]
\centering
\caption{\areaSCODTpr{}$\downarrow$ in \% points for the \Algo{} approximated likelihood ratio when varying the portion of OOD samples $\pi_O^{\rm tr}$ in the mixture. Results are shown for practically usable plugin double-score strategies on ID ImageNet. The evaluation data is identical in all settings. The best single-score and double-score strategies using contemporary scores are also shown; note that they are selected based on test data performance. Settings where the linear combination of $\hat r(x)$ and $\hat g(x)$ outperforms all other strategies are marked in bold.}
%\resizebox{\columnwidth}{!}{%
\begin{tabular}{l|r|c|c|c|c|c||c|c}
 & \backslashbox{\it Score}{\it $\pi_O^{\rm tr}$} & $0.1$ & $0.2$ & $0.3$ & $0.4$ & $0.5$ & S. Score & D. Score \\ \hline
\multirow{2}{*}[-0.05cm]{\rotatebox{0}{ssb\_hard}} 
& SIRC%(MSP, -LR) 
& 17.36
& 17.34
& 16.84
& 16.80
& 8.63
& \multirow{2}{*}[-0.05cm]{\rotatebox{0}{17.43}} 
& \multirow{2}{*}[-0.05cm]{\rotatebox{0}{16.01}} 
\\ \cline{2-7}
& Linear%(1-MSP, LR)
& 17.20
& \bf13.91
& \bf10.94
& \bf9.67
& \bf5.88
&
&
\\ \hline
\multirow{2}{*}[-0.05cm]{\rotatebox{0}{ninco}} 
& SIRC%(MSP, -LR) 
& 13.27
& 13.26
& 12.79
& 13.56
& 11.16
& \multirow{2}{*}[-0.05cm]{\rotatebox{0}{13.50}} 
& \multirow{2}{*}[-0.05cm]{\rotatebox{0}{11.75}}
\\ \cline{2-7}
& Linear%(1-MSP, LR)
& 12.41
& 12.68
& \bf10.11
& \bf10.27
& \bf7.29
&
&
\\ \hline
\multirow{2}{*}[-0.05cm]{\rotatebox{0}{inaturalist}} 
& SIRC%(MSP, -LR) 
& 7.66
& 8.18
& 8.59
& 8.02
& 6.58
& \multirow{2}{*}[-0.05cm]{\rotatebox{0}{6.95}} 
& \multirow{2}{*}[-0.05cm]{\rotatebox{0}{5.53}}
\\ \cline{2-7}
& Linear%(1-MSP, LR)
& \bf4.54
& \bf4.36
& \bf4.30
& \bf4.27
& \bf4.12
&
&
\\ \hline
\multirow{2}{*}[-0.05cm]{\rotatebox{0}{textures}} 
& SIRC%(MSP, -LR) 
& 11.61
& 10.84
& 11.84
& 10.85
& 8.89
& \multirow{2}{*}[-0.05cm]{\rotatebox{0}{7.03}} 
& \multirow{2}{*}[-0.05cm]{\rotatebox{0}{5.91}}
\\ \cline{2-7}
& Linear%(1-MSP, LR)
& \bf5.52
& \bf4.98
& \bf4.83
& \bf4.60
& \bf4.24
&
&
\\ \hline
\multirow{2}{*}[-0.05cm]{\rotatebox{0}{openimage\_o}} 
& SIRC%(MSP, -LR) 
& 10.42
& 10.39
& 9.84
& 9.60
& 8.88
& \multirow{2}{*}[-0.05cm]{\rotatebox{0}{8.85}} 
& \multirow{2}{*}[-0.05cm]{\rotatebox{0}{7.41}}
\\ \cline{2-7}
& Linear%(1-MSP, LR)
& 9.11
& 7.68
& 7.67
& \bf7.17
& \bf6.25
&
&
\\ \hline
\end{tabular}
%}
\label{tab:changing_prior_p_o}
\end{table}

\clearpage
\section{Results on ImageNet-1K}
\label{asec:imagenet}
\begin{table}[!htbp]
\centering
\caption{The metrics defined in \cref{assec:metrics} shown in \% points for selective classifiers constructed from an ImageNet ID classifier $h(x)$ and selectors $c(x)=\leftbb s(x)\leq \lambda \rightbb$ for a representative sample of single-scores $s(x)$. Results are shown for in-distribution ImageNet using a relative cost of $\alpha=0.5$. The best results are marked in bold, with the second best underlined.}
\resizebox{\linewidth}{!}{%
\begin{tabular}{ll||c||c|c||c|c||c|c||c|c||c|c}
\multicolumn{2}{c||}{\multirow{2}{*}{\backslashbox{\it \footnotesize Score}{\it \footnotesize Dataset}}} & ImageNet & \multicolumn{2}{c||}{\small ssb\_hard} & \multicolumn{2}{c||}{\small ninco} & \multicolumn{2}{c||}{\small inaturalist} & \multicolumn{2}{c||}{\small textures} & \multicolumn{2}{c}{\small openimage\_o}\\
& & \tiny AuRC$\downarrow$ & \tiny AuROC$\uparrow$ & \tiny \areaSCODTpr$\downarrow$ & \tiny AuROC$\uparrow$ & \tiny \areaSCODTpr$\downarrow$ & \tiny AuROC$\uparrow$ & \tiny \areaSCODTpr$\downarrow$ & \tiny AuROC$\uparrow$ & \tiny \areaSCODTpr$\downarrow$ & \tiny AuROC$\uparrow$ & \tiny \areaSCODTpr$\downarrow$ \\ \hline
\multirow{9}{*}{\rotatebox{90}{\bf Single Score}} & \footnotesize ASH & \footnotesize11.32
& \footnotesize \underline{72.88} & \footnotesize19.03 & \footnotesize\bf83.44 & \footnotesize\underline{13.75} & \footnotesize\bf97.06 & \footnotesize\bf6.94 & \footnotesize\bf96.90 & \footnotesize\bf7.02 & \footnotesize\bf93.25 & \footnotesize\bf8.85 \\ \cline{2-13}
& \footnotesize EBO & \footnotesize11.27
&\footnotesize 72.07 &\footnotesize 19.59 &\footnotesize 79.70 &\footnotesize 15.78&\footnotesize 90.63 &\footnotesize 10.31&\footnotesize 88.70 &\footnotesize 11.28&\footnotesize 89.05 &\footnotesize 11.10\\ \cline{2-13}
& \footnotesize GradNorm & \footnotesize19.79
&\footnotesize 71.89 &\footnotesize 23.94 &\footnotesize 74.01 &\footnotesize 22.88&\footnotesize 93.89 &\footnotesize 12.95&\footnotesize 92.04 &\footnotesize 13.87 &\footnotesize 84.82 &\footnotesize 17.48\\ \cline{2-13}
& \footnotesize $L_1$-norm & \footnotesize25.02
&\footnotesize 60.47 &\footnotesize 32.27 &\footnotesize 52.89 &\footnotesize 36.06&\footnotesize 46.57 &\footnotesize 39.22&\footnotesize 68.27 &\footnotesize 28.37&\footnotesize 59.30 &\footnotesize 32.85\\ \cline{2-13}
& \footnotesize MLS & \footnotesize \underline{10.73}
&\footnotesize 72.50 &\footnotesize 19.11 &\footnotesize 80.40 &\footnotesize 15.16&\footnotesize 91.16 &\footnotesize 9.78&\footnotesize 88.39 &\footnotesize 11.17&\footnotesize 89.16 &\footnotesize 10.78\\ \cline{2-13}
& \footnotesize MSP & \footnotesize\bf6.95
&\footnotesize 72.09 &\footnotesize \bf17.43 &\footnotesize 79.95 &\footnotesize \bf13.50&\footnotesize 88.40 &\footnotesize 9.27&\footnotesize 82.43 &\footnotesize 12.26&\footnotesize 84.85 &\footnotesize 11.04\\ \cline{2-13}
& \footnotesize ODIN & \footnotesize11.32
&\footnotesize 71.74 &\footnotesize 19.79 &\footnotesize 77.76 &\footnotesize 16.77&\footnotesize 91.16 &\footnotesize 10.07&\footnotesize 89.00 &\footnotesize 11.15&\footnotesize 88.23 &\footnotesize 11.54\\ \cline{2-13}
& \footnotesize ReAct & \footnotesize10.79
&\footnotesize \bf73.02 &\footnotesize \underline{18.88} &\footnotesize \underline{81.72} &\footnotesize 14.53&\footnotesize \underline{96.34} &\footnotesize \underline{7.22}&\footnotesize \underline{92.78} &\footnotesize \underline{9.00}&\footnotesize \underline{91.86} &\footnotesize \underline{9.46}\\ \cline{2-13}
& \footnotesize Residual & \footnotesize24.75
&\footnotesize 43.31 &\footnotesize 40.72 &\footnotesize 52.80 &\footnotesize 35.97&\footnotesize 50.70 &\footnotesize 37.02&\footnotesize 87.57 &\footnotesize 18.59 &\footnotesize 60.90 &\footnotesize 31.92\\ 
\end{tabular}
}
\label{tab:imagenet_single_score}
\end{table}
\begin{table}[htbp]
\centering
\caption{\areaSCODTpr$\downarrow$ in \% points for \textit{tuned} selective classifiers constructed from an ID classifier $h(x)$ and selectors $c(x)=\leftbb s(x)\leq \lambda \rightbb$. Results are shown for ID ImageNet and several possible OOD datasets. Rows of the Table correspond to different scores $s(x)$. The relative cost is $\alpha = 0.5$. The best results per dataset are highlighted in green. The best results with contemporary OODD scores are shown in bold.}
\begin{tabular}{ll|c|c|c|c|c}
\multicolumn{2}{c|}{{\backslashbox{\it  Score}{\it  Dataset}}} &  ssb\_hard &  ninco &  inaturalist &  textures & openimage\_o\\ \hline
\multirow{9}{*}{\rotatebox{90}{\bf Single Score}} &  ASH \cite{Djurisic-ASH-ICLR2023}
& 19.04 & 13.76 & 6.95 & 7.03 & 8.85
\\ \cline{2-7}
&  EBO \cite{Weitang-EBO-NIPS2020}
& 19.60 & 15.78 & 10.32 & 11.28 & 11.11
\\ \cline{2-7}
&  GradNorm~\cite{Huang-Importance-NIPS2021}
& 23.95 & 22.89 & 12.95 & 13.88 & 17.49
\\ \cline{2-7}
&  $L_1$-norm \cite{Huang-Importance-NIPS2021}   
& 32.27 & 36.06 & 39.22 & 28.37 & 32.86
\\ \cline{2-7}
&  MLS  \cite{Hendrycks-MLS-ICML2022}      
& 19.12 & 15.17 & 9.79 & 11.17 & 10.79
\\ \cline{2-7}
&  MSP \cite{Hendrycks-baseline-ICLR17}            
& 17.43 & 13.50 & 9.27 & 12.26 & 11.05
\\ \cline{2-7}
&  ODIN \cite{liang2018enhancing}             
& 19.79 & 16.78 & 10.08 & 11.16 & 11.54
\\ \cline{2-7}
&  ReAct \cite{Sun-NIPS2021}             
& 18.88 & 14.53 & 7.22 & 9.00 & 9.46
\\ \cline{2-7}
&  Residual \cite{Wang-ViM-CVPR2022}      
& 40.72 & 35.98 & 37.03 & 18.59 & 31.93
\\ \hline
&  \Algo{} LR
& 14.25 & 16.67 & 12.70 & 13.19 & 15.65
\\ \cline{2-7}
&  Clean LR
& 14.14 & 14.04 & 11.22 & 11.75 & 14.00
\\
\hline\hline
\multirow{8}{*}{\rotatebox{90}{{\bf Linear} (MSP, -)}} 
& ASH \cite{Djurisic-ASH-ICLR2023}
& 16.81 & 11.86 & 5.57 & \bf5.91 & 7.45
\\ \cline{2-7}
&  EBO \cite{Weitang-EBO-NIPS2020}
& 17.02 & 12.95 & 8.26 & 9.98 & 9.42
\\ \cline{2-7}
&  GradNorm~\cite{Huang-Importance-NIPS2021}
& 17.36 & 13.43 & 7.74 & 9.60 & 10.39
\\ \cline{2-7}
&  $L_1$-norm \cite{Huang-Importance-NIPS2021}   
& 17.19 & 13.50 & 9.29 & 11.88 & 11.04
\\ \cline{2-7}
&  MLS  \cite{Hendrycks-MLS-ICML2022}      
& 17.03 & 12.95 & 8.25 & 10.08 & 9.43
\\ \cline{2-7}
&  ODIN \cite{liang2018enhancing}             
& 16.96 & 13.06 & 7.95 & 9.54 & 9.36
\\ \cline{2-7}
&  ReAct \cite{Sun-NIPS2021}             
& 16.65 & 12.26 & 5.92 & 7.99 & 8.06
\\ \cline{2-7}
&  Residual \cite{Wang-ViM-CVPR2022}      
& 17.42 & 13.22 & 8.99 & 6.65 & 10.16
\\ \hline
&  \Algo{} LR
& \cellcolor{\bestcolor}5.88 & 7.29 & 4.12 & 4.24 & 6.25
\\ \cline{2-7}
&  Clean LR
& \cellcolor{\bestcolor}5.88 & 6.67 & \cellcolor{\bestcolor}4.04 & \cellcolor{\bestcolor}4.23 & \cellcolor{\bestcolor}5.82
\\
\hline\hline
\multirow{8}{*}{\rotatebox{90}{{\bf SIRC} (MSP, -)}} & ASH \cite{Djurisic-ASH-ICLR2023}
& 16.73 & \bf11.75 & \bf5.53 & 6.09 & \bf7.41
\\ \cline{2-7}
&  EBO \cite{Weitang-EBO-NIPS2020}
& 17.01 & 12.93 & 8.23 & 10.01 & 9.41
\\ \cline{2-7}
&  GradNorm~\cite{Huang-Importance-NIPS2021}
& \bf16.01 & 12.33 & 6.42 & 7.99 & 8.78
\\ \cline{2-7}
&  $L_1$-norm \cite{Huang-Importance-NIPS2021}   
& 16.88 & 13.43 & 9.28 & 11.58 & 10.87
\\ \cline{2-7}
&  MLS  \cite{Hendrycks-MLS-ICML2022}      
& 17.02 & 12.93 & 8.23 & 10.14 & 9.44
\\ \cline{2-7}
&  ODIN \cite{liang2018enhancing}             
& 16.96 & 13.03 & 7.97 & 9.76 & 9.43
\\ \cline{2-7}
&  ReAct \cite{Sun-NIPS2021}             
& 16.60 & 12.23 & 5.91 & 8.12 & 8.06
\\ \cline{2-7}
&  Residual \cite{Wang-ViM-CVPR2022}      
& 17.41 & 13.13 & 8.89 & 6.22 & 9.90
\\ \hline
&  \Algo{} LR
& 8.63 & 11.16 & 6.58 & 8.89 & 8.88
\\ \cline{2-7}
&  Clean LR
& 6.59 & \cellcolor{\bestcolor}6.61 & 4.78 & 7.45 & 6.21
\end{tabular}
\label{tab:imagenet_posthoc_double_scores_tuned}
\end{table}

\begin{table}[htbp]
\centering
\caption{\areaSCODTpr$\downarrow$ in \% points for \textit{plugin} selective classifiers constructed from an ID classifier $h(x)$ and selectors $c(x)=\leftbb s(x)\leq \lambda \rightbb$. Results are shown for ID ImageNet and several possible OOD datasets. Rows of the Table correspond to different scores $s(x)$. The relative cost is $\alpha = 0.5$. The best results per dataset are highlighted in green. The best results with contemporary OODD scores are shown in bold.}
\begin{tabular}{ll|c|c|c|c|c}
\multicolumn{2}{c|}{{\backslashbox{\it  Score}{\it  Dataset}}} &  ssb\_hard &  ninco &  inaturalist &  textures & openimage\_o\\ \hline
\multirow{9}{*}{\rotatebox{90}{\bf Single Score}} &  ASH \cite{Djurisic-ASH-ICLR2023}
& 19.04 & 13.76 & 6.95 & 7.03 & 8.85
\\ \cline{2-7}
&  EBO \cite{Weitang-EBO-NIPS2020}
& 19.60 & 15.78 & 10.32 & 11.28 & 11.11
\\ \cline{2-7}
&  GradNorm~\cite{Huang-Importance-NIPS2021}
& 23.95 & 22.89 & 12.95 & 13.88 & 17.49
\\ \cline{2-7}
&  $L_1$-norm \cite{Huang-Importance-NIPS2021}   
& 32.27 & 36.06 & 39.22 & 28.37 & 32.86
\\ \cline{2-7}
&  MLS  \cite{Hendrycks-MLS-ICML2022}      
& 19.12 & 15.17 & 9.79 & 11.17 & 10.79
\\ \cline{2-7}
&  MSP \cite{Hendrycks-baseline-ICLR17}            
& 17.43 & 13.50 & 9.27 & 12.26 & 11.05
\\ \cline{2-7}
&  ODIN \cite{liang2018enhancing}             
& 19.79 & 16.78 & 10.08 & 11.16 & 11.54
\\ \cline{2-7}
&  ReAct \cite{Sun-NIPS2021}             
& 18.88 & 14.53 & 7.22 & 9.00 & 9.46
\\ \cline{2-7}
&  Residual \cite{Wang-ViM-CVPR2022}      
& 40.72 & 35.98 & 37.03 & 18.59 & 31.93
\\ \hline
&  \Algo{} LR
& 14.25 & 16.67 & 12.70 & 13.19 & 15.65
\\ \cline{2-7}
&  Clean LR
& 14.14 & 14.04 & 11.22 & 11.75 & 14.00
\\
\hline\hline
\multirow{8}{*}{\rotatebox{90}{{\bf Linear} (MSP, -)}} 
& ASH \cite{Djurisic-ASH-ICLR2023}
& 18.74 & 13.47 & 6.82 & \bf6.95 & \bf8.66
\\ \cline{2-7}
&  EBO \cite{Weitang-EBO-NIPS2020}
& 18.86 & 14.97 & 9.67 & 10.88 & 10.56
\\ \cline{2-7}
&  GradNorm~\cite{Huang-Importance-NIPS2021}
& 23.92 & 22.86 & 12.97 & 13.91 & 17.48
\\ \cline{2-7}
&  $L_1$-norm \cite{Huang-Importance-NIPS2021}   
& 32.11 & 35.87 & 39.01 & 28.20 & 32.67
\\ \cline{2-7}
&  MLS  \cite{Hendrycks-MLS-ICML2022}      
& 18.55 & 14.56 & 9.36 & 10.89 & 10.40
\\ \cline{2-7}
&  ODIN \cite{liang2018enhancing}             
& 17.40 & 13.51 & 9.27 & 12.21 & 11.02
\\ \cline{2-7}
&  ReAct \cite{Sun-NIPS2021}             
& 17.85 & 13.37 & \bf6.76 & 8.62 & 8.77
\\ \cline{2-7}
&  Residual \cite{Wang-ViM-CVPR2022}      
& 27.47 & 21.46 & 18.50 & 9.43 & 17.02
\\ \hline
&  \Algo{} LR
& 6.37 & \cellcolor{\bestcolor}7.44 & \cellcolor{\bestcolor}4.16 & \cellcolor{\bestcolor}4.38 & \cellcolor{\bestcolor}6.48
\\ \cline{2-7}
&  Clean LR
& \cellcolor{\bestcolor}6.10 & 8.05 & 4.35 & 4.48 & 7.38
\\
\hline\hline
\multirow{8}{*}{\rotatebox{90}{{\bf SIRC} (MSP, -)}} & ASH \cite{Djurisic-ASH-ICLR2023}
& 17.23 & 13.03 & 8.01 & 10.62 & 9.98
\\ \cline{2-7}
&  EBO \cite{Weitang-EBO-NIPS2020}
& 17.33 & 13.37 & 9.03 & 11.83 & 10.71
\\ \cline{2-7}
&  GradNorm~\cite{Huang-Importance-NIPS2021}
& \bf17.00 & \bf12.96 & 7.70 & 10.24 & 9.80
\\ \cline{2-7}
&  $L_1$-norm \cite{Huang-Importance-NIPS2021}   
& 17.21 & 13.54 & 9.50 & 11.90 & 10.96
\\ \cline{2-7}
&  MLS  \cite{Hendrycks-MLS-ICML2022}      
& 17.34 & 13.38 & 9.04 & 11.86 & 10.72
\\ \cline{2-7}
&  ODIN \cite{liang2018enhancing}             
& 17.38 & 13.45 & 9.16 & 12.10 & 10.92
\\ \cline{2-7}
&  ReAct \cite{Sun-NIPS2021}             
& 17.23 & 13.20 & 8.46 & 11.54 & 10.39
\\ \cline{2-7}
&  Residual \cite{Wang-ViM-CVPR2022}      
& 17.50 & 13.43 & 9.20 & 9.54 & 10.62
\\ \hline
&  \Algo{} LR
& 14.28 & 12.67 & 8.03 & 10.51 & 9.75
\\ \cline{2-7}
&  Clean LR
& 13.24 & 10.94 & 6.38 & 10.49 & 8.59
\end{tabular}
\label{tab:imagenet_posthoc_double_scores_plugin}
\end{table}

\clearpage
\section{Results on CIFAR-10}
\label{asec:cifar10}

\begin{table}[htbp]
\centering
\caption{The mean and standard deviation over 3-folds of metrics defined in~\cref{assec:metrics}. The results are shown in \% points for selective classifiers constructed from a CIFAR-10 ID classifier $h(x)$ and selectors $c(x)=\leftbb s(x)\leq \lambda \rightbb$ for a representative sample of single-scores $s(x)$. Results are shown for in-distribution CIFAR-10 using a relative cost of $\alpha=0.5$. The best results are marked in bold, with the second best underlined.}
\resizebox{\linewidth}{!}{%
\begin{tabular}{ll||c||c|c||c|c||c|c}
\multicolumn{2}{c||}{\multirow{2}{*}{\backslashbox{\it \footnotesize Score}{\it \footnotesize Dataset}}} & ID cifar10 & \multicolumn{2}{c||}{ cifar100} & \multicolumn{2}{c||}{ mnist} & \multicolumn{2}{c}{ places365} \\
& & \footnotesize AuRC$\downarrow$ & \footnotesize AuROC$\uparrow$ & \footnotesize AuSRT$\downarrow$ & \footnotesize AuROC$\uparrow$ & \footnotesize AuSRT$\downarrow$ & \footnotesize AuROC$\uparrow$ & \footnotesize AuSRT$\downarrow$ \\ \hline
\multirow{10}{*}{\rotatebox{90}{\bf Single Score}} 
& ASH 
& $1.84 \pm 0.27$ & $74.11 \pm 1.55$ & $13.87 \pm 0.89$ & $83.16 \pm 4.66$ & $9.34 \pm 2.35$ & $79.89 \pm 3.69$ & $10.97 \pm 1.71$
\\ \cline{2-9}
&  EBO            
& $0.90 \pm 0.10$ & $86.36 \pm 0.58$ & $7.27 \pm 0.33$ & $94.32 \pm 2.53$ & $3.29 \pm 1.22$ & $89.25 \pm 0.78$ & $5.82 \pm 0.35$
\\ \cline{2-9}
&  GradNorm            
& $3.89 \pm 0.38$ & $54.43 \pm 1.59$ & $24.73 \pm 0.92$ & $63.72 \pm 7.37$ & $20.08 \pm 3.54$ & $60.50 \pm 5.33$ & $21.69 \pm 2.47$
\\ \cline{2-9}
&  KNN         
& $\mathbf{0.56 \pm 0.04}$ & $\mathbf{89.73 \pm 0.14}$ & $\mathbf{5.41 \pm 0.09}$ & $94.26 \pm 0.38$ & $3.15 \pm 0.17$ & $\mathbf{91.77 \pm 0.23}$ & $\mathbf{4.40 \pm 0.11}$
\\ \cline{2-9}
&  MLS              
& $0.89 \pm 0.10$ & $86.31 \pm 0.59$ & $7.29 \pm 0.33$ & $94.15 \pm 2.48$ & $3.37 \pm 1.19$ & $89.14 \pm 0.76$ & $5.87 \pm 0.34$
\\ \cline{2-9}
& MSP
& $\underline{0.60 \pm 0.03}$ & $87.19 \pm 0.33$ & $6.70 \pm 0.18$ & $92.63 \pm 1.57$ & $3.98 \pm 0.77$ & $88.92 \pm 0.47$ & $5.84 \pm 0.22$
\\ \cline{2-9}
&  ODIN              
& $1.23 \pm 0.22$ & $82.18 \pm 1.87$ & $9.53 \pm 1.04$ & $\mathbf{95.24 \pm 1.96}$ & $\mathbf{3.00 \pm 0.91}$ & $85.07 \pm 1.24$ & $8.08 \pm 0.69$
\\ \cline{2-9}
&  ReAct              
& $0.86 \pm 0.11$ & $85.93 \pm 0.83$ & $7.46 \pm 0.46$ & $92.81 \pm 3.03$ & $4.02 \pm 1.49$ & $\underline{90.35 \pm 0.78}$ & $\underline{5.26 \pm 0.34}$
\\ \cline{2-9}
&  VIM
& $0.79 \pm 0.05$ & $\underline{87.75 \pm 0.28}$ & $\underline{6.52 \pm 0.16}$ & $\underline{94.76 \pm 0.38}$ & $\underline{3.02 \pm 0.19}$ & $89.49 \pm 0.39$ & $5.65 \pm 0.22$
\\ 
\end{tabular}
}

\vspace{1cm}

\resizebox{\linewidth}{!}{
\begin{tabular}{ll||c||c|c||c|c||c|c}
\multicolumn{2}{c||}{\multirow{2}{*}{\backslashbox{\it \footnotesize Score}{\it \footnotesize Dataset}}} & ID cifar10 & \multicolumn{2}{c||}{ svhn} & \multicolumn{2}{c||}{ textures} & \multicolumn{2}{c}{ tin}\\
& & \footnotesize AuRC$\downarrow$ & \footnotesize AuROC$\uparrow$ & \footnotesize AuSRT$\downarrow$ & \footnotesize AuROC$\uparrow$ & \footnotesize AuSRT$\downarrow$ & \footnotesize AuROC$\uparrow$ & \footnotesize AuSRT$\downarrow$ \\ \hline
\multirow{10}{*}{\rotatebox{90}{\bf Single Score}} 
& ASH 
& $1.84 \pm 0.27$ & $73.46 \pm 6.41$ & $14.19 \pm 3.28$ & $77.45 \pm 2.39$ & $12.20 \pm 1.33$ & $76.44 \pm 0.61$ & $12.70 \pm 0.44$
\\ \cline{2-9}
&  EBO            
& $0.90 \pm 0.10$ & $91.79 \pm 0.98$ & $4.56 \pm 0.45$ & $89.47 \pm 0.70$ & $5.72 \pm 0.40$ & $88.80 \pm 0.36$ & $6.05 \pm 0.22$
\\ \cline{2-9}
&  GradNorm            
& $3.89 \pm 0.38$ & $53.91 \pm 6.36$ & $24.99 \pm 3.23$ & $52.07 \pm 4.09$ & $25.91 \pm 2.23$ & $55.37 \pm 0.41$ & $24.26 \pm 0.39$
\\ \cline{2-9}
&  KNN         
& $\mathbf{0.56 \pm 0.04}$ & $\underline{92.67 \pm 0.30}$ & $\underline{3.94 \pm 0.13}$ & $\underline{93.16 \pm 0.24}$ & $\underline{3.70 \pm 0.14}$ & $\mathbf{91.56 \pm 0.26}$ & $\mathbf{4.50 \pm 0.15}$
\\ \cline{2-9}
&  MLS              
& $0.89 \pm 0.10$ & $91.69 \pm 0.94$ & $4.60 \pm 0.43$ & $89.41 \pm 0.71$ & $5.74 \pm 0.41$ & $88.72 \pm 0.36$ & $6.08 \pm 0.23$
\\ \cline{2-9}
& MSP
& $\underline{0.60 \pm 0.03}$ & $91.46 \pm 0.40$ & $4.57 \pm 0.19$ & $89.89 \pm 0.71$ & $5.36 \pm 0.37$ & $88.87 \pm 0.19$ & $5.86 \pm 0.11$
\\ \cline{2-9}
&  ODIN              
& $1.23 \pm 0.22$ & $84.58 \pm 0.77$ & $8.33 \pm 0.43$ & $86.94 \pm 2.26$ & $7.15 \pm 1.24$ & $83.55 \pm 1.84$ & $8.84 \pm 1.03$
\\ \cline{2-9}
&  ReAct              
& $0.86 \pm 0.11$ & $89.12 \pm 3.19$ & $5.87 \pm 1.59$ & $89.38 \pm 1.49$ & $5.74 \pm 0.80$ & $88.29 \pm 0.44$ & $6.29 \pm 0.28$
\\ \cline{2-9}
&  VIM
& $0.79 \pm 0.05$ & $\mathbf{94.51 \pm 0.48}$ & $\mathbf{3.14 \pm 0.22}$ & $\mathbf{95.16 \pm 0.34}$ & $\mathbf{2.82 \pm 0.19}$ & $\underline{89.62 \pm 0.33}$ & $\underline{5.59 \pm 0.19}$
\\ 
\end{tabular}
}

\end{table}

\begin{table}[htbp]
\centering
\caption{The mean and standard deviation over 3-folds of \areaSCODTpr$\downarrow$ in \% points for \textit{tuned} selective classifiers constructed from an ID classifier $h(x)$ and selectors $c(x)=\leftbb s(x)\leq \lambda \rightbb$. Results are shown for ID CIFAR-10 and several possible OOD datasets. Rows of the Table correspond to different scores $s(x)$. The relative cost is $\alpha = 0.5$. The best results per dataset are highlighted in green. The best results with contemporary OODD scores are shown in bold.}
\resizebox{\linewidth}{!}{%
\begin{tabular}{ll|c|c|c|c|c|c}
\multicolumn{2}{c|}{{\backslashbox{\it  Score}{\it  Dataset}}} &  cifar100 &  mnist &  places365 &  svhn & textures & tin\\ \hline
\multirow{9}{*}{\rotatebox{90}{\bf Single Score}} 
&  ASH \cite{Djurisic-ASH-ICLR2023} 
& $13.87 \pm 0.89$ & $9.34 \pm 2.35$ & $10.97 \pm 1.71$ & $14.19 \pm 3.28$ & $12.20 \pm 1.33$ & $12.70 \pm 0.44$
\\ \cline{2-8}
&  EBO \cite{Weitang-EBO-NIPS2020}            
& $7.27 \pm 0.33$ & $3.29 \pm 1.22$ & $5.82 \pm 0.35$ & $4.56 \pm 0.45$ & $5.72 \pm 0.40$ & $6.05 \pm 0.22$
\\ \cline{2-8}
&  GradNorm~\cite{Huang-Importance-NIPS2021}             
& $24.73 \pm 0.92$ & $20.08 \pm 3.54$ & $21.69 \pm 2.47$ & $24.99 \pm 3.23$ & $25.91 \pm 2.23$ & $24.26 \pm 0.39$
\\ \cline{2-8}
&  KNN~\cite{pmlr-v162-sun22d}
& $5.41 \pm 0.09$ & $3.15 \pm 0.17$ & $\mathbf{4.40 \pm 0.11}$ & $3.94 \pm 0.13$ & $3.70 \pm 0.14$ & $\mathbf{4.50 \pm 0.15}$
\\ \cline{2-8}
&  MLS  \cite{Hendrycks-MLS-ICML2022}                 
& $7.29 \pm 0.33$ & $3.37 \pm 1.19$ & $5.87 \pm 0.34$ & $4.60 \pm 0.43$ & $5.74 \pm 0.41$ & $6.08 \pm 0.23$
\\ \cline{2-8}
& MSP \cite{Hendrycks-baseline-ICLR17}  
& $6.70 \pm 0.18$ & $3.98 \pm 0.77$ & $5.84 \pm 0.22$ & $4.57 \pm 0.19$ & $5.36 \pm 0.37$ & $5.86 \pm 0.11$
\\ \cline{2-8}
&  ODIN \cite{liang2018enhancing}               
& $9.53 \pm 1.04$ & $3.00 \pm 0.91$ & $8.08 \pm 0.69$ & $8.33 \pm 0.43$ & $7.15 \pm 1.24$ & $8.84 \pm 1.03$
\\ \cline{2-8}
&  ReAct \cite{Sun-NIPS2021}                
& $7.46 \pm 0.46$ & $4.02 \pm 1.49$ & $5.26 \pm 0.34$ & $5.87 \pm 1.59$ & $5.74 \pm 0.80$ & $6.29 \pm 0.28$
\\ \cline{2-8}
%&  Residual           
%& $52.17 \pm 0.17$ & $52.17 \pm 0.17$ & $52.17 \pm 0.17$ & $52.17 \pm 0.17$ & $52.17 \pm 0.17$ & $52.17 \pm 0.17$
%\\ \cline{2-8}
&  VIM~\cite{Wang-ViM-CVPR2022}           
& $6.52 \pm 0.16$ & $3.02 \pm 0.19$ & $5.65 \pm 0.22$ & $3.14 \pm 0.22$ & $2.82 \pm 0.19$ & $5.59 \pm 0.19$
\\ \hline
& \Algo{} LR
& $10.80 \pm 0.11$ & $2.17 \pm 0.17$ & $7.89 \pm 0.95$ & $2.25 \pm 0.12$ & $3.97 \pm 0.30$ & $2.73 \pm 0.07$
\\
\cline{2-8}
&  Clean LR
& $4.24 \pm 0.18$ & $2.17 \pm 0.16$ & $5.99 \pm 0.13$ & $2.14 \pm 0.13$ & $3.01 \pm 0.16$ & $2.31 \pm 0.18$
\\
\hline\hline
\multirow{8}{*}{\rotatebox{90}{{\bf Linear} (MSP, -)}} 
&  ASH \cite{Djurisic-ASH-ICLR2023} 
& $6.60 \pm 0.20$ & $3.79 \pm 0.79$ & $5.63 \pm 0.28$ & $4.51 \pm 0.26$ & $5.25 \pm 0.38$ & $5.75 \pm 0.12$ 
\\ \cline{2-8}
&  EBO \cite{Weitang-EBO-NIPS2020}            
& $6.28 \pm 0.18$ & $3.03 \pm 1.03$ & $5.20 \pm 0.27$ & $3.96 \pm 0.30$ & $4.84 \pm 0.34$ & $5.31 \pm 0.12$
\\ \cline{2-8}
&  GradNorm~\cite{Huang-Importance-NIPS2021}             
& $6.69 \pm 0.18$ & $3.93 \pm 0.81$ & $5.80 \pm 0.24$ & $4.55 \pm 0.22$ & $5.35 \pm 0.37$ & $5.85 \pm 0.11$
\\ \cline{2-8}
&  KNN~\cite{pmlr-v162-sun22d}
& $\mathbf{5.40 \pm 0.08}$ & $3.12 \pm 0.19$ & $\mathbf{4.40 \pm 0.11}$ & $3.91 \pm 0.13$ & $3.70 \pm 0.14$ & $\mathbf{4.50 \pm 0.15}$
\\ \cline{2-8}
&  MLS  \cite{Hendrycks-MLS-ICML2022}                 
& $6.31 \pm 0.18$ & $3.11 \pm 1.01$ & $5.25 \pm 0.27$ & $4.01 \pm 0.28$ & $4.88 \pm 0.34$ & $5.36 \pm 0.12$
\\ \cline{2-8}
&  ODIN \cite{liang2018enhancing}               
& $6.52 \pm 0.21$ & $\mathbf{2.67 \pm 0.81}$ & $5.63 \pm 0.22$ & $4.48 \pm 0.16$ & $4.80 \pm 0.39$ & $5.73 \pm 0.14$
\\ \cline{2-8}
&  ReAct \cite{Sun-NIPS2021}                
& $6.33 \pm 0.23$ & $3.31 \pm 1.00$ & $5.00 \pm 0.22$ & $4.24 \pm 0.43$ & $4.86 \pm 0.45$ & $5.38 \pm 0.17$
\\ \cline{2-8}
%&  Residual           
%& $6.71 \pm 0.18$ & $3.99 \pm 0.77$ & $5.85 \pm 0.22$ & $4.58 \pm 0.20$ & $5.36 \pm 0.36$ & $5.87 \pm 0.11$
%\\ \cline{2-8}
&  VIM~\cite{Wang-ViM-CVPR2022}
& $6.15 \pm 0.18$ & $2.88 \pm 0.11$ & $5.29 \pm 0.09$ & $\mathbf{3.10 \pm 0.21}$ & $\mathbf{2.81 \pm 0.19}$ & $5.28 \pm 0.17$
\\ \hline
& \Algo{} LR
& $5.07 \pm 0.10$ & \cellcolor{\bestcolor}$0.29 \pm 0.02$ & $3.75 \pm 0.46$ & $0.34 \pm 0.02$ & $1.55 \pm 0.12$ & $0.98 \pm 0.05$
\\
\cline{2-8}
&  Clean LR
& \cellcolor{\bestcolor}$3.02 \pm 0.06$ & \cellcolor{\bestcolor}$0.29 \pm 0.02$ & \cellcolor{\bestcolor}$3.20 \pm 0.05$ & \cellcolor{\bestcolor}$0.30 \pm 0.02$ & \cellcolor{\bestcolor}$1.15 \pm 0.11$ & \cellcolor{\bestcolor}$0.42 \pm 0.00$
\\
\hline\hline
\multirow{8}{*}{\rotatebox{90}{{\bf SIRC} (MSP, -)}} 
&  ASH \cite{Djurisic-ASH-ICLR2023} 
& $6.57 \pm 0.19$ & $3.77 \pm 0.80$ & $5.51 \pm 0.39$ & $4.51 \pm 0.26$ & $5.24 \pm 0.38$ & $5.73 \pm 0.11$
\\ \cline{2-8}
&  EBO \cite{Weitang-EBO-NIPS2020}            
& $6.30 \pm 0.18$ & $3.10 \pm 1.01$ & $5.23 \pm 0.27$ & $3.99 \pm 0.29$ & $4.87 \pm 0.34$ & $5.34 \pm 0.12$
\\ \cline{2-8}
&  GradNorm~\cite{Huang-Importance-NIPS2021}             
& $6.66 \pm 0.18$ & $3.89 \pm 0.82$ & $5.72 \pm 0.29$ & $4.53 \pm 0.22$ & $5.33 \pm 0.38$ & $5.83 \pm 0.11$
\\ \cline{2-8}
&  KNN~\cite{pmlr-v162-sun22d}          
& $5.94 \pm 0.13$ & $3.35 \pm 0.40$ & $4.91 \pm 0.14$ & $4.16 \pm 0.19$ & $4.17 \pm 0.24$ & $5.03 \pm 0.12$
\\ \cline{2-8}
&  MLS  \cite{Hendrycks-MLS-ICML2022}                 
& $6.33 \pm 0.19$ & $3.18 \pm 0.98$ & $5.29 \pm 0.26$ & $4.04 \pm 0.27$ & $4.91 \pm 0.35$ & $5.38 \pm 0.12$
\\ \cline{2-8}
&  ODIN \cite{liang2018enhancing}               
& $6.51 \pm 0.21$ & $2.86 \pm 0.82$ & $5.63 \pm 0.22$ & $4.47 \pm 0.16$ & $4.81 \pm 0.38$ & $5.73 \pm 0.14$
\\ \cline{2-8}
&  ReAct \cite{Sun-NIPS2021}                
& $6.33 \pm 0.23$ & $3.35 \pm 0.97$ & $5.08 \pm 0.24$ & $4.25 \pm 0.41$ & $4.88 \pm 0.45$ & $5.40 \pm 0.16$
\\ \cline{2-8}
%&  Residual           
%& $6.71 \pm 0.18$ & $3.99 \pm 0.77$ & $5.85 \pm 0.22$ & $4.58 \pm 0.20$ & $5.36 \pm 0.36$ & $5.87 \pm 0.11$
%\\ \cline{2-8}
&  VIM~\cite{Wang-ViM-CVPR2022}
& $6.14 \pm 0.16$ & $2.92 \pm 0.20$ & $5.27 \pm 0.02$ & $3.17 \pm 0.23$ & $2.93 \pm 0.23$ & $5.30 \pm 0.14$
\\ \hline
& \Algo{} LR
& $5.09 \pm 0.23$ & \cellcolor{\bestcolor}$0.29 \pm 0.02$ & $5.04 \pm 0.18$ & $0.36 \pm 0.03$ & $1.81 \pm 0.30$ & $1.72 \pm 0.28$
\\
\cline{2-8}
&  Clean LR
& $3.89 \pm 0.61$ & \cellcolor{\bestcolor}$0.29 \pm 0.02$ & $5.40 \pm 0.18$ & $0.32 \pm 0.02$ & $1.78 \pm 0.12$ & $0.66 \pm 0.02$
\\
\end{tabular}
}
\end{table}

\begin{table}[htbp]
\centering
\caption{The mean and standard deviation over 3-folds \areaSCODTpr$\downarrow$ in \% points for \textit{plugin} selective classifiers constructed from an ID classifier $h(x)$ and selectors $c(x)=\leftbb s(x)\leq \lambda \rightbb$. Results are shown for ID CIFAR-10 and several possible OOD datasets. Rows of the Table correspond to different scores $s(x)$. The relative cost is $\alpha = 0.5$. The best results per dataset are highlighted in green. The best results with contemporary OODD scores are shown in bold.}
\resizebox{\linewidth}{!}{%
\begin{tabular}{ll|c|c|c|c|c|c}
\multicolumn{2}{c|}{{\backslashbox{\it  Score}{\it  Dataset}}} &  cifar100 &  mnist &  places365 &  svhn & textures & tin\\ \hline
\multirow{9}{*}{\rotatebox{90}{\bf Single Score}} 
&  ASH \cite{Djurisic-ASH-ICLR2023} 
& $13.87 \pm 0.89$ & $9.34 \pm 2.35$ & $10.97 \pm 1.71$ & $14.19 \pm 3.28$ & $12.20 \pm 1.33$ & $12.70 \pm 0.44$
\\ \cline{2-8}
&  EBO \cite{Weitang-EBO-NIPS2020}            
& $7.27 \pm 0.33$ & $3.29 \pm 1.22$ & $5.82 \pm 0.35$ & $4.56 \pm 0.45$ & $5.72 \pm 0.40$ & $6.05 \pm 0.22$
\\ \cline{2-8}
&  GradNorm~\cite{Huang-Importance-NIPS2021}             
& $24.73 \pm 0.92$ & $20.08 \pm 3.54$ & $21.69 \pm 2.47$ & $24.99 \pm 3.23$ & $25.91 \pm 2.23$ & $24.26 \pm 0.39$
\\ \cline{2-8}
&  KNN~\cite{pmlr-v162-sun22d}          
& $\mathbf{5.41 \pm 0.09}$ & $3.15 \pm 0.17$ & $\mathbf{4.40 \pm 0.11}$ & $3.94 \pm 0.13$ & $3.70 \pm 0.14$ & $\mathbf{4.50 \pm 0.15}$
\\ \cline{2-8}
&  MLS  \cite{Hendrycks-MLS-ICML2022}                 
& $7.29 \pm 0.33$ & $3.37 \pm 1.19$ & $5.87 \pm 0.34$ & $4.60 \pm 0.43$ & $5.74 \pm 0.41$ & $6.08 \pm 0.23$
\\ \cline{2-8}
& MSP \cite{Hendrycks-baseline-ICLR17}  
& $6.70 \pm 0.18$ & $3.98 \pm 0.77$ & $5.84 \pm 0.22$ & $4.57 \pm 0.19$ & $5.36 \pm 0.37$ & $5.86 \pm 0.11$
\\ \cline{2-8}
&  ODIN \cite{liang2018enhancing}               
& $9.53 \pm 1.04$ & $\mathbf{3.00 \pm 0.91}$ & $8.08 \pm 0.69$ & $8.33 \pm 0.43$ & $7.15 \pm 1.24$ & $8.84 \pm 1.03$
\\ \cline{2-8}
&  ReAct \cite{Sun-NIPS2021}                
& $7.46 \pm 0.46$ & $4.02 \pm 1.49$ & $5.26 \pm 0.34$ & $5.87 \pm 1.59$ & $5.74 \pm 0.80$ & $6.29 \pm 0.28$
\\ \cline{2-8}
%&  Residual           
%& $52.17 \pm 0.17$ & $52.17 \pm 0.17$ & $52.17 \pm 0.17$ & $52.17 \pm 0.17$ & $52.17 \pm 0.17$ & $52.17 \pm 0.17$
%\\ \cline{2-8}
&  VIM~\cite{Wang-ViM-CVPR2022}        
& $6.52 \pm 0.16$ & $3.02 \pm 0.19$ & $5.65 \pm 0.22$ & $\mathbf{3.14 \pm 0.22}$ & $\mathbf{2.82 \pm 0.19}$ & $5.59 \pm 0.19$
\\ \hline
& \Algo{} LR
& $10.80 \pm 0.11$ & $2.17 \pm 0.17$ & $7.89 \pm 0.95$ & $2.25 \pm 0.12$ & $3.97 \pm 0.30$ & $2.73 \pm 0.07$
\\
\cline{2-8}
&  Clean LR
& $4.24 \pm 0.18$ & $2.17 \pm 0.16$ & $5.99 \pm 0.13$ & $2.14 \pm 0.13$ & $3.01 \pm 0.16$ & $2.31 \pm 0.18$
\\
\hline\hline
\multirow{8}{*}{\rotatebox{90}{{\bf Linear} (MSP, -)}} 
&  ASH \cite{Djurisic-ASH-ICLR2023} 
& $13.60 \pm 0.84$ & $9.13 \pm 2.30$ & $10.78 \pm 1.68$ & $13.69 \pm 3.07$ & $11.91 \pm 1.30$ & $12.43 \pm 0.41$
\\ \cline{2-8}
&  EBO \cite{Weitang-EBO-NIPS2020}            
& $7.28 \pm 0.33$ & $3.37 \pm 1.20$ & $5.87 \pm 0.35$ & $4.59 \pm 0.44$ & $5.74 \pm 0.40$ & $6.08 \pm 0.22$
\\ \cline{2-8}
&  GradNorm~\cite{Huang-Importance-NIPS2021}             
& $24.71 \pm 0.91$ & $20.07 \pm 3.53$ & $21.68 \pm 2.47$ & $24.97 \pm 3.22$ & $25.89 \pm 2.22$ & $24.24 \pm 0.39$
\\ \cline{2-8}
&  KNN~\cite{pmlr-v162-sun22d}          
& $5.69 \pm 0.09$ & $3.45 \pm 0.30$ & $4.79 \pm 0.16$ & $4.06 \pm 0.10$ & $4.11 \pm 0.15$ & $4.84 \pm 0.16$
\\ \cline{2-8}
&  MLS  \cite{Hendrycks-MLS-ICML2022}                 
& $7.31 \pm 0.34$ & $3.45 \pm 1.17$ & $5.92 \pm 0.34$ & $4.64 \pm 0.42$ & $5.77 \pm 0.41$ & $6.12 \pm 0.23$
\\ \cline{2-8}
&  ODIN \cite{liang2018enhancing}               
& $6.79 \pm 0.21$ & $3.87 \pm 0.80$ & $5.93 \pm 0.23$ & $4.64 \pm 0.20$ & $5.32 \pm 0.39$ & $5.96 \pm 0.12$
\\ \cline{2-8}
&  ReAct \cite{Sun-NIPS2021}                
& $7.49 \pm 0.44$ & $4.11 \pm 1.45$ & $5.34 \pm 0.34$ & $5.87 \pm 1.51$ & $5.79 \pm 0.77$ & $6.33 \pm 0.26$
\\ \cline{2-8}
%&  Residual           
%& $6.71 \pm 0.18$ & $4.00 \pm 0.77$ & $5.85 \pm 0.22$ & $4.58 \pm 0.20$ & $5.36 \pm 0.36$ & $5.87 \pm 0.11$
%\\ \cline{2-8}
&  VIM~\cite{Wang-ViM-CVPR2022}           
& $6.49 \pm 0.17$ & $\mathbf{3.00 \pm 0.17}$ & $5.61 \pm 0.21$ & $3.16 \pm 0.21$ & $2.86 \pm 0.19$ & $5.55 \pm 0.18$ 
\\ \hline
& \Algo{} LR
& $8.93 \pm 0.34$ & \cellcolor{\bestcolor}$0.30 \pm 0.02$ & $4.30 \pm 0.38$ & $0.37 \pm 0.03$ & $1.73 \pm 0.12$ & $1.20 \pm 0.08$
\\
\cline{2-8}
&  Clean LR
& \cellcolor{\bestcolor} $3.66 \pm 0.12$ & \cellcolor{\bestcolor}$0.30 \pm 0.02$ & \cellcolor{\bestcolor}$4.16 \pm 0.07$ & \cellcolor{\bestcolor}$0.32 \pm 0.02$ & \cellcolor{\bestcolor}$1.25 \pm 0.12$ & \cellcolor{\bestcolor}$0.48 \pm 0.0$
\\
\hline\hline
\multirow{8}{*}{\rotatebox{90}{{\bf SIRC} (MSP, -)}} 
&  ASH \cite{Djurisic-ASH-ICLR2023} 
& $6.70 \pm 0.18$ & $3.94 \pm 0.80$ & $5.79 \pm 0.26$ & $4.61 \pm 0.24$ & $5.36 \pm 0.39$ & $5.86 \pm 0.11$
\\ \cline{2-8}
&  EBO \cite{Weitang-EBO-NIPS2020}            
& $6.61 \pm 0.18$ & $3.76 \pm 0.85$ & $5.70 \pm 0.23$ & $4.43 \pm 0.20$ & $5.24 \pm 0.37$ & $5.74 \pm 0.11$
\\ \cline{2-8}
&  GradNorm~\cite{Huang-Importance-NIPS2021}             
& $6.76 \pm 0.18$ & $3.99 \pm 0.81$ & $5.86 \pm 0.25$ & $4.62 \pm 0.21$ & $5.42 \pm 0.40$ & $5.91 \pm 0.12$
\\ \cline{2-8}
&  KNN~\cite{pmlr-v162-sun22d}          
& $6.53 \pm 0.18$ & $3.81 \pm 0.71$ & $5.61 \pm 0.21$ & $4.48 \pm 0.20$ & $5.12 \pm 0.36$ & $5.67 \pm 0.11$
\\ \cline{2-8}
&  MLS  \cite{Hendrycks-MLS-ICML2022}                 
& $6.62 \pm 0.18$ & $3.77 \pm 0.85$ & $5.71 \pm 0.23$ & $4.44 \pm 0.20$ & $5.24 \pm 0.37$ & $5.75 \pm 0.11$
\\ \cline{2-8}
&  ODIN \cite{liang2018enhancing}               
& $6.62 \pm 0.19$ & $3.49 \pm 0.82$ & $5.77 \pm 0.22$ & $4.52 \pm 0.18$ & $5.08 \pm 0.37$ & $5.81 \pm 0.12$
\\ \cline{2-8}
&  ReAct \cite{Sun-NIPS2021}                
& $6.62 \pm 0.19$ & $3.81 \pm 0.85$ & $5.68 \pm 0.22$ & $4.49 \pm 0.26$ & $5.24 \pm 0.39$ & $5.75 \pm 0.12$
\\ \cline{2-8}
%&  Residual           
%& $6.70 \pm 0.18$ & $3.98 \pm 0.77$ & $5.84 \pm 0.22$ & $4.57 \pm 0.19$ & $5.36 \pm 0.37$ & $5.86 \pm 0.11$
%\\ \cline{2-8}
&  VIM~\cite{Wang-ViM-CVPR2022}           
& $6.57 \pm 0.18$ & $3.78 \pm 0.67$ & $5.71 \pm 0.18$ & $4.29 \pm 0.25$ & $4.59 \pm 0.39$ & $5.73 \pm 0.12$
\\ \hline
& \Algo{} LR
& $6.02 \pm 0.25$ & \cellcolor{\bestcolor}$0.30 \pm 0.02$ & $5.72 \pm 0.22$ & $0.44 \pm 0.06$ & $2.76 \pm 0.37$ & $3.40 \pm 0.49$
\\
\cline{2-8}
&  Clean LR
& $5.28 \pm 0.57$ & \cellcolor{\bestcolor}$0.30 \pm 0.02$ & $5.73 \pm 0.2$ & $0.36 \pm 0.03$ & $2.83 \pm 0.17$ & $1.26 \pm 0.07$
\\
\end{tabular}
}
\end{table}

\clearpage

\section{Results on CIFAR-100}
\label{asec:cifar100}

\begin{table}[htbp]
\centering
\caption{The mean and standard deviation over 3-folds of metrics defined in~\cref{assec:metrics}. Results are shown in \% points for selective classifiers constructed from a CIFAR-100 ID classifier $h(x)$ and selectors $c(x)=\leftbb s(x)\leq \lambda \rightbb$ for a representative sample of single-scores $s(x)$. Results are shown for in-distribution CIFAR-100 using a relative cost of $\alpha=0.5$. The best results are marked in bold, with the second best underlined.}
\resizebox{\linewidth}{!}{%
\begin{tabular}{ll||c||c|c||c|c||c|c}
\multicolumn{2}{c||}{\multirow{2}{*}{\backslashbox{\it \footnotesize Score}{\it \footnotesize Dataset}}} & ID cifar100 & \multicolumn{2}{c||}{ cifar10} & \multicolumn{2}{c||}{ mnist} & \multicolumn{2}{c||}{ places365} \\
& & \footnotesize AuRC$\downarrow$ & \footnotesize AuROC$\uparrow$ & \footnotesize AuSRT$\downarrow$ & \footnotesize AuROC$\uparrow$ & \footnotesize AuSRT$\downarrow$ & \footnotesize AuROC$\uparrow$ & \footnotesize AuSRT$\downarrow$ \\ \hline
\multirow{10}{*}{\rotatebox{90}{\bf Single Score}} 
& ASH 
& $8.34 \pm 0.20$ & $76.48 \pm 0.30$ & $15.93 \pm 0.24$ & $77.23 \pm 0.46$ & $15.56 \pm 0.13$ & $78.76 \pm 0.16$ & $14.79 \pm 0.05$
\\ \cline{2-9}
&  EBO            
& $7.83 \pm 0.06$ & $\underline{79.05 \pm 0.11}$ & $14.39 \pm 0.07$ & $79.18 \pm 1.37$ & $14.33 \pm 0.67$ & $79.52 \pm 0.23$ & $14.16 \pm 0.13$
\\ \cline{2-9}
&  GradNorm            
& $14.17 \pm 0.18$ & $70.32 \pm 0.20$ & $21.92 \pm 0.16$ & $65.35 \pm 1.12$ & $24.41 \pm 0.48$ & $69.69 \pm 0.17$ & $22.24 \pm 0.14$
\\ \cline{2-9}
&  KNN         
& $\underline{7.28 \pm 0.12}$ & $77.02 \pm 0.25$ & $15.13 \pm 0.08$ & $\underline{82.36 \pm 1.52}$ & $\underline{12.46 \pm 0.79}$ & $79.43 \pm 0.47$ & $13.93 \pm 0.18$
\\ \cline{2-9}
&  MLS              
& $7.59 \pm 0.07$ & $\mathbf{79.21 \pm 0.10}$ & $\underline{14.19 \pm 0.08}$ & $78.91 \pm 1.47$ & $14.34 \pm 0.72$ & $\underline{79.75 \pm 0.24}$ & $13.92 \pm 0.14$
\\ \cline{2-9}
& MSP
& $\mathbf{6.19 \pm 0.12}$ & $78.47 \pm 0.07$ & $\mathbf{13.86 \pm 0.09}$ & $76.08 \pm 1.86$ & $15.06 \pm 0.93$ & $79.22 \pm 0.29$ & $\mathbf{13.49 \pm 0.16}$
\\ \cline{2-9}
&  ODIN              
& $8.13 \pm 0.02$ & $78.18 \pm 0.14$ & $14.98 \pm 0.07$ & $\mathbf{83.79 \pm 1.31}$ & $\mathbf{12.17 \pm 0.64}$ & $79.45 \pm 0.26$ & $14.34 \pm 0.14$
\\ \cline{2-9}
&  ReAct              
& $7.66 \pm 0.02$ & $78.65 \pm 0.05$ & $14.50 \pm 0.02$ & $78.37 \pm 1.59$ & $14.65 \pm 0.79$ & $\mathbf{80.03 \pm 0.11}$ & $\underline{13.82 \pm 0.06}$
\\ \cline{2-9}
& VIM
& $8.79 \pm 0.18$ & $72.21 \pm 0.41$ & $18.29 \pm 0.21$ & $81.89 \pm 1.02$ & $13.45 \pm 0.59$ & $75.85 \pm 0.37$ & $16.47 \pm 0.10$
\\ 
\end{tabular}
}

\vspace{1cm}

\resizebox{\linewidth}{!}{%
\begin{tabular}{ll||c||c|c||c|c||c|c}
\multicolumn{2}{c||}{\multirow{2}{*}{\backslashbox{\it \footnotesize Score}{\it \footnotesize Dataset}}} & ID cifar100 & \multicolumn{2}{c||}{ svhn} & \multicolumn{2}{c||}{ textures} & \multicolumn{2}{c}{ tin}\\
& & \footnotesize AuRC$\downarrow$ & \footnotesize AuROC$\uparrow$ & \footnotesize AuSRT$\downarrow$ & \footnotesize AuROC$\uparrow$ & \footnotesize AuSRT$\downarrow$ & \footnotesize AuROC$\uparrow$ & \footnotesize AuSRT$\downarrow$ \\ \hline
\multirow{10}{*}{\rotatebox{90}{\bf Single Score}} 
& ASH 
& $8.34 \pm 0.20$ & $\mathbf{85.60 \pm 1.40}$ & $\mathbf{11.37 \pm 0.60}$ & $80.72 \pm 0.70$ & $13.81 \pm 0.25$ & $79.92 \pm 0.20$ & $14.21 \pm 0.17$
\\ \cline{2-9}
&  EBO            
& $7.83 \pm 0.06$ & $82.03 \pm 1.74$ & $12.90 \pm 0.90$ & $78.35 \pm 0.83$ & $14.74 \pm 0.45$ & $82.76 \pm 0.08$ & $12.54 \pm 0.04$
\\ \cline{2-9}
&  GradNorm            
& $14.17 \pm 0.18$ & $76.95 \pm 4.73$ & $18.61 \pm 2.40$ & $64.58 \pm 0.13$ & $24.79 \pm 0.15$ & $69.95 \pm 0.79$ & $22.11 \pm 0.38$
\\ \cline{2-9}
&  KNN         
& $\underline{7.28 \pm 0.12}$ & $\underline{84.15 \pm 1.09}$ & $\underline{11.57 \pm 0.48}$ & $\underline{83.66 \pm 0.83}$ & $\underline{11.81 \pm 0.44}$ & $\mathbf{83.34 \pm 0.16}$ & $\mathbf{11.97 \pm 0.07}$
\\ \cline{2-9}
&  MLS              
& $7.59 \pm 0.07$ & $81.65 \pm 1.49$ & $12.97 \pm 0.78$ & $78.39 \pm 0.84$ & $14.60 \pm 0.46$ & $\underline{82.90 \pm 0.05}$ & $12.35 \pm 0.05$
\\ \cline{2-9}
& MSP
& $\mathbf{6.19 \pm 0.12}$ & $78.42 \pm 0.89$ & $13.89 \pm 0.51$ & $77.32 \pm 0.71$ & $14.44 \pm 0.41$ & $82.07 \pm 0.17$ & $\underline{12.06 \pm 0.13}$
\\ \cline{2-9}
&  ODIN              
& $8.13 \pm 0.02$ & $74.54 \pm 0.76$ & $16.80 \pm 0.39$ & $79.33 \pm 1.08$ & $14.40 \pm 0.55$ & $81.63 \pm 0.08$ & $13.25 \pm 0.04$
\\ \cline{2-9}
&  ReAct              
& $7.66 \pm 0.02$ & $83.01 \pm 0.97$ & $12.33 \pm 0.48$ & $80.15 \pm 0.46$ & $13.76 \pm 0.23$ & $82.88 \pm 0.08$ & $12.39 \pm 0.04$
\\ \cline{2-9}
& VIM
& $8.79 \pm 0.18$ & $83.14 \pm 3.71$ & $12.83 \pm 1.77$ & $\mathbf{85.91 \pm 0.78}$ & $\mathbf{11.44 \pm 0.46}$ & $77.76 \pm 0.16$ & $15.52 \pm 0.13$
\\ 
\end{tabular}
}

\end{table}

\begin{table}[htbp]
\centering
\caption{The mean and standard deviation over 3-folds \areaSCODTpr$\downarrow$ in \% points for \textit{tuned} selective classifiers constructed from an ID classifier $h(x)$ and selectors $c(x)=\leftbb s(x)\leq \lambda \rightbb$. Results are shown for ID CIFAR-100 and several possible OOD datasets. Rows of the Table correspond to different scores $s(x)$. The relative cost is $\alpha = 0.5$. The best results per dataset are highlighted in green. The best results with contemporary OODD scores are shown in bold.}
\resizebox{\linewidth}{!}{%
\begin{tabular}{ll|c|c|c|c|c|c}
\multicolumn{2}{c|}{{\backslashbox{\it  Score}{\it  Dataset}}} &  cifar10 &  mnist &  places365 &  svhn & textures & tin\\ \hline
\multirow{9}{*}{\rotatebox{90}{\bf Single Score}} 
& ASH \cite{Djurisic-ASH-ICLR2023} 
& $15.93 \pm 0.24$ & $15.56 \pm 0.13$ & $14.79 \pm 0.05$ & $11.37 \pm 0.60$ & $13.81 \pm 0.25$ & $14.21 \pm 0.17$
\\ \cline{2-8}
&  EBO \cite{Weitang-EBO-NIPS2020}            
& $14.39 \pm 0.07$ & $14.33 \pm 0.67$ & $14.16 \pm 0.13$ & $12.90 \pm 0.90$ & $14.74 \pm 0.45$ & $12.54 \pm 0.04$
\\ \cline{2-8}
&  GradNorm~\cite{Huang-Importance-NIPS2021}             
& $21.92 \pm 0.16$ & $24.41 \pm 0.48$ & $22.24 \pm 0.14$ & $18.61 \pm 2.40$ & $24.79 \pm 0.15$ & $22.11 \pm 0.38$
\\ \cline{2-8}
&  KNN~\cite{pmlr-v162-sun22d}          
& $15.13 \pm 0.08$ & $12.46 \pm 0.79$ & $13.93 \pm 0.18$ & $11.57 \pm 0.48$ & $11.81 \pm 0.44$ & $11.97 \pm 0.07$
\\ \cline{2-8}
&  MLS \cite{Hendrycks-MLS-ICML2022}              
& $14.19 \pm 0.08$ & $14.34 \pm 0.72$ & $13.92 \pm 0.14$ & $12.97 \pm 0.78$ & $14.60 \pm 0.46$ & $12.35 \pm 0.05$
\\ \cline{2-8}
& MSP \cite{Hendrycks-baseline-ICLR17}  
& $13.86 \pm 0.09$ & $15.06 \pm 0.93$ & $13.49 \pm 0.16$ & $13.89 \pm 0.51$ & $14.44 \pm 0.41$ & $12.06 \pm 0.13$
\\ \cline{2-8}
&  ODIN \cite{liang2018enhancing}               
& $14.98 \pm 0.07$ & $12.17 \pm 0.64$ & $14.34 \pm 0.14$ & $16.80 \pm 0.39$ & $14.40 \pm 0.55$ & $13.25 \pm 0.04$
\\ \cline{2-8}
&  ReAct \cite{Sun-NIPS2021}                
& $14.50 \pm 0.02$ & $14.65 \pm 0.79$ & $13.82 \pm 0.06$ & $12.33 \pm 0.48$ & $13.76 \pm 0.23$ & $12.39 \pm 0.04$
\\ \cline{2-8}
&  VIM \cite{Wang-ViM-CVPR2022}           
& $18.29 \pm 0.21$ & $13.45 \pm 0.59$ & $16.47 \pm 0.10$ & $12.83 \pm 1.77$ & $11.44 \pm 0.46$ & $15.52 \pm 0.13$
\\ 
\hline
& \Algo{} LR 
& $19.18 \pm 0.18$ & $11.29 \pm 0.09$ & $16.66 \pm 0.60$ & $11.53 \pm 0.20$ & $14.54 \pm 0.20$ & $12.28 \pm 0.22$
\\ 
\cline{2-8}
&  Clean LR
& $12.10 \pm 0.24$ & $11.33 \pm 0.10$ & $15.60 \pm 0.29$ & $10.98 \pm 0.11$ & $12.93 \pm 0.16$ & $11.78 \pm 0.12$
\\
\hline\hline
\multirow{8}{*}{\rotatebox{90}{{\bf Linear} (MSP , -)}} 
& ASH \cite{Djurisic-ASH-ICLR2023} 
& $13.70 \pm 0.10$ & $13.96 \pm 0.51$ & $13.17 \pm 0.10$ & $\mathbf{10.98 \pm 0.59}$ & $12.96 \pm 0.08$ & $11.90 \pm 0.12$
\\ \cline{2-8}
&  EBO \cite{Weitang-EBO-NIPS2020}            
& $13.56 \pm 0.09$ & $13.99 \pm 0.79$ & $13.23 \pm 0.14$ & $12.58 \pm 0.86$ & $14.03 \pm 0.41$ & $11.67 \pm 0.06$ 
\\ \cline{2-8}
&  GradNorm~\cite{Huang-Importance-NIPS2021}             
& $13.86 \pm 0.09$ & $14.91 \pm 0.88$ & $13.48 \pm 0.16$ & $13.31 \pm 0.96$ & $14.40 \pm 0.38$ & $12.06 \pm 0.13$
\\ \cline{2-8}
&  KNN~\cite{pmlr-v162-sun22d}          
& $13.58 \pm 0.04$ & $12.31 \pm 0.83$ & $\mathbf{13.03 \pm 0.18}$ & $11.39 \pm 0.44$ & $11.59 \pm 0.47$ & $\mathbf{11.35 \pm 0.11}$
\\ \cline{2-8}
&  MLS \cite{Hendrycks-MLS-ICML2022}              
& $13.56 \pm 0.09$ & $14.06 \pm 0.82$ & $13.23 \pm 0.14$ & $12.72 \pm 0.77$ & $14.04 \pm 0.41$ & $11.68 \pm 0.07$
\\ \cline{2-8}
&  ODIN \cite{liang2018enhancing}               
& $13.66 \pm 0.09$ & $\mathbf{11.83 \pm 0.69}$ & $13.17 \pm 0.15$ & $13.85 \pm 0.51$ & $13.48 \pm 0.48$ & $11.81 \pm 0.09$
\\ \cline{2-8}
&  ReAct \cite{Sun-NIPS2021}                
& $13.60 \pm 0.08$ & $14.24 \pm 0.86$ & $13.17 \pm 0.12$ & $12.09 \pm 0.46$ & $13.49 \pm 0.24$ & $11.65 \pm 0.06$
\\ \cline{2-8}
&  VIM \cite{Wang-ViM-CVPR2022}           
& $13.74 \pm 0.05$ & $12.62 \pm 0.71$ & $13.12 \pm 0.18$ & $11.41 \pm 1.11$ & $\mathbf{10.30 \pm 0.42}$ & $11.70 \pm 0.14$
\\ 
\hline
& \Algo{} LR 
& $9.62 \pm 0.07$ & $3.09 \pm 0.06$ & $8.57 \pm 0.27$ & $3.20 \pm 0.05$ & $6.78 \pm 0.23$ & $4.22 \pm 0.12$
\\ 
\cline{2-8}
&  Clean LR
& \cellcolor{\bestcolor}$6.63 \pm 0.17$ & \cellcolor{\bestcolor}$3.08 \pm 0.06$ & \cellcolor{\bestcolor}$7.71 \pm 0.22$ & \cellcolor{\bestcolor}$3.10 \pm 0.06$ & \cellcolor{\bestcolor}$5.73 \pm 0.08$ & \cellcolor{\bestcolor}$3.60 \pm 0.16$
\\
\hline\hline
\multirow{8}{*}{\rotatebox{90}{{\bf SIRC} (MSP , -)}} 
& ASH \cite{Djurisic-ASH-ICLR2023} 
& $13.68 \pm 0.10$ & $13.94 \pm 0.50$ & $13.14 \pm 0.10$ & $11.15 \pm 0.43$ & $12.95 \pm 0.05$ & $11.87 \pm 0.12$
\\ \cline{2-8}
&  EBO \cite{Weitang-EBO-NIPS2020}            
& $\mathbf{13.54 \pm 0.09}$ & $14.03 \pm 0.84$ & $13.21 \pm 0.14$ & $12.66 \pm 0.81$ & $14.01 \pm 0.42$ & $11.66 \pm 0.06$
\\ \cline{2-8}
&  GradNorm~\cite{Huang-Importance-NIPS2021}             
& $13.69 \pm 0.09$ & $14.74 \pm 0.85$ & $13.36 \pm 0.15$ & $12.74 \pm 1.09$ & $14.24 \pm 0.37$ & $11.98 \pm 0.11$
\\ \cline{2-8}
&  KNN~\cite{pmlr-v162-sun22d}          
& $13.55 \pm 0.04$ & $12.59 \pm 0.91$ & $\mathbf{13.03 \pm 0.17}$ & $11.60 \pm 0.32$ & $11.88 \pm 0.47$ & $\mathbf{11.35 \pm 0.10}$
\\ \cline{2-8}
&  MLS \cite{Hendrycks-MLS-ICML2022}              
& $13.55 \pm 0.09$ & $14.12 \pm 0.86$ & $13.21 \pm 0.14$ & $12.81 \pm 0.73$ & $14.03 \pm 0.42$ & $11.67 \pm 0.07$
\\ \cline{2-8}
&  ODIN \cite{liang2018enhancing}               
& $13.64 \pm 0.09$ & $12.31 \pm 0.76$ & $13.13 \pm 0.15$ & $13.85 \pm 0.51$ & $13.57 \pm 0.47$ & $11.80 \pm 0.09$ 
\\ \cline{2-8}
&  ReAct \cite{Sun-NIPS2021}                
& $13.58 \pm 0.08$ & $14.24 \pm 0.87$ & $13.15 \pm 0.12$ & $12.31 \pm 0.49$ & $13.59 \pm 0.29$ & $11.63 \pm 0.06$
\\ \cline{2-8}
&  VIM \cite{Wang-ViM-CVPR2022}           
& $13.70 \pm 0.05$ & $12.48 \pm 0.73$ & $13.08 \pm 0.19$ & $11.38 \pm 1.04$ & $10.34 \pm 0.41$ & $11.64 \pm 0.15$
\\ 
\hline
& \Algo{} LR 
& $9.41 \pm 0.31$ & $3.09 \pm 0.06$ & $12.81 \pm 0.28$ & $3.21 \pm 0.07$ & $7.49 \pm 0.81$ & $6.55 \pm 0.46$
\\ 
\cline{2-8}
&  Clean LR
& $7.67 \pm 0.57$ & $3.09 \pm 0.06$ & $13.16 \pm 0.26$ & $3.12 \pm 0.06$ & $7.20 \pm 1.51$ & $4.28 \pm 0.50$
\\
\end{tabular}
}
\end{table}

\begin{table}[htbp]
\centering
\caption{The mean and standard deviation over 3-folds \areaSCODTpr$\downarrow$ in \% points for \textit{plugin} selective classifiers constructed from an ID classifier $h(x)$ and selectors $c(x)=\leftbb s(x)\leq \lambda \rightbb$. Results are shown for ID CIFAR-100 and several possible OOD datasets. Rows of the Table correspond to different scores $s(x)$. The relative cost is $\alpha = 0.5$. The best results per dataset are highlighted in green. The best results with contemporary OODD scores are shown in bold.}
\resizebox{\linewidth}{!}{%
\begin{tabular}{ll|c|c|c|c|c|c}
\multicolumn{2}{c|}{{\backslashbox{\it  Score}{\it  Dataset}}} &  cifar10 &  mnist &  places365 &  svhn & textures & tin\\ \hline
\multirow{9}{*}{\rotatebox{90}{\bf Single Score}} 
& ASH \cite{Djurisic-ASH-ICLR2023} 
& $15.93 \pm 0.24$ & $15.56 \pm 0.13$ & $14.79 \pm 0.05$ & $\mathbf{11.37 \pm 0.60}$ & $13.81 \pm 0.25$ & $14.21 \pm 0.17$
\\ \cline{2-8}
&  EBO \cite{Weitang-EBO-NIPS2020}            
& $14.39 \pm 0.07$ & $14.33 \pm 0.67$ & $14.16 \pm 0.13$ & $12.90 \pm 0.90$ & $14.74 \pm 0.45$ & $12.54 \pm 0.04$
\\ \cline{2-8}
&  GradNorm~\cite{Huang-Importance-NIPS2021}             
& $21.92 \pm 0.16$ & $24.41 \pm 0.48$ & $22.24 \pm 0.14$ & $18.61 \pm 2.40$ & $24.79 \pm 0.15$ & $22.11 \pm 0.38$
\\ \cline{2-8}
&  KNN \cite{pmlr-v162-sun22d}          
& $15.13 \pm 0.08$ & $12.46 \pm 0.79$ & $13.93 \pm 0.18$ & $11.57 \pm 0.48$ & $11.81 \pm 0.44$ & $11.97 \pm 0.07$
\\ \cline{2-8}
&  MLS \cite{Hendrycks-MLS-ICML2022}                 
& $14.19 \pm 0.08$ & $14.34 \pm 0.72$ & $13.92 \pm 0.14$ & $12.97 \pm 0.78$ & $14.60 \pm 0.46$ & $12.35 \pm 0.05$
\\ \cline{2-8}
& MSP \cite{Hendrycks-baseline-ICLR17}  
& $13.86 \pm 0.09$ & $15.06 \pm 0.93$ & $13.49 \pm 0.16$ & $13.89 \pm 0.51$ & $14.44 \pm 0.41$ & $12.06 \pm 0.13$
\\ \cline{2-8}
&  ODIN \cite{liang2018enhancing}               
& $14.98 \pm 0.07$ & $12.17 \pm 0.64$ & $14.34 \pm 0.14$ & $16.80 \pm 0.39$ & $14.40 \pm 0.55$ & $13.25 \pm 0.04$
\\ \cline{2-8}
&  ReAct \cite{Sun-NIPS2021}                
& $14.50 \pm 0.02$ & $14.65 \pm 0.79$ & $13.82 \pm 0.06$ & $12.33 \pm 0.48$ & $13.76 \pm 0.23$ & $12.39 \pm 0.04$
\\ \cline{2-8}
&  VIM \cite{Wang-ViM-CVPR2022}           
& $18.29 \pm 0.21$ & $13.45 \pm 0.59$ & $16.47 \pm 0.10$ & $12.83 \pm 1.77$ & $11.44 \pm 0.46$ & $15.52 \pm 0.13$
\\ 
\hline
& \Algo{} LR 
& $19.18 \pm 0.18$ & $11.29 \pm 0.09$ & $16.66 \pm 0.60$ & $11.53 \pm 0.20$ & $14.54 \pm 0.20$ & $12.28 \pm 0.22$
\\ 
\cline{2-8}
&  Clean LR
& $12.10 \pm 0.24$ & $11.33 \pm 0.10$ & $15.60 \pm 0.29$ & $10.98 \pm 0.11$ & $12.93 \pm 0.16$ & $11.78 \pm 0.12$
\\
\hline\hline
\multirow{8}{*}{\rotatebox{90}{{\bf Linear} (MSP , -)}} 
& ASH \cite{Djurisic-ASH-ICLR2023} 
& $15.66 \pm 0.20$ & $15.41 \pm 0.19$ & $14.56 \pm 0.05$ & $11.44 \pm 0.52$ & $13.74 \pm 0.21$ & $13.91 \pm 0.15$
\\ \cline{2-8}
&  EBO \cite{Weitang-EBO-NIPS2020}            
& $14.18 \pm 0.08$ & $14.30 \pm 0.71$ & $13.92 \pm 0.14$ & $12.91 \pm 0.82$ & $14.57 \pm 0.45$ & $12.32 \pm 0.05$
\\ \cline{2-8}
&  GradNorm~\cite{Huang-Importance-NIPS2021}             
& $21.91 \pm 0.16$ & $24.40 \pm 0.48$ & $22.22 \pm 0.13$ & $18.60 \pm 2.39$ & $24.78 \pm 0.15$ & $22.09 \pm 0.37$
\\ \cline{2-8}
&  KNN \cite{pmlr-v162-sun22d}          
& $14.20 \pm 0.04$ & $\mathbf{14.12 \pm 0.85}$ & $13.46 \pm 0.17$ & $13.00 \pm 0.21$ & $13.36 \pm 0.42$ & $\mathbf{11.81 \pm 0.12}$
\\ \cline{2-8}
&  MLS \cite{Hendrycks-MLS-ICML2022}                 
& $14.08 \pm 0.08$ & $14.35 \pm 0.75$ & $13.79 \pm 0.14$ & $13.01 \pm 0.74$ & $14.51 \pm 0.45$ & $12.24 \pm 0.06$
\\ \cline{2-8}
&  ODIN \cite{liang2018enhancing}               
& $13.94 \pm 0.10$ & $14.95 \pm 0.87$ & $13.58 \pm 0.18$ & $14.09 \pm 0.51$ & $14.43 \pm 0.44$ & $12.17 \pm 0.14$
\\ \cline{2-8}
&  ReAct \cite{Sun-NIPS2021}                
& $14.28 \pm 0.02$ & $14.60 \pm 0.84$ & $13.61 \pm 0.07$ & $12.42 \pm 0.42$ & $13.69 \pm 0.23$ & $12.19 \pm 0.02$
\\ \cline{2-8}
&  VIM \cite{Wang-ViM-CVPR2022}           
& $17.32 \pm 0.16$ & $13.13 \pm 0.61$ & $15.64 \pm 0.13$ & $12.32 \pm 1.63$ & $\mathbf{10.99 \pm 0.45}$ & $14.54 \pm 0.13$
\\ 
\hline
& \Algo{} LR 
& $13.13 \pm 0.13$ & \cellcolor{\bestcolor}$3.10 \pm 0.06$ & $11.60 \pm 0.37$ & $3.25 \pm 0.06$ & $7.12 \pm 0.28$ & $4.49 \pm 0.21$
\\ 
\cline{2-8}
&  Clean LR
& \cellcolor{\bestcolor} $7.56 \pm 0.20$ & \cellcolor{\bestcolor}$3.10 \pm 0.06$ & \cellcolor{\bestcolor}$11.27 \pm 0.14$ & \cellcolor{\bestcolor}$3.15 \pm 0.06$ & \cellcolor{\bestcolor}$6.07 \pm 0.10$ & \cellcolor{\bestcolor}$3.75 \pm 0.24$
\\
\hline\hline
\multirow{8}{*}{\rotatebox{90}{{\bf SIRC} (MSP , -)}} 
& ASH \cite{Djurisic-ASH-ICLR2023} 
& $13.82 \pm 0.09$ & $14.85 \pm 0.85$ & $13.41 \pm 0.15$ & $13.50 \pm 0.44$ & $14.21 \pm 0.37$ & $12.02 \pm 0.12$
\\ \cline{2-8}
&  EBO \cite{Weitang-EBO-NIPS2020}            
& $13.82 \pm 0.09$ & $14.97 \pm 0.93$ & $13.44 \pm 0.16$ & $13.77 \pm 0.53$ & $14.38 \pm 0.41$ & $12.00 \pm 0.12$
\\ \cline{2-8}
&  GradNorm~\cite{Huang-Importance-NIPS2021}             
& $13.82 \pm 0.09$ & $14.95 \pm 0.88$ & $13.46 \pm 0.16$ & $13.54 \pm 0.64$ & $14.39 \pm 0.38$ & $12.09 \pm 0.11$
\\ \cline{2-8}
&  KNN \cite{pmlr-v162-sun22d}          
& $\mathbf{13.80 \pm 0.08}$ & $14.75 \pm 0.93$ & $13.37 \pm 0.17$ & $13.48 \pm 0.46$ & $14.01 \pm 0.42$ & $11.91 \pm 0.13$
\\ \cline{2-8}
&  MLS \cite{Hendrycks-MLS-ICML2022}                 
& $13.82 \pm 0.09$ & $14.97 \pm 0.93$ & $13.45 \pm 0.16$ & $13.78 \pm 0.53$ & $14.38 \pm 0.41$ & $12.00 \pm 0.12$
\\ \cline{2-8}
&  ODIN \cite{liang2018enhancing}               
& $13.84 \pm 0.09$ & $14.89 \pm 0.93$ & $13.44 \pm 0.16$ & $13.93 \pm 0.51$ & $14.37 \pm 0.41$ & $12.03 \pm 0.12$
\\ \cline{2-8}
&  ReAct \cite{Sun-NIPS2021}                
& $13.81 \pm 0.09$ & $14.96 \pm 0.92$ & $13.43 \pm 0.16$ & $13.73 \pm 0.52$ & $14.34 \pm 0.40$ & $11.98 \pm 0.12$
\\ \cline{2-8}
&  VIM \cite{Wang-ViM-CVPR2022}           
& $13.84 \pm 0.08$ & $14.71 \pm 0.96$ & $\mathbf{13.34 \pm 0.17}$ & $13.48 \pm 0.37$ & $13.45 \pm 0.42$ & $11.95 \pm 0.14$
\\ 
\hline
& \Algo{} LR 
& $12.36 \pm 0.23$ & \cellcolor{\bestcolor}$3.10 \pm 0.06$ & $13.46 \pm 0.14$ & $3.39 \pm 0.20$ & $11.16 \pm 1.14$ & $10.23 \pm 0.36$
\\ 
\cline{2-8}
&  Clean LR
& $10.70 \pm 0.72$ & \cellcolor{\bestcolor}$3.10 \pm 0.06$ & $13.45 \pm 0.16$ & \cellcolor{\bestcolor}$3.15 \pm 0.06$ & $10.44 \pm 1.51$ & $6.26 \pm 0.92$   
\\
\end{tabular}
}
\end{table}

\clearpage

% Theorem 1, 2, and 3
\section{Proof of Theorem~\ref{thm:BayesClsIsOptimal}}

The proof is essentially identical to that of \cite[Theorem 1]{Franc-Optimal-JMLR2023}. We include it here for the sake of completeness.

For any ID classifier $h$ and a
stochastic selector $c$, the definition of $h_B$ allows to derive $\Rsel(h_B,c) \le \Rsel(h,c)$ as follows:
\begin{align*}
\Rsel(h_B,c)&=\frac{1}{\tpr(c)}\int\limits_{\SX}\sum\limits_{y\in\SY}p(x,y)\,\ell(y,h_B(x))\,c(x)\,dx \\
&= \frac{1}{\tpr(c)}\int\limits_{\SX}p(x)c(x)\left(\sum\limits_{y\in\SY} p(y\,|\,x)\,\ell(y,h_B(x))\right)\,dx\\
&\le \frac{1}{\tpr(c)}\int\limits_{\SX}p(x)c(x)\left(\sum\limits_{y\in\SY} p(y\,|\,x)\,\ell(y,h(x))\right)\,dx\\
&=\frac{1}{\tpr(c)}\int\limits_{\SX}\sum\limits_{y\in\SY}p(x,y)\,\ell(y,h(x))\,c(x)\,dx\\
&=\Rsel(h,c)\,.
\end{align*}

\section{Proof of Theorem~\ref{thm:SingleCostModel}}

If $c^*$ is an optimal solution to 
Problem~\ref{prob:SingleCostModelFixedCls} fulfilling $\tpr(c^*)= a\cdot \tpr_{\rm min}$, where $a>1$, then $c=c^*/a$ is also an optimal solution such that $\tpr(c)=\tpr_{\rm min}$. Hence, to find an optimal solution to Problem~\ref{prob:SingleCostModelFixedCls}, it suffices to minimize the objective function
\begin{align*}
&\frac{1-\relCost}{\tpr_{\rm min}}\int_{\SX} R(h_B,x) c(x) dx + \relCost \int_{\SX} p_O(x)c(x)dx \\ &= \int_{\SX} \left[ \frac{1-\relCost}{\tpr_{\rm min}}R(h_B,x)+\relCost p_O(x) \right] c(x)dx
\end{align*}
subject to
\begin{align*}
\int_{\SX}p_I(x)c(x) \ge \tpr_{\rm min} \,.
\end{align*}

By~\cite[Theorem 3]{Franc-Optimal-JMLR2023}, for a bounded risk function $f: \SX \to \Re_+$, a probability distribution $p: \SX \to \Re_+$, and $b\in \Re_+$, whenever the problem
    \begin{equation*}
       \min_{c\in[0,1]^\SX} \int_{\SX} f(x)c(x)dx
\qquad\mbox{s.t.}\qquad \int_{\SX} p(x)c(x) \geq b
\end{equation*}
is feasible, the set of its optimal solutions contains
\begin{equation*}
   c^*(x)=\left \{ \begin{array}{rcl}
     0 & \mbox{if} & \frac{f(x)}{p(x)} > \lambda \\
     \tau & \mbox{if} & \frac{f(x)}{p(x)} = \lambda \\
     1 & \mbox{if} & \frac{f(x)}{p(x)} < \lambda 
   \end{array}
   \right .
\end{equation*}
for suitable $\tau\in [0,1]$ and $\lambda\in \Re$. This implies that Problem~\ref{prob:SingleCostModelFixedCls} has an optimal solution $c^*$ induced by the score function $s(x)=\frac{1-\relCost}{\tpr_{\rm min}}r(x)+\relCost g(x)$ and the threshold value $\lambda$. It is further easy to see that the score $s(x)$ can be replaced by $s'(x)=r(x)+\frac{\relCost \tpr_{\rm min}}{1-\relCost} g(x)$ if the threshold is set to $\frac{\tpr_{\rm min}}{1-\relCost} \lambda$.

\section{Proof of Theorem~\ref{thm:SCODisNotPAC}}
It was proved in~\cite{Zhen-IsOODLearnable-NIPS2022}  that the problem
\begin{equation}\label{prob:PACnotLearnable}
    \min_{h\in\SY^\SX,c\in[0,1]^\SX} R_C(h,c)
\end{equation}
where
\begin{equation*}
\begin{split}
    R_C(h,c) = \int_{\SX} \Bigl[(1-\pi_O)R(h,x)c(x)+L_{FN}(1-\pi_O)p_I(x)(1-c(x))\\+L_{FP}\pi_O  p_O(x)c(x) \Bigr] dx
\end{split}
\end{equation*}
is not PAC learnable for any triple of constants $\pi_O\in (0,1)$, $L_{FN}>0$, $L_{FP}>0$.

Observe that
\[
R_C(h,c) = L_{FN}(1-\pi_O) + \int_{\SX} G(h,x)c(x) dx
\]
where
\[
G(h,x) = (1-\pi_O)R(h,x)-L_{FN}(1-\pi_O)p_I(x)+L_{FP}\pi_O  p_O(x) \,.
\]
An optimal solution $(h_B, c^*)$ to Problem~\ref{prob:PACnotLearnable} can thus be established by prescribing $c^*(x)=1$ whenever $G(h_B,x)<0$, and $c^*(x)=0$ whenever $G(h_B,x)\ge0$. This means that the considered optimal solution is equivalently determined by a score function $s_1(x)=r(x)+\frac{\pi_O L_{FP}}{1-\pi_O}g(x)$ and a threshold $\lambda_1=L_{FN}$.

By Theorem~\ref{thm:SingleCostModel}, there are optimal solutions to Problem~\ref{prob:SingleCostModelFixedCls} determined by the score function $s_2(x)= r(x) + \frac{\relCost\tpr_{\rm min}}{1-\relCost} g(x)$ and a threshold $\lambda_2>0$.
Hence, for a given instance of Problem~\ref{prob:SingleCostModelFixedCls} with an optimal solution $(h_B, c^*)$ consistent with the score function $s_2$, we obtain an instance of Problem~\ref{prob:PACnotLearnable} for which $(h_B, c^*)$ is also an optimal solution just by setting $\pi_O:=\relCost$, $L_{FP}:=\tpr_{\rm min}$, and $L_{FN}:=\lambda_2$. In accordance with the proof of Theorem~\ref{thm:SingleCostModel}, we assume that $\tpr(c^*)=\tpr_{\rm min}$.

Let us now consider we have a solution $(h,c)$ to Problem~\ref{prob:SingleCostModelFixedCls} such that
\begin{equation}
R_S(h,c)-R_S(h_B,c^*) \le \veps_1 \,, \label{eq:PAC_proof_1}
\end{equation}
\begin{equation}
\tpr(c) \ge \tpr_{\rm min} - \veps_2 \label{eq:PAC_proof_2}
\end{equation}
for some $\veps_1, \veps_2 \in \Re_+$.

Our goal is to show that
\begin{equation}
R_C(h,c) - R_C(h_B,c^*) \le \veps_1\tpr(c) + \relCost \veps_2 \fpr(c) \,. \label{eq:PAC_proof_3}
\end{equation}
As the right-hand side expression can be made arbitrarily close to zero by choosing sufficiently small values of $\veps_1$ and $\veps_2$, inequality~\equ{eq:PAC_proof_3} enforces that Problem~\ref{prob:SingleCostModelFixedCls} is not PAC learnable, otherwise Problem~\ref{prob:PACnotLearnable} would be PAC learnable as well, which is impossible due to ~\cite[Theorem 4]{Zhen-IsOODLearnable-NIPS2022}.

Inequality~\equ{eq:PAC_proof_1} implies
\[
\frac{1-\relCost}{\tpr(c)} \int_{\SX} R(h,x)c(x)dx - \frac{1-\relCost}{\tpr_{\rm min}} \int_{\SX} R(h_B,c^*)c^*(x)dx + \relCost\, \fpr(c) - \relCost\, \fpr(c^*) \le \veps_1
\]
which can further be rewritten to
\begin{align}\label{eq:PAC_proof_4}
& (1-\relCost) \int_{\SX} R(h,x)c(x)dx - (1-\relCost) \int_{\SX} R(h_B,c^*)c^*(x)dx 
\\&
\le \relCost\, \tpr(c) \left(\fpr(c^*) - \fpr(c)\right) \\ \nonumber
& \qquad + (1-\relCost) \frac{\tpr(c) - \tpr_{\rm min}}{\tpr_{\rm min}} \int_{\SX} R(h_B,x)c^*(x)dx + \veps_1 \tpr(c) \,.
\end{align}

The score function $s_2$ ensures that
\[
R(h_B,x)c^*(x)+\frac{\relCost}{1-\relCost}\tpr_{\rm min}p_O(x)c^*(x) \le \lambda_2 p_I(x)c^*(x)
\]
for all $x\in\SX$, yielding
\begin{equation}\label{eq:PAC_proof_5}
    \int_{\SX} R(h_B,x)c^*(x)dx + \frac{\relCost}{1-\relCost} \tpr_{\rm min} \fpr(c^*) \le \lambda_2 \tpr(c^*) = \lambda_2\tpr_{\rm \min} \,.
\end{equation}

Now, using~\equ{eq:PAC_proof_2}, \equ{eq:PAC_proof_4} and~\equ{eq:PAC_proof_5}, we derive
\begin{align*}
    R_C&(h,c) - R_C(h_B,c^*) = (1-\relCost) \int_{\SX} R(h,x)c(x)dx \\
    & - (1-\relCost) \int_{\SX} R(h_B,c^*)c^*(x)dx
    - \lambda_2 (1-\relCost) \tpr(c) + \lambda_2 (1-\relCost)\tpr_{\rm min} \\
    & + \relCost\,\tpr_{\rm min}\fpr(c) - \relCost\,\tpr_{\rm min}\fpr(c^*) \\
    \le\, & \veps_1 \tpr(c) + (1-\relCost) \frac{\tpr(c) - \tpr_{\rm min}}{\tpr_{\rm min}} \left(\lambda_2\tpr_{\rm min} - \frac{\relCost}{1-\relCost} \tpr_{\rm min} \fpr(c^*) \right) \\ 
    & + \relCost\, \tpr(c)(\fpr(c^*)-\fpr(c)) - \lambda_2(1-\relCost)(\tpr(c) -\tpr_{\rm min}) \\ & -\relCost\,\tpr_{\rm min}(\fpr(c^*)-\fpr(c)) \\
    =\, & \veps_1\tpr(c)+\lambda_2(1-\relCost)(\tpr(c)-\tpr_{\rm min})+\relCost\,\fpr(c^*)(\tpr_{\rm min}-\tpr(c)) \\
    & -\lambda_2(1-\relCost)(\tpr(c)-\tpr_{\rm min}) - \relCost(\fpr(c^*)-\fpr(c))(\tpr_{\rm min} - \tpr(c)) \\
    =\, & \veps_1 \tpr(c) + \relCost\, \fpr(c) (\tpr(c)-\tpr_{\rm min}) \le \veps_1 \tpr(c) + \relCost \veps_2 \fpr(c) \,.
\end{align*}

\section{Proof of Theorem~\ref{thm:CorrectedSigmoid}}

Assume there exists a map $\#\phi\colon\SX\rightarrow\Re^d$ that renders the ID and ODD data normal distributed, i.e.,
\begin{equation}    
  \label{equ:NormalAssumption}
  \begin{array}{rcl}
   p_I(x) &= &(2\pi)^{-\frac{d}{2}}\det{\#C}^{-\frac{1}{2}}\exp \big ( -\frac{1}{2}(\#\phi(x)-\#\mu_I)^T\#C^{-1}(\#\phi(x)-\#\mu_I)\big )\\
   p_O(x) &= &(2\pi)^{-\frac{d}{2}}\det{\#C}^{-\frac{1}{2}}\exp \big ( -\frac{1}{2}(\#\phi(x)-\#\mu_O)^T\#C^{-1}(\#\phi(x)-\#\mu_O)\big )
   \end{array}
\end{equation}
where $(\#\mu_I,\#C)$ and $(\#\mu_O,\#C)$ are mean and covariance matrix of the ID and OOD, respectively. Under assumption~\cref{equ:NormalAssumption}, the OOD/ID likelihood ratio reads
\begin{equation}
  \label{equ:LikelihoodRatioNormal}
    \frac{p_O(x)}{p_I(x)} = \exp \big ( \frac{1}{2}(\#\mu_O^T\#C^{-1}\#\mu_O - \#\mu_I^T\#C^{-1}\#\mu_I \big ) + \#\phi(x)^T \#C^{-1}\big(\#\mu_I-\#\mu_O) \big )\:.    
\end{equation}
From~\equ{equ:IDvsUnlabMixture}, which defines the mixture of ID and unlabeled mixture of ID and OOD, we can derive that
\begin{equation}
\label{equ:MixturePosterior}
\begin{array}{rcl}
    p(z=I\mid x) &= &\dfrac{p(x,z=I)}{p(x,z=I)+p(x,z=U)}\\
    &=& \dfrac{(1-\pi_U)p_I(x)}{(1-\pi_U)p_I(x)+\pi_U \pi_O p_O(x)+(1-\pi_O)\pi_U p_I(x)}\\
    &=& \dfrac{1}{1+\dfrac{\pi_U \pi_O}{(1-\pi_U)} \dfrac{p_O(x)}{p_I(x)}+\pi_U\dfrac{(1-\pi_O)}{(1-\pi_U)} }\:.
\end{array}
\end{equation}
After substituting~\cref{equ:LikelihoodRatioNormal} to~\cref{equ:MixturePosterior}, we obtain
\begin{equation}
\begin{array}{rcl}
(z=I\mid x;\#\theta,a) &=& \dfrac{1}{\splitfrac{1+\pi_U\dfrac{(1-\pi_O)}{(1-\pi_U)}+\exp\bigg(\ln\Big(\dfrac{\pi_U\pi_O}{(1-\pi_U)} \Big)}{+ \frac{1}{2}(\#\mu_O^T\#C^{-1}\#\mu_O - \#\mu_I^T\#C^{-1}\#\mu_I \big )+ \#\phi(x)^T \#C^{-1}\big(\#\mu_I-\#\mu_O) \bigg)}}  \\
   &=& \dfrac{1}{1+|a|+\exp\big( \#\theta^T[\#\phi(x);1]\big)}
\end{array}
\end{equation}
where
$$
  \#\theta = \bigg [\#C^{-1}\big(\#\mu_I-\#\mu_O)  ;  \ln\Big(\dfrac{\pi_U \pi_O}{(1-\pi_U)} \Big)+ \frac{1}{2}(\#\mu_O^T\#C^{-1}\#\mu_O - \#\mu_I^T\#C^{-1}\#\mu_I \big )\bigg ]
$$
and
$$
a= \pi_U\dfrac{(1-\pi_O)}{(1-\pi_U)}\:.
$$
Note that we use $\#u=[\#v;b]$ to denote a column vector obtained by extending the vector $\#v$ by a new coordinate $b$. This ends the proof.

\clearpage  % TODO REVIEW/FINAL: This \clearpage needs to be removed from both review and camera-ready versions.

%\bibliographystyle{splncs04}
%\bibliography{main}
\end{document}